\documentclass[10pt,twocolumn,letterpaper]{article}

\usepackage[pagenumbers]{cvpr} 

\usepackage{graphicx}
\usepackage{amsmath}
\usepackage{amssymb}
\usepackage{booktabs}
\usepackage{enumitem}
\usepackage{colortbl}
\usepackage{multirow}
\usepackage{overpic}

\usepackage[pagebackref,breaklinks,colorlinks]{hyperref}

\newcommand{\ourmethod}{NOPS\xspace}
\newcommand{\CC}[1]{\cellcolor{#1}}
\definecolor{novelcolor}{rgb}{0.8, 1., 0.9}

\usepackage[capitalize]{cleveref}
\crefname{section}{Sec.}{Secs.}
\Crefname{section}{Section}{Sections}
\Crefname{table}{Table}{Tables}
\crefname{table}{Tab.}{Tabs.}

\newcommand\blfootnote[1]{%
  \begingroup
  \renewcommand\thefootnote{}\footnote{#1}%
  \addtocounter{footnote}{-1}%
  \endgroup
}

\begin{document}

\title{Novel Class Discovery for 3D Point Cloud Semantic Segmentation}

\author{
Luigi Riz$^1$ \quad 
Cristiano Saltori$^2$ \quad 
Elisa Ricci$^{1,2}$ \quad 
Fabio Poiesi$^1$\vspace{5px}\\
\normalsize{
$^1$Fondazione Bruno Kessler \quad 
$^2$University of Trento
}}


\newcommand{\lblfig}[1]{\label{fig:#1}}

\twocolumn[{%
\renewcommand\twocolumn[1][]{#1}%
\vspace{-1cm}
\maketitle
\thispagestyle{empty}
\begin{center}
    \centering
    \includegraphics[width=\textwidth]{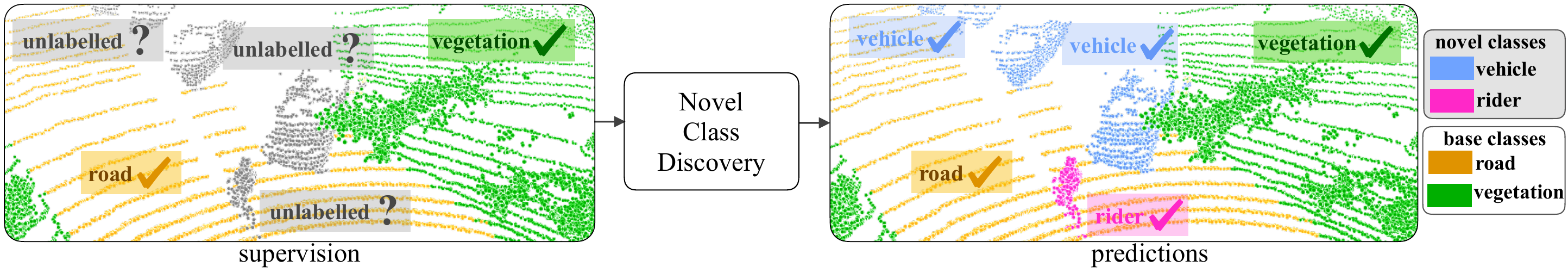}
    \vspace{-.5cm}
    \captionof{figure}{Novel Class Discovery for 3D point cloud semantic segmentation seeks to recognise novel classes by clustering unlabelled novel points with similar semantic features and by exploiting only the knowledge of a set of labelled samples corresponding to the base classes.}
    \label{fig:teaser}
\end{center}%
}]


\blfootnote{This project has received funding from the European Union’s Horizon Europe research and innovation programme under grant agreement No 101058589.
This work was also partially supported by the PRIN project LEGO-AI (Prot.~2020TA3K9N), the EU ISFP PROTECTOR (101034216) project and the EU H2020 MARVEL (957337) project and, it was carried out in the Vision and Learning joint laboratory of FBK and UNITN.}

\begin{abstract}

Novel class discovery (NCD) for semantic segmentation is the task of learning a model that can segment unlabelled (novel) classes using only the supervision from labelled (base) classes. This problem has recently been pioneered for 2D image data, but no work exists for 3D point cloud data. In fact, the assumptions made for 2D are loosely applicable to 3D in this case. This paper is presented to advance the state of the art on point cloud data analysis in four directions. Firstly, we address the new problem of NCD for point cloud semantic segmentation. Secondly, we show that the transposition of the only existing NCD method for 2D semantic segmentation to 3D data is suboptimal. Thirdly, we present a new method for NCD based on online clustering that exploits uncertainty quantification to produce prototypes for pseudo-labelling the points of the novel classes. Lastly, we introduce a new evaluation protocol to assess the performance of NCD for point cloud semantic segmentation. We thoroughly evaluate our method on SemanticKITTI and SemanticPOSS datasets, showing that it can significantly outperform the baseline. Project page: \url{https://github.com/LuigiRiz/NOPS}.
\end{abstract}

\section{Introduction}\label{sec:intro}

As humans, we are fairly skilled in organising new visual knowledge (novelty) into homogeneous groups, even when we do not know what we are observing. However, machines cannot perform this task satisfactorily without our supervision. The challenge here is mainly in the formulation of discriminative latent representations of the real world and in the quantification of the uncertainty when the novelty is observed \cite{han2019learning,zhong2021neighborhood,zhao2022novel}. The work of Han et al.~\cite{han2019learning} pioneered the formulation of the Novel Class Discovery (NCD) problem by defining it as the task that aims to classify the samples of an unlabelled dataset into different classes, i.e.~the \textit{novel samples}, by exploiting the knowledge of a set of labelled samples, i.e.~the \textit{base samples}. Note that the classes in the labelled and unlabelled datasets are disjoint.

NCD has been explored in the 2D image domain for classification \cite{han2019learning,fini2021unified,zhong2021neighborhood} and later for semantic segmentation \cite{zhao2022novel}. In particular, Zhao et al.~\cite{zhao2022novel} presented the first approach for NCD for 2D semantic segmentation. Two key assumptions were made by the authors. Firstly, at most one novel class is allowed in each image. Secondly, the new class belongs to a foreground object that can be found via saliency detection (e.g.~a man on a bicycle, where the bicycle is the novel class). Thanks to these assumptions, the authors could pool the features of each image into a single latent representation and cluster the representations of the whole dataset to discover clusters of novel classes. We argue that these two are important constraints that are not applicable to generic 3D data, in particular to point clouds captured with LiDAR sensors in real-world city-scale scenarios. One point cloud can contain more than one novel class, and the saliency for 3D data cannot be leveraged in the same way as that for 2D data. Although they are both related to the attraction of human fixations, 3D saliency is more related to the regional importance of 3D surfaces rather than the foreground/background distinction~\cite{Ran2021}. Our motivation in exploring NCD for the 3D setting is mainly driven by addressing these shortcomings.

In this paper, we explore the new problem of NCD for 3D point cloud semantic segmentation (see Fig.~\ref{fig:teaser}).
Given a partially human-annotated dataset, we jointly learn base and novel semantic classes by clustering unlabelled points with similar semantic features.
We adapt the method of Zhao et al.~\cite{zhao2022novel} (Entropy-based Uncertainty Modeling and Self-training - EUMS) for point cloud data and use it as our baseline.
We go beyond their formulation and, inspired by~\cite{caron2020unsupervised}, we integrate batch-level (online) clustering in our method and update prototypes during training in order to make clustering computationally tractable.
Cluster assignments are then used as training pseudo-labels.
We also exploit over-clustering to achieve a higher clustering accuracy as in EUMS.
Because point clouds contain multiple semantic classes, we cannot guarantee that all the classes appear in the point clouds within each batch, some will be missing.
Therefore, we design a queuing strategy to store important features over training time, which will be used for pseudo-labelling in the case of missing classes.
We further introduce a strategy for exploiting the pseudo-label uncertainty to promote the creation of reliable prototypes that we then exploit to produce higher-quality pseudo-labels.
Lastly, we produce two augmented views of the same point cloud and impose pseudo-label consistency amongst them.
We evaluate our approach on SemanticKITTI \cite{behley2019semantickitti, geiger2012cvpr, behley2021ijrr} and SemanticPOSS \cite{pan2020semanticposs}, introducing an evaluation protocol for NCD and point cloud segmentation that can be adopted in future works.
We empirically show that our approach largely outperforms our baseline in both datasets.
We also perform an extensive ablation study to demonstrate the importance of the different components of our method.

\vspace{1mm}
\noindent To summarise, our contributions are:
\setlist{nolistsep}
\begin{itemize}
    \item We address the new problem of NCD for 3D semantic segmentation;
    \item We show that the transposition of the only existing NCD method for 2D semantic segmentation \cite{zhao2022novel} to 3D data is suboptimal;
    \item We present a new method for NCD based on online clustering and uncertainty estimation, which we name it NOPS (NOvel Point Segmentation);
    \item We introduce a new evaluation protocol to assess the performance of NCD for 3D semantic segmentation.
\end{itemize}

\section{Related work}

\begin{figure*}[t]
    \centering
    \includegraphics[width=\textwidth]{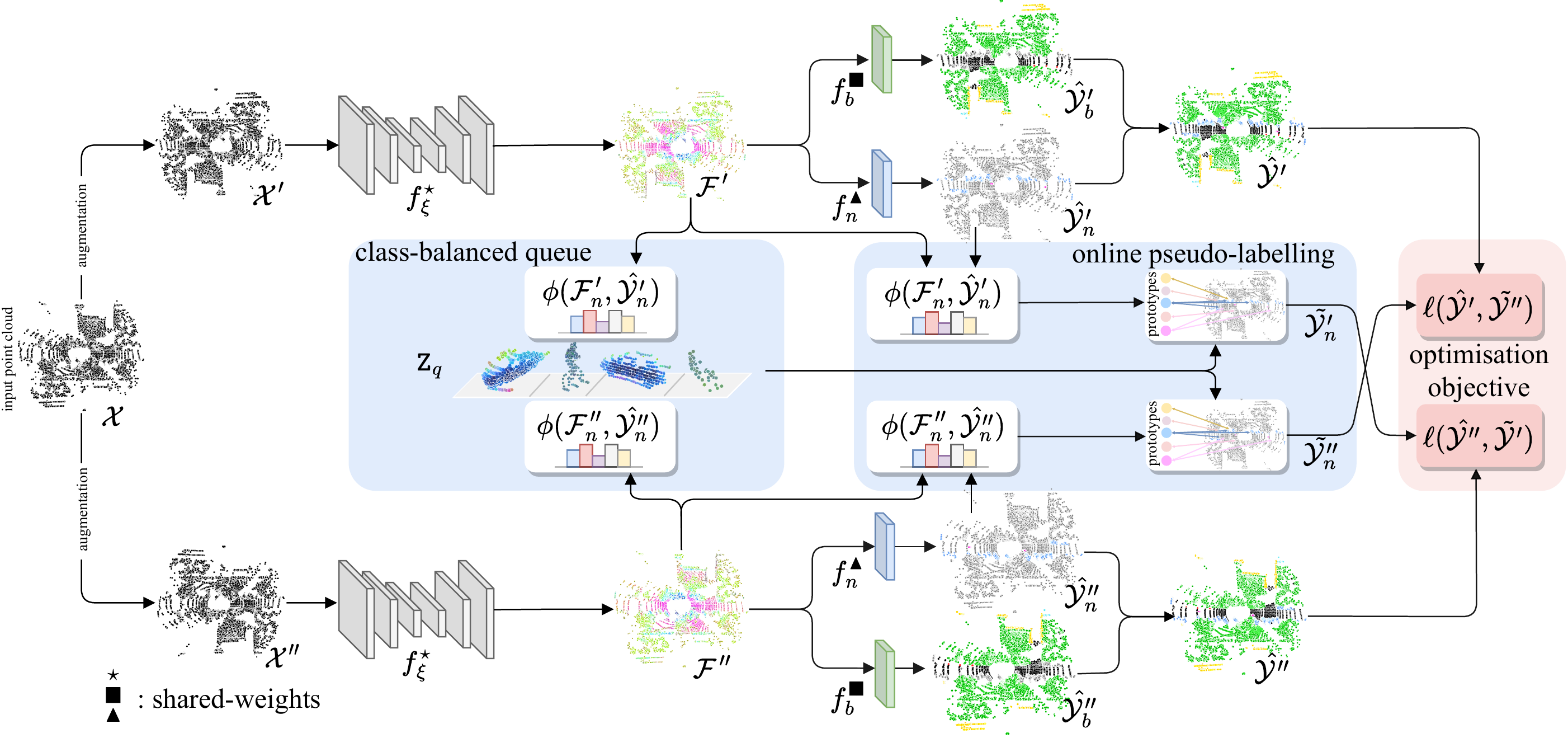}
    \vspace{-.6cm}
    \caption{Overview of \ourmethod.
    We random augment the input point cloud twice and extract point-level features $\mathcal{F}$ with the shared model $f_\xi$. 
    $\mathcal{F}$ are used to obtain pseudo-labels in the online pseudo-labelling. 
    We forward $\mathcal{F}$ to a novel $f_n$ and a base $f_b$ segmentation layer to output the novel and base predictions, respectively. We optimise our network by minimising a global objective function based on cross entropy.}
    \label{fig:main_chart}
\end{figure*}

\noindent \textbf{Point cloud semantic segmentation} can be performed at the point level~\cite{qi2017pointnet++}, on range view maps~\cite{ronneberger2015u}, and by voxelising the input points~\cite{zhou2018voxelnet}. 
Point-level networks process the input without intermediate representations. Examples of these include PointNet~\cite{qi2017pointnet}, PointNet++~\cite{qi2017pointnet++}, RandLA-Net~\cite{hu2020randla}, and KPConv~\cite{thomas2019kpconv}.
PointNet~\cite{qi2017pointnet} and PointNet++~\cite{qi2017pointnet++} are based on a series of multi-layer perceptron where PointNet++ introduces global and local feature aggregation at multiple scales. 
RandLA-Net~\cite{hu2020randla} uses random sampling, attentive pooling, and local spatial encoding. 
KPConv~\cite{thomas2019kpconv} employs flexible and deformable convolutions in a continuous input space. 
Point-level networks are computationally inefficient when large-scale point clouds are processed. 
Range view architectures~\cite{milioto2019rangenet++} and voxel-based approaches~\cite{choy20194d} are more computationally efficient than their point-level counterpart. 
The former requires projecting the input points on a 2D dense map, processing input maps with 2D convolutional filters~\cite{ronneberger2015u}, and re-projecting predictions to the initial 3D space. 
SqueezeSeg networks~\cite{wu2018squeezeseg, wu2019squeezesegv2}, 3D-MiniNet~\cite{alonso2020MiniNet3D}, RangeNet++~\cite{milioto2019rangenet++}, and PolarNet~\cite{zhang2020polarnet} are examples of this category. 
Although they are more efficient, these approaches tend to lose information during the projection and re-projection phase.
The latter includes 3D quantisation-based approaches that discretise the input points into a 3D voxel grid and employ 3D convolutions~\cite{zhou2018voxelnet} or 3D sparse convolutions~\cite{SubmanifoldSparseConvNet, choy20194d} to predict per-voxel classes. VoxelNet~\cite{zhou2018voxelnet}, SparseConv~\cite{SubmanifoldSparseConvNet, 3DSemanticSegmentationWithSubmanifoldSparseConvNet}, MinkowskiNet~\cite{choy20194d}, Cylinder3D~\cite{zhu2021cylindrical}, and (AF)$^2$-S3Net~\cite{ran2021af2s3net} are architectures belonging to this category. These approaches tackle point cloud segmentation in the supervised settings, whereas we tackle novel class discovery with labelled base classes and unlabelled novel classes.

\noindent \textbf{Novel class discovery} (NCD) is explored for 2D classification~\cite{han2019learning, zhong2021neighborhood, fini2021unified, joseph2022novel, roy2022class, jia2021joint, zhong2021openmix, vaze2022generalized, yang2022divide} and 2D segmentation~\cite{zhao2022novel}. 
NCD is more complex than standard semi-supervised learning~\cite{souly2017semi, zhang2020wcp, tang2016large}. In semi-supervised learning, labelled and unlabelled samples belong to the same classes, while in NCD, novel and base samples belong to disjoint classes. 
Han et al.~\cite{han2019learning} pioneered the NCD problem for 2D image classification. 
A classification model is pre-trained on a set of base classes and used as feature extractor for the novel classes. 
They then train a classifier for the novel classes using the pseudo-labels produced by the pre-trained model. 
Zhong et al.~\cite{zhong2021neighborhood} introduced neighbourhood contrastive learning to generate discriminative representations for clustering. 
They retrieve and aggregate pseudo-positive pairs with contrastive learning, encouraging the model to learn more discriminative representations. 
Hard negatives are obtained by mixing labelled and unlabelled samples in the feature space.
UNO~\cite{fini2021unified} unifies the two previous works by using a unique classification loss function for both base and novel classes, where pseudo-labels are processed together with ground-truth labels. 
NCD without Forgetting~\cite{joseph2022novel} and FRoST~\cite{roy2022class} further extend NCD to the incremental learning setting. 
EUMS~\cite{zhao2022novel} is the only approach analysing the NCD problem for semantic segmentation. Unlike image classification, the model has to classify each pixel and handle multiple classes in each image. EUMS consists of a multi-stage pipeline using a saliency model to cluster the latent representations of novel classes to produce pseudo-labels. Moreover, entropy-based uncertainty and self-training are used to overcome noisy pseudo-labels while improving the model performance on the novel classes. In this work, we tackle the problem of NCD in 3D point cloud semantic segmentation. Unlike previous works, our problem inherits the challenges from the fields of 2D semantic segmentation~\cite{deeplabv3plus2018, chen2017deeplab} and 3D point cloud segmentation~\cite{choy20194d, saltori2022cosmix, milioto2019rangenet++}. From 2D semantic segmentation, it inherits the additional challenges of multiple novel classes in the same image and the strong class unbalance. From 3D point cloud segmentation, we inherit the sparsity of input data, the different density of point cloud regions and the inability to identify foreground and background. The latter are not present in 2D segmentation~\cite{zhao2022novel}. 
Unlike \cite{zhao2022novel} that use K-Means, we formulate clustering as an optimal transport problem to avoid degenerate solutions (i.e.~all data points may be assigned to the same label and learn a constant representation)~\cite{Asano2020, mei2022data}.
Lastly, related to EUMS, REAL is proposed for open-world 3D semantic segmentation~\cite{cen2022open}, where both known and unknown points have to be segmented.
Unlike NOPS, all the unknown points belong to a single class and it is the task of a human annotator to separately label the novel classes.
Then, these labels are used to update the base model by incrementally learning the novel classes.


\section{Proposed approach}

\subsection{Overview}

Given an input point cloud, we produce two augmented views that are processed with the same deep neural network to extract point-level features.
These features are used to obtain pseudo-labels in the \textit{online pseudo-labelling} step through the Sinkhorn-Knopp algorithm~\cite{cuturi2013sinkhorn} (Sec.~\ref{sec:pseudo_labelling}).
Concurrently, we process the same features with the last network layers to segment novel and base classes.
These features are stored in the \textit{class-balanced queue} to mitigate the problem of batches with missing classes (Sec.~\ref{sec:class_balanced_queue}).
We exploit pseudo-label values (class probabilities) to filter out uncertain points, thus adding to the queue only high-quality points (Sec.~\ref{sec:unc_train}).
Lastly, we train our network by minimising the \textit{optimisation objective} function through a swapped prediction task based on the computed pseudo-labels (Sec.~\ref{sec:optimization_objective}).
Fig.~\ref{fig:main_chart} shows the block diagram of NOPS.

\subsection{Problem formulation}\label{sec:problem_formulation}

Let $\mathrm{X} = \{\mathcal{X}\}$ be a dataset of 3D point clouds captured in different scenes.
$\mathcal{X}$ is a set composed of a base set $\mathcal{X}_b$ and a novel set $\mathcal{X}_n$, s.t.~$\mathcal{X} = \mathcal{X}_b \cup \mathcal{X}_n$.
The semantic categories that can be present in our point clouds are $\mathcal{C} = \mathcal{C}_b \cup \mathcal{C}_n$, where $\mathcal{C}_b$ is the set of base classes and $\mathcal{C}_n$ is the set of novel classes, s.t.~$\mathcal{C}_b \cap \mathcal{C}_n = \emptyset$.
Each $\mathcal{X} \in \mathrm{X}$ is composed of a finite but unknown number of 3D points $\mathcal{X} = \{(\mathbf{x}, c)\}$, where $\mathbf{x} \in \mathbb{R}^3$ is the coordinate of the a point and $c$ is its semantic class.
We know the class of the point $(\mathbf{x}, c)$, s.t.~$\mathbf{x} \in \mathcal{X}_b$ and $c \in \mathcal{C}_b$, but we do not know the class of the point $(\mathbf{x}, c)$, s.t.~$x \in \mathcal{X}_n$ and $c \in \mathcal{C}_n$.
No points in $\mathcal{X}_n$ belong to one of the base classes $\mathcal{C}_b$.
As in \cite{han2019learning, zhong2021neighborhood, zhao2022novel}, we assume that the number of classes to discover is known, i.e.~$|\mathcal{C}_n| = C_n$.
We aim to design a computational approach that trains a deep neural network $f_{\mathbf{\Theta}}$ that can segment all the points of a given point cloud, thus learning to jointly segment base classes $\mathcal{C}_b$ and novel classes $\mathcal{C}_n$.
$\mathbf{\Theta}$ are the weights of our deep neural network.
$f_\mathbf{\Theta}$ is composed of two heads, $f_\mathbf{\Theta} = f_\xi \circ \{f_b, f_n\}$, where $f_b$ is the segmentation head for the base classes, $f_n$ is the segmentation head for the novel classes, $f_\xi$ is the feature extractor network and $\circ$ is the composition operator (Fig.~\ref{fig:main_chart}).

\subsection{Online pseudo-labelling}\label{sec:pseudo_labelling}

We formulate pseudo-labelling as the assignment of novel points to the class-prototypes  learnt during training~\cite{caron2020unsupervised}.
Let $\mathtt{P} \in \mathbb{R}^{D \times \rho}$ be the class prototypes, where $D$ is the size of the output features from $f_\xi$ and $\rho$ is the number of prototypes.
Let $\mathtt{Z} \in \mathbb{R}^{D \times m}$ be the normalised output features extracted from $f_\xi$, where $m$ is the number of points of the point cloud.
$m$ it is not known a priori and it can differ across point clouds.
We aim to find the assignment $\mathtt{Q} \in \mathbb{R}^{\rho \times m}$ s.t.~all the points in the batch are equally partitioned across the $\rho$ prototypes.
This equipartition ensures that the feature representations of the points belonging to different novel classes are well-separated, thus preventing the case in which the novel class feature representations collapse into a unique solution. 
Caron et al.~\cite{caron2020unsupervised} employs an arbitrary large number of prototypes $\rho$ to effectively organise the feature space produced by $f_\xi$. 
They discard $\mathtt{P}$ after training. 
In contrast, we learn exactly $\rho = C_n$ class prototypes and propose to use $\mathtt{P}$ as the weights for our new class segmentation head $f_n$, which outputs the $C_n$ logits for the new classes. 
In order to optimise the assignment $\mathtt{Q}$, we maximise the similarity between the features of the new points and the learned prototypes as
\begin{equation}
    \label{eq:sk_problem}
    \max_{\mathtt{Q} \in \mathcal{Q}} \,\, \text{Tr}(\mathtt{Q}^\top \mathtt{P}^\top \mathtt{Z}) + \epsilon H(\mathtt{Q})  \rightarrow \mathtt{Q}^*,
\end{equation}
where $H$ is the entropy function, $\epsilon$ is the parameter that determines the smoothness of the assignment and $\mathtt{Q}^*$ is our sought solution. 
Asano et al.~\cite{Asano2020} enforce the equipartioning constraint by requiring $\mathtt{Q}$ to belong to a transportation polytope and perform this optimisation on the whole dataset at once (offline).
This operation with point cloud data is computationally impractical.
Therefore, we formulate the transportation polytope such that the optimisation is performed online, which consist of processing only the points within the batch being processed
\begin{equation}
    \mathcal{Q} = \left\{ \mathtt{Q} \in \mathbb{R}^{C_n \times m}_+ | \mathtt{Q} \mathtt{1}_m = \frac{1}{C_n} \mathtt{1}_{C_n}, \mathtt{Q}^\top \mathtt{1}_{C_n} = \frac{1}{m} \mathtt{1}_m \right\},
\end{equation}
where $\mathtt{1}_\star$ represents a vector of ones of dimension $\star$.
These constraints ensure that each class prototype is selected on average at least $m / C_n$ times in each batch. 
The solution $\mathtt{Q}^*$ can take the form of a normalised exponential matrix
\begin{equation}
    \mathtt{Q}^* = \text{diag}(\alpha) \exp \left( {\frac{\mathtt{P}^\top \mathtt{Z}}{\epsilon}} \right) \text{diag}(\beta),
\end{equation}
where $\alpha$ and $\beta$ are renormalization vectors that are computed iteratively with the Sinkhorn-Knopp algorithm \cite{cuturi2013sinkhorn, mei2023overlap}.
We then transpose the optimised soft assignment $\mathtt{Q}^* \in \mathbb{R}^{C_n \times m}_+$ to obtain the soft pseudo-labels for each of the $m$ novel points being processed within each batch.

We empirically found that training can be more effective if pseudo-labels are smoother in the first training epochs and peaked in the last training epochs.
Therefore, we introduce a linear decay of $\epsilon$ during training.

\noindent \textbf{Multi-headed segmentation:}
A single segmentation head may converge to a suboptimal feature space, thus producing suboptimal prototype solutions.
To further improve the segmentation quality, we use multiple novel class segmentation heads to optimise $f_\Theta$ based on different training solutions.
Different solutions increase the likelihood of producing a diverse partitioning of the feature space as they regularise with each other (they share the same backbone) \cite{ji2019invariant}.
In practise, we concatenate the logits of the base class segmentation head with the outputs of each novel class segmentation head and we separately evaluate their loss for each novel class segmentation head at training time.

We task our network to over-cluster novel points, using segmentation heads that output $o\cdot C_n$ logits, where $o$ is the over-clustering factor. 
Previous studies empirically showed that this is beneficial to learn more informative features \cite{caron2020unsupervised, fini2021unified, mei2022data, ji2019invariant}. We observed the same and concur that over-clustering can be useful for increasing expressivity of the feature representations. 
The over-clustering heads are then discarded at inference time.

\subsection{Class-balanced queuing}\label{sec:class_balanced_queue}

Soft pseudo-labelling described in Sec.~\ref{sec:pseudo_labelling} produces an equipartite matching between the novel points and the class centroids.
However, it is highly likely that batches are sampled with point clouds containing novel classes with different cardinalities when dealing with 3D data.
It is also likely that some scenes may contain only a subset of the novel classes. 
Therefore, enforcing the equipartitioning constraint in each batch of the dataset could affect the learning of less-frequent (long-tail) classes.
As a solution, we introduce a queue $\mathtt{Z}_q$ containing a randomly extracted portion of the features of the novel points from the previous iterations.
We use these additional data to mitigate the potential class imbalance that may occur during training. 
In practise, we compute $\mathtt{Z} \leftarrow \mathtt{Z} \oplus \mathtt{Z}_q$, where $\oplus$ is the concatenation operator, and execute the Sinkhorn-Knopp algorithm on this augmented version of $\mathtt{Z}$.
Then, we retain only the pseudo-labels for the first $m$ columns of $\mathtt{Q}^*$.

\subsection{Uncertainty-aware training and queuing}\label{sec:unc_train}

We propose to carefully select novel points for training $f_\Theta$ with fewer but more reliable pseudo-labels and to build a more effective queue $\mathtt{Z}_q$.
We perform this selection by applying a threshold to the class probabilities of the novel class pseudo-labels.
We found that seeking a fixed threshold for all the novel classes, that is also compatible with the variations of the class probabilities during training, is impractical.
Therefore, we employ an adaptive threshold based on the class probabilities within each batch.

Our selection strategy operates as follows.
Let $\tau_c$ be the adaptive threshold for the points of the novel class $c \in \mathcal{C}_n$.
Firstly, we extract the novel points that have the greatest class probability for the class $c$.
Secondly, we compute $\tau_c$ as the $p$-th percentile of the class probabilities of these novel points.
Lastly, we retain the novel points of class $c$ whose class probability is above the threshold $\tau_c$.
We define this selection strategy as the function
\begin{equation}
    \phi : (\mathcal{F}_n, \hat{\mathcal{Y}_n}) \times p \mapsto (\bar{\mathcal{F}}_n),
\end{equation}
where $\mathcal{F}_n$ is the set of feature vectors extracted from $f_\xi$ and $\hat{\mathcal{Y}}_n$ is the set of class probabilities predicted by the network for these points.
$\bar{\mathcal{F}}_n$ are both processed by the Sinkhorn-Knopp algorithm to generate our pseudo-labels and added to $\mathtt{Z}_q$ to make it more effective.

\subsection{Optimisation objective}\label{sec:optimization_objective}

We optimise $f_\Theta$ by using the weighted Cross Entropy objective based on the labels $\mathcal{Y}_b$ of the base samples and the pseudo-labels $\tilde{\mathcal{Y}}_n$ of the novel samples.
We formulate a swapped prediction task based on these pseudo-labels \cite{caron2020unsupervised}.
Secifically, we begin by generating two different augmentations of $\mathcal{X}$ that we define as $\mathcal{X}^\prime$ and $\mathcal{X}^{\prime\prime}$.
We use the known one-hot labels for $\mathcal{Y}_b$ and the predicted soft pseudo-labels for $\tilde{\mathcal{Y}}_n$.
We predict the novel pseudo-labels $\tilde{\mathcal{Y}}_n^\prime$ and $\tilde{\mathcal{Y}}_n^{\prime\prime}$ of the respective point clouds $\mathcal{X}^\prime$ and $\mathcal{X}^{\prime\prime}$ with our approach.
Then, we enforce prediction consistency between the swapped pseudo-labels of the two augmentations as
\begin{equation}
    \label{eq:swapped_pred_task}
    \mathcal{L}(\mathcal{X}) = \ell(\hat{\mathcal{Y}}^\prime, \tilde{\mathcal{Y}}^{\prime\prime}) + \ell(\hat{\mathcal{Y}}^{\prime\prime}, \tilde{\mathcal{Y}}^{\prime}),
\end{equation}
where $\hat{\mathcal{Y}}^\prime = \hat{\mathcal{Y}}^\prime_b \cup \hat{\mathcal{Y}}^\prime_n$ (same for $\hat{\mathcal{Y}}^{\prime\prime}$),
$\tilde{\mathcal{Y}}^\prime = \mathcal{Y}_b^\prime \cup \tilde{\mathcal{Y}}_n^\prime$ (same for $\tilde{\mathcal{Y}}^{\prime\prime}$)
and $\ell$ is the weighted Cross Entropy loss.
We use separate segmentation heads for base classes and novel classes.
The weights of the loss for the base classes are computed based on their occurrence frequency in the training set.
The weights of the loss for the novel classes are all set equally as their occurrence frequency in the dataset is unknown.

\section{Adapting NCD for 2D images to 3D} \label{sec:adaptation_Zhao}

Another contribution of this work is to adapt the method proposed by Zhao et al.~\cite{zhao2022novel} for NCD for 2D semantic segmentation (EUMS) to 3D data. 
Our empirical evaluation (see Sec.~\ref{sec:experiments}) shows that the transposition of EUMS to the 3D domain has some limitations.
In particular, as described in Sec.~\ref{sec:intro}, EUMS uses two assumptions: \textbf{I}) the novel classes belong to the foreground and \textbf{II}) each image can contain at most one novel class.
This allows EUMS to leverage a saliency detection model to produce a foreground mask and a segmentation model pre-trained on the base classes to determine which portion of the image is background.
The portion of the image that belongs to both the foreground mask and the background mask is where features are then pooled.
EUMS computes a feature representation for each image by average pooling the features of the pixels belonging the unknown portion.
The feature representations of all the images in the dataset are clustered with K-Means by using the number of classes to discover as the target number of clusters.
EUMS shows that overclustering and entropy-based modelling can be exploited to improve the results.
The affiliation of a point to its cluster is used to produce hard pseudo-labels that are in turn used along with the ground-truth labels to fine-tune the pre-trained model.

With 3D point clouds, there is no concept of foreground and background (in contrast with \textbf{I}). Our adaptation is designed to discover the classes of all the unlabelled points (in contrast with \textbf{II}).
Therefore, given the unlabelled points of each point cloud, we randomly extract a subset of these by setting a ratio (e.g.~30\%) with upper bound (e.g.~1K) on the number of points to select.
We compute and collect their features for all the point clouds in the dataset and apply K-Means on the whole set of features.
Note that this clustering step is computationally expensive, and we had to use High Performance Computing to execute it.
The subsampling of the points was necessary to fit the data in the RAM (see Sec.~\ref{sec:experiments} for a detailed analysis).
Once the cluster prototypes are computed, we produce the hard pseudo-labels.
To enrich the set of pseudo-labels, we propagate the pseudo-label of each point to its nearest neighbour in the coordinate space.
This allows us to expand the subset of pseudo-labelled randomly selected points.
We also implement the other steps of overclustering and entropy-based modelling to boost the results.
Lastly, we fine-tune our model with these pseudo-labels.
We name our transposition of EUMS as EUMS$^\dag$ and report its block diagram in the Supplementary Material.

\section{Experimental results}\label{sec:experiments}

\subsection{Experimental setup}

\noindent \textbf{Datasets.} 
We evaluate our approach on SemanticKITTI~\cite{behley2019semantickitti, geiger2012cvpr, behley2021ijrr} and SemanticPOSS~\cite{pan2020semanticposs}.
SemanticKITTI~\cite{behley2019semantickitti} consists of 43,552 point cloud acquisitions with point-level annotations of 19 semantic classes.
Based on the official benchmark guidelines~\cite{behley2019semantickitti}, we use sequence $08$ for validation and the other sequences for training.
SemanticPOSS~\cite{pan2020semanticposs} consists of 2,988 real-world point cloud acquisitions with point-level annotations of 13 semantic classes.
Based on the official benchmark guidelines~\cite{pan2020semanticposs}, we use sequence $03$ for validation and the other sequences for training.

\noindent \textbf{Experimental protocol for 3D NCD.}
Similarly to what proposed by \cite{zhao2022novel} in the 2D domain, we create different splits of each dataset to validate the NCD performance.
We create four splits for SemanticKITTI and SemanticPOSS.
We refer to these splits as SemanticKITTI-$n^i$ and SemanticPOSS-$n^i$, where $i$ indexes the split.
In each set, the novel classes and the base classes correspond to unlabelled and labelled points, respectively.
Tabs.~\ref{tab:KITTI_folds} \& \ref{tab:POSS_folds} detail the splits of our datasets.
These splits are selected based on their class distribution in the dataset and on the semantic relationship between novel and base classes, e.g.~in KITTI-$4^3$ the base class \textit{motorcycle} can be helpful to discover the novel class \textit{motorcyclist}.
We report additional details about the selection process in the Supplementary Material.

We quantify the performance by using the mean Intersection over Union (mIoU), which is defined as the average IoU across the considered classes \cite{behley2019semantickitti}.
We provide separate mIoU values for the base and novel classes.
We also report the overall mIoU computed across all the classes in the dataset for completeness.

\begin{table}[t]
    \centering
    \caption{SemanticKITTI splits, is defined as KITTI-$n^i$, where $n$ is the number of novel classes and $i$ is the split index.}
    \label{tab:KITTI_folds}
    \vspace{-.2cm}
    \resizebox{\columnwidth}{!}{
    \begin{tabular}{ll}
        \toprule
        Split & Novel Classes \\
        \midrule
        KITTI-$5^0$ & \textit{building}, \textit{road}, \textit{sidewalk}, \textit{terrain}, \textit{vegetation} \\
        KITTI-$5^1$ & \textit{car}, \textit{fence}, \textit{other-ground}, \textit{parking}, \textit{trunk} \\
        KITTI-$5^2$ & \textit{motorcycle}, \textit{other-vehicle}, \textit{pole}, \textit{traffic-sign}, \textit{truck} \\
        KITTI-$4^3$ & \textit{bicycle}, \textit{bicyclist}, \textit{motorcyclist}, \textit{person} \\
        \bottomrule
    \end{tabular}
    }
\end{table}
\begin{table}[t]
    \centering
    \caption{SemanticPOSS splits, defined as POSS-$n^i$, where $n$ is the number of novel classes and $i$ is the split index.}
    \label{tab:POSS_folds}
    \vspace{-.2cm}
    \resizebox{.7\columnwidth}{!}{
    \begin{tabular}{ll}
        \toprule
        Split & Novel Classes  \\
        \midrule
        POSS-$4^0$ & \textit{building}, \textit{car}, \textit{ground}, \textit{plants} \\
        POSS-$3^1$ & \textit{bike}, \textit{fence}, \textit{person} \\
        POSS-$3^2$ & \textit{pole}, \textit{traffic-sign}, \textit{trunk} \\
        POSS-$3^3$ & \textit{cone-stone}, \textit{rider}, \textit{trashcan} \\
        \bottomrule
    \end{tabular}
    }
\end{table}

\noindent \textbf{Implementation Details.}
We implement our network based on a MinkowskiUNet-34C network~\cite{choy20194d}.
Point-level features are extracted from the penultimate layer.
The segmentation heads are implemented as linear layers, producing output logits for each point in the batched point clouds.
We train our network for 10 epochs.
We use the SGD optimizer, with momentum 0.9 and weight decay 0.0001. 
Our learning rate scheduler consists of linear warm-up and cosine annealing, with $lr_{max} = 10^{-2}$ and $lr_{min} = 10^{-5}$. 
We train with batch size equal to 4. 
We employ 5 segmentation heads, that are used in synergy with an equal number of over-clustering heads, with $o = 3$.
In $\phi$, we set $p=0.5$ for SemanticKITTI-$n^i$ and $p=0.3$ for SemanticPOSS-$n^i$.
We adapted the implementation of the Sinkhorn-Knopp algorithm~\cite{cuturi2013sinkhorn} from the code provided by \cite{caron2020unsupervised}, with the introduction of the queue and an in-place normalisation steps.
Similarly to \cite{caron2020unsupervised}, we set $n_{sk\_iters}=3$, while we adopt a linear decay for $\epsilon$, with $\epsilon_{start}=0.3, \epsilon_{end}=0.05$.

\subsection{Quantitative analysis}

\noindent \textbf{Segmentation quality.}
Tabs.~\ref{tab:results_poss} \& \ref{tab:results_semantickitti} report the quantitative results on SemanticPOSS and SemanticKITTI, respectively.
We report the \textit{Full supervision} setting as our upper bound.

On SemanticPOSS, we outperform EUMS$^\dag$ on three out of four splits with an improvement of $18.3$ mIoU on POSS-$4^0$, $~9.0$ mIoU on POSS-$3^1$ and $0.6$ on POSS-$3^2$.
In these splits, \ourmethod shows a large improvement on all the novel classes, with the exception of the class \textit{fence} in POSS-$3^1$ and \textit{traffic-sign} in POSS-$3^3$.
Differently, we deem the lower performance in POSS-$3^3$ is due to the difficulty and scarce presence of these novel classes.
The advantage of EUMS$^\dag$ is the clustering on the whole dataset that enables a complete visibility of all the novel classes.
On average, \ourmethod achieves $21.40$ mIoU, improving over EUMS$^\dag$ of $6.5$ mIoU.

On SemanticKITTI, we outperform EUMS$^\dag$ on all the four splits, improving by $12.6$ mIoU on KITTI-$5^0$, $1.2$ mIoU on KITTI-$5^1$, $4.2$mIoU on KITTI-$5^2$ and $5.3$ mIoU on KITTI-$4^3$.
\ourmethod outperforms EUMS$^\dag$ by a large margin on the majority of the novel classes. 
Exceptions are the class \textit{sidewalk} in KITTI-$5^0$, \textit{car} in KITTI-$5^1$, \textit{motorcycle} in KITTI-$5^2$ and \textit{motorcyclist} in KITTI-$4^3$.
On average, \ourmethod achieves $22.84$ mIoU, improving over EUMS$^\dag$ of $5.8$ mIoU.
Interestingly, \ourmethod outperforms also the supervised upper bound on the class \textit{trunk} in KITTI-$5^1$.

\noindent \textbf{Computational time.}
\ourmethod outperforms EUMS$^\dag$ in terms of computational time.
Firstly, EUMS$^\dag$ requires a pre-training step and a fine-tuning step, i.e.~30 training epochs in total. 
Then, EUMS$^\dag$ requires a large amount of memory (up to 200 GB memory for KITTI-$5^0$) to store the data required for clustering, taking several hours (50 hrs) to complete the training procedure. 
Differently, \ourmethod achieves superior performance with 10 training epochs, by using less memory (10 GB max) and a lower computational time (up to 25 hrs for KITTI-$5^0$).
We run these tests using one GPU Tesla A40-48GB.

\begin{table*}[t]
    \centering
    \caption{Novel class discovery results on SemanticPOSS. 
    \ourmethod outperforms EUMS$^\dag$ on three out of four splits. 
    Full supervision: model trained with labels for base and novel classes. EUMS$^\dag$: baseline described in Sec.~\ref{sec:adaptation_Zhao}. Highlighted values are the novel classes in each split.}
    \vspace{-.2cm}
    \label{tab:results_poss}
    \tabcolsep 6pt
    \resizebox{\textwidth}{!}{%
    \begin{tabular}{l|l|ccccccccccccc|ccc}
        \toprule
        \multirow{2}{*}{\textbf{Split}} & \multirow{2}{*}{\textbf{Model}} & \multirow{2}{*}{\rotatebox{45}{\textbf{bike}}} & \multirow{2}{*}{\rotatebox{45}{\textbf{build.}}} & \multirow{2}{*}{\rotatebox{45}{\textbf{car}}} & \multirow{2}{*}{\rotatebox{45}{\textbf{cone.}}} & \multirow{2}{*}{\rotatebox{45}{\textbf{fence}}} & \multirow{2}{*}{\rotatebox{45}{\textbf{grou.}}} & \multirow{2}{*}{\rotatebox{45}{\textbf{pers.}}} & \multirow{2}{*}{\rotatebox{45}{\textbf{plants}}} & \multirow{2}{*}{\rotatebox{45}{\textbf{pole}}} & \multirow{2}{*}{\rotatebox{45}{\textbf{rider}}} & \multirow{2}{*}{\rotatebox{45}{\textbf{traf.}}} & \multirow{2}{*}{\rotatebox{45}{\textbf{trashc.}}} & \multirow{2}{*}{\rotatebox{45}{\textbf{trunk}}} &  \multicolumn{3}{c}{\textbf{mIoU}} \\
         &  &  &  &  &  &  &  &  &  &  &  &  &  &  & \textbf{Novel} & \textbf{Base} & \textbf{All}\\
        \midrule
        
         &  Full supervision & 43.20 & 71.30 & 33.00 & 32.50 & 44.60  & 78.50 & 61.80 & 73.90 & 30.90 & 54.70 & 26.70 & 11.00 & 19.30 & - & - & 44.72 \\
        \midrule
        
        \multirow{2}{*}{POSS-$4^0$} & EUMS\dag \cite{zhao2022novel} & 25.67 & \CC{novelcolor}3.98 & \CC{novelcolor}0.56 & 16.44 & 29.40 & \CC{novelcolor}36.76 & 43.84 & \CC{novelcolor}28.46 & 13.13 & 26.75 & 18.18 & 3.34 & 16.91 & \CC{novelcolor}17.44 & 21.52 & 20.26 \\
         & \ourmethod (Ours) & 35.47 & \CC{novelcolor}\textbf{30.35} & \CC{novelcolor}\textbf{1.24} & 13.52 & 24.13 & \CC{novelcolor}\textbf{69.14} & 44.70 & \CC{novelcolor}\textbf{42.07} & 19.19 & 47.65 & 24.44 & 8.17 & 21.82 & \CC{novelcolor}\textbf{35.70} & 26.57 & 29.38 \\
        \midrule
        
        \multirow{2}{*}{POSS-$3^1$} & EUMS\dag \cite{zhao2022novel} & \CC{novelcolor}15.17 & 67.98 & 28.02 & 23.98 & \CC{novelcolor}\textbf{11.88} & 75.07 & \CC{novelcolor}35.98 & 74.46 & 26.91 & 48.56 & 26.00 & 5.60 & 23.05 & \CC{novelcolor}21.01 & 39.96 & 35.59 \\
         & \ourmethod (Ours) & \CC{novelcolor}\textbf{29.35} & 71.35 & 28.70 & 12.21 & \CC{novelcolor}3.94 & 78.24 & \CC{novelcolor}\textbf{56.78} & 74.21 & 18.29 & 38.88 & 23.31 & 13.74 & 23.51 & \CC{novelcolor}\textbf{30.02} & 38.24 & 36.35 \\
        \midrule
        
        \multirow{2}{*}{POSS-$3^2$} & EUMS\dag \cite{zhao2022novel} & 40.14 & 69.45 & 27.67 & 13.50 & 34.86 & 76.03 & 54.66 & 75.59 & \CC{novelcolor}5.27 & 39.22 & \CC{novelcolor}\textbf{7.79} & 8.52 & \CC{novelcolor}11.85 & \CC{novelcolor}8.31 & 43.96 & 35.74 \\
         & \ourmethod (Ours) & 37.16 & 71.81 & 29.74 & 14.64 & 28.38 & 77.53 & 52.09 & 73.00 & \CC{novelcolor}\textbf{11.51} & 47.11 & \CC{novelcolor}0.54 & 10.20 & \CC{novelcolor}\textbf{14.79} & \CC{novelcolor}\textbf{8.95} & 44.17 & 36.04 \\
        \midrule
        
        \multirow{2}{*}{POSS-$3^3$} & EUMS\dag \cite{zhao2022novel} & 41.17 & 70.68 & 28.08 & \CC{novelcolor}\textbf{4.34} & 38.27 & 76.66 & 38.29 & 75.35 & 25.76 & \CC{novelcolor}\textbf{34.34} & 28.31 & \CC{novelcolor}0.36 & 24.40 & \CC{novelcolor}\textbf{13.01} & 44.70 & 37.38 \\
         & \ourmethod (Ours) & 38.55 & 70.36 & 30.91 & \CC{novelcolor}0.00 & 29.38 & 76.50 & 55.98 & 71.84 & 17.03 & \CC{novelcolor}31.87 & 26.15 & \CC{novelcolor}\textbf{0.95} & 22.57 & \CC{novelcolor}10.94 & 43.93 & 36.32 \\
        \bottomrule
        \addlinespace[2.5pt]
         \multicolumn{12}{c}{} & \multirow{2}{*}{Avg} & \multicolumn{2}{|l|}{EUMS\dag \cite{zhao2022novel}} & \CC{novelcolor}14.94  & 37.54 & 32.24 \\
         \multicolumn{12}{c}{} &  & \multicolumn{2}{|l|}{\ourmethod (Ours)} & \CC{novelcolor}\textbf{21.40}  & 38.23 & 34.52 \\
         \cmidrule[1pt]{13-18}
    \end{tabular}
    }
\end{table*}
\begin{table*}[t]
    \centering
    \caption{Novel class discovery results on SemanticKITTI.
    \ourmethod outperforms EUMS$^\dag$ on all four splits. 
    Full supervision: model trained with annotations for base and novel classes. EUMS$^\dag$: baseline described in Sec.~\ref{sec:adaptation_Zhao}. Highlighted values are the novel classes in each split.}
    \vspace{-.2cm}
    \label{tab:results_semantickitti}
    \tabcolsep 3pt
    \resizebox{\textwidth}{!}{%
    \begin{tabular}{l|l|ccccccccccccccccccc|ccc}
        \toprule
        \multirow{2}{*}{\textbf{Split}} & \multirow{2}{*}{\textbf{Model}} & \multirow{2}{*}{\rotatebox{45}{\textbf{bi.cle}}} & \multirow{2}{*}{\rotatebox{45}{\textbf{b.clst}}} & \multirow{2}{*}{\rotatebox{45}{\textbf{build.}}} & \multirow{2}{*}{\rotatebox{45}{\textbf{car}}} & \multirow{2}{*}{\rotatebox{45}{\textbf{fence}}} & \multirow{2}{*}{\rotatebox{45}{\textbf{mt.cle}}} & \multirow{2}{*}{\rotatebox{45}{\textbf{m.clst}}} & \multirow{2}{*}{\rotatebox{45}{\textbf{oth-g.}}} & \multirow{2}{*}{\rotatebox{45}{\textbf{oth-v.}}} & \multirow{2}{*}{\rotatebox{45}{\textbf{park.}}} & \multirow{2}{*}{\rotatebox{45}{\textbf{pers.}}} & \multirow{2}{*}{\rotatebox{45}{\textbf{pole}}} & \multirow{2}{*}{\rotatebox{45}{\textbf{road}}} & \multirow{2}{*}{\rotatebox{45}{\textbf{sidew.}}} & \multirow{2}{*}{\rotatebox{45}{\textbf{terra.}}} & \multirow{2}{*}{\rotatebox{45}{\textbf{traff.}}} & \multirow{2}{*}{\rotatebox{45}{\textbf{truck}}} & \multirow{2}{*}{\rotatebox{45}{\textbf{trunk}}} & \multirow{2}{*}{\rotatebox{45}{\textbf{veget.}}} & \multicolumn{3}{c}{\textbf{mIoU}}\\
         &  &  &  &  &  &  &  &  &  &  &  &  &  &  &  &  &  &  &  &  & \textbf{Novel} & \textbf{Base} & \textbf{All}\\
        \midrule
        
         & Full supervision & 6.30 & 39.50 & 85.40 & 90.00 & 23.20 & 20.30 & 5.70 & 3.90 & 18.00 & 28.90 & 31.00 & 40.60 & 90.90 & 74.60 & 62.10 & 20.50 & 62.90 & 46.20 & 83.90 & - & - & 43.89 \\
        \midrule
        
        \multirow{2}{*}{KITTI-$5^0$} & EUMS\dag \cite{zhao2022novel} & 5.28 & 39.96 & \CC{novelcolor}15.77 & 79.20 & 9.03 & 16.89 & 2.52 & 0.07 & 11.39 & 14.40 & 12.67 & 29.17 & \CC{novelcolor}42.58 & \CC{novelcolor}\textbf{26.10} & \CC{novelcolor}0.05 & 10.30 & 47.37 & 37.92 & \CC{novelcolor}38.35 & \CC{novelcolor}24.57 & 21.08 & 23.11 \\
          & \ourmethod (Ours) & 5.59 & 47.76 & \CC{novelcolor}\textbf{52.68} & 82.60 & 13.76 & 25.55 & 1.36 & 1.66 & 14.52 & 19.80 & 25.86 & 32.12 & \CC{novelcolor}\textbf{56.74} & \CC{novelcolor}8.08 & \CC{novelcolor}\textbf{23.84} & 14.28 & 49.41 & 36.18 & \CC{novelcolor}\textbf{44.17} & \CC{novelcolor}\textbf{37.10} & 24.70 & 29.62 \\
        \midrule
        
        \multirow{2}{*}{KITTI-$5^1$} & EUMS\dag \cite{zhao2022novel} & 7.53 & 42.41 & 79.97 & \CC{novelcolor}\textbf{76.77} & \CC{novelcolor}\textbf{8.62} & 19.58 & 1.39 & \CC{novelcolor}\textbf{0.57} & 12.03 & \CC{novelcolor}14.14 & 13.95 & 40.74 & 86.32 & 66.45 & 56.29 & 11.97 & 44.79 & \CC{novelcolor}20.94 & 72.40 & \CC{novelcolor}24.21 & 37.06 & 35.62 \\
         & \ourmethod (Ours) & 7.36 & 51.23 & 84.53 & \CC{novelcolor}50.87 & \CC{novelcolor}7.27 & 28.93 & 1.76 & \CC{novelcolor}0.00 & 22.20 & \CC{novelcolor}\textbf{19.39} & 30.42 & 37.61 & 90.07 & 72.18 & 60.75 & 16.78 & 57.34 & \CC{novelcolor}\textbf{49.25} & 85.12 & \CC{novelcolor}\textbf{25.36} & 43.09 & 40.69 \\
        \midrule
        
        \multirow{2}{*}{KITTI-$5^2$} & EUMS\dag \cite{zhao2022novel} & 8.26 & 50.78 & 82.98 & 88.05 & 17.88 & \CC{novelcolor}\textbf{2.75} & 2.32 & 0.17 & \CC{novelcolor}3.16 & 25.40 & 24.98 & \CC{novelcolor}20.20 & 88.30 & 71.04 & 57.85 & \CC{novelcolor}8.63 & \CC{novelcolor}\textbf{27.16} & 38.36 & 76.95 & \CC{novelcolor}12.38 & 42.22 & 36.59 \\
         & \ourmethod (Ours) & 6.72 & 49.24 & 86.36 & 90.79 & 23.68 & \CC{novelcolor}2.69 & 0.58 & 1.87 & \CC{novelcolor}\textbf{15.46} & 29.48 & 27.92 & \CC{novelcolor}\textbf{36.39} & 90.26 & 73.39 & 61.21 & \CC{novelcolor}\textbf{17.83} & \CC{novelcolor}10.32 & 46.16 & 84.29 & \CC{novelcolor}\textbf{16.54} & 44.80 & 39.72 \\
        \midrule
        
        \multirow{2}{*}{KITTI-$4^3$} & EUMS\dag \cite{zhao2022novel} & \CC{novelcolor}\textbf{3.95} & \CC{novelcolor}2.47 & 80.10 & 87.21 & 16.81 & 14.02 & \CC{novelcolor}\textbf{14.98} & 0.31 & 14.13 & 20.77 & \CC{novelcolor}6.80 & 37.59 & 86.79 & 66.50 & 55.26 & 16.20 & 40.62 & 38.37 & 76.15 & \CC{novelcolor}7.05 & 43.39 & 35.74 \\
         & \ourmethod (Ours) & \CC{novelcolor}2.32 & \CC{novelcolor}\textbf{27.83} & 86.04 & 89.89 & 23.06 & 24.47 & \CC{novelcolor}2.92 & 3.06 & 18.19 & 30.09 & \CC{novelcolor}\textbf{16.32} & 39.90 & 90.65 & 73.51 & 61.04 & 17.40 & 49.76 & 44.01 & 83.18 & \CC{novelcolor}\textbf{12.35} & 48.95 & 41.24 \\
        \bottomrule
        \addlinespace[2.5pt]
        \multicolumn{17}{c}{} & \multirow{2}{*}{Avg} & \multicolumn{3}{|l|}{EUMS\dag \cite{zhao2022novel}} & \CC{novelcolor}17.05  & 35.94 & 32.76 \\
        \multicolumn{16}{c}{} & & & \multicolumn{3}{|l|}{\ourmethod (Ours)} & \CC{novelcolor}\textbf{22.84}  & 40.38 & 37.73 \\
        \cmidrule[1pt]{18-24} 

    \end{tabular}
    }
\end{table*}

\subsection{Qualitative analysis}

\begin{figure*}[t]
\centering
    \setlength\tabcolsep{2.5pt}
    \begin{tabular}{ccccc}
    \raggedright
        \begin{overpic}[width=0.19\textwidth]{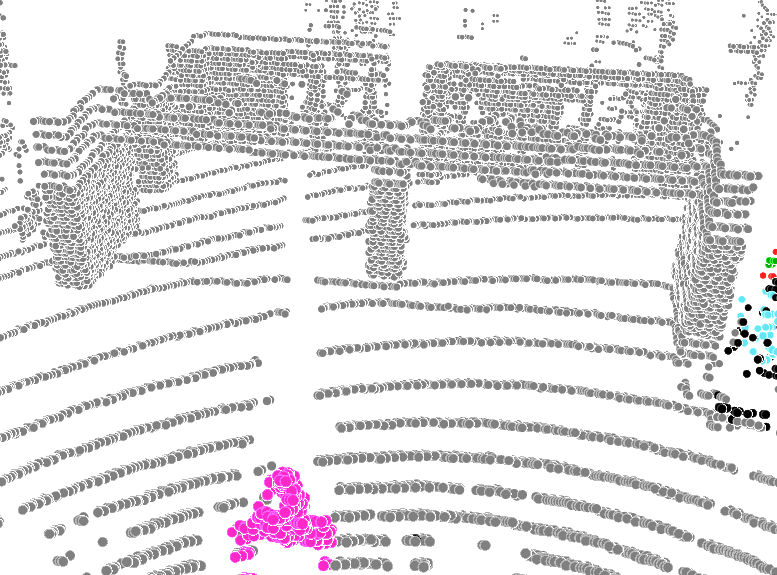}
        \put(20,75){\color{black}\footnotesize \textbf{EUMS$^\dag$~\cite{zhao2022novel} base}}
        \put(130,75){\color{black}\footnotesize \textbf{EUMS$^\dag$~\cite{zhao2022novel} novel}}
        \put(225,75){\color{black}\footnotesize \textbf{\ourmethod base (Ours)}}
        \put(330,75){\color{black}\footnotesize \textbf{\ourmethod novel (Ours)}}
        \put(460,75){\color{black}\footnotesize \textbf{GT}}
        \put(-10, 20){\small\rotatebox{90}{\color{black}\footnotesize \textbf{POSS-$3^1$}}}
        \end{overpic} &  
        \begin{overpic}[width=0.19\textwidth]{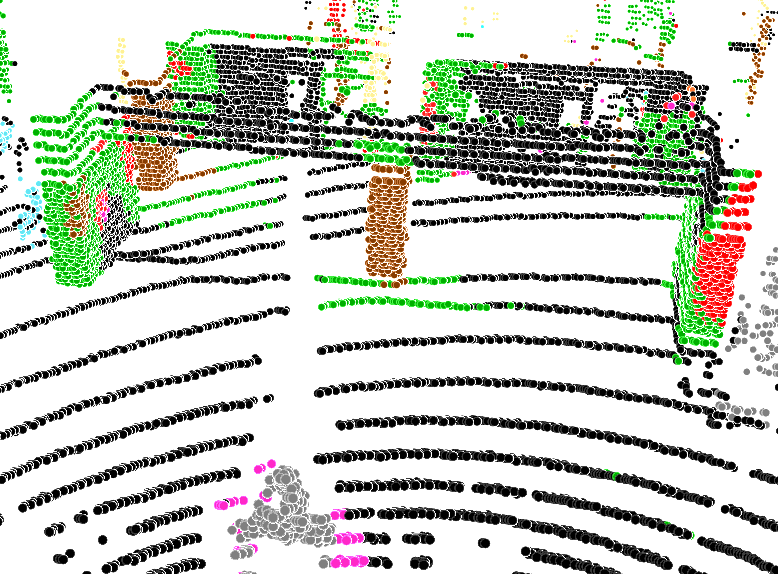}
        \end{overpic} &
        \begin{overpic}[width=0.19\textwidth]{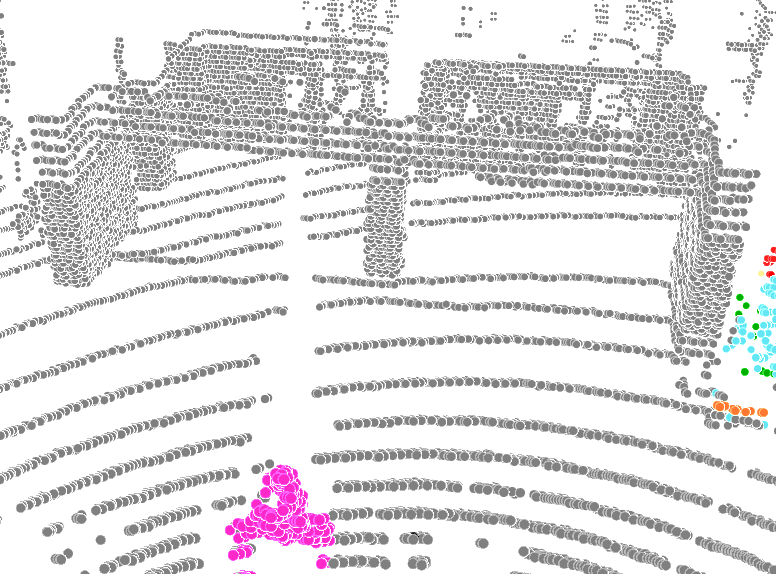}
        \end{overpic} &
        \begin{overpic}[width=0.19\textwidth]{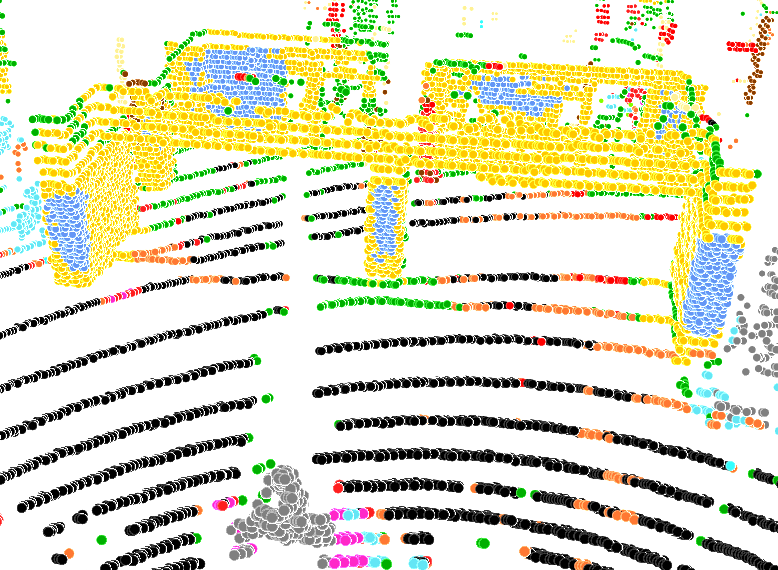}
        \end{overpic}
        \begin{overpic}[width=0.19\textwidth]{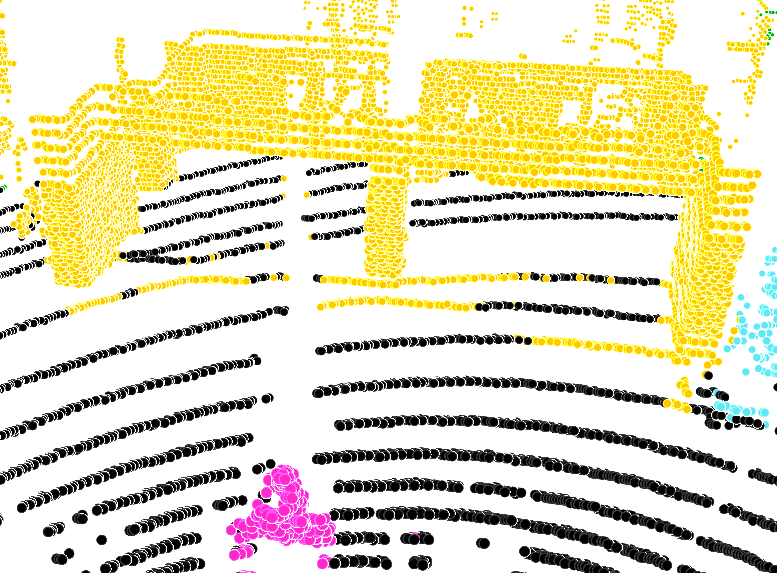}
        \end{overpic}
        \\
        \multicolumn{5}{c}{
        \begin{overpic}[width=0.99\textwidth]{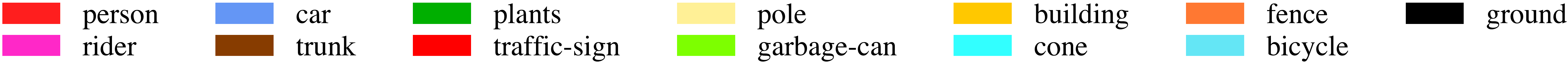}
        \end{overpic}}
        \\
        \begin{overpic}[width=0.19\textwidth]{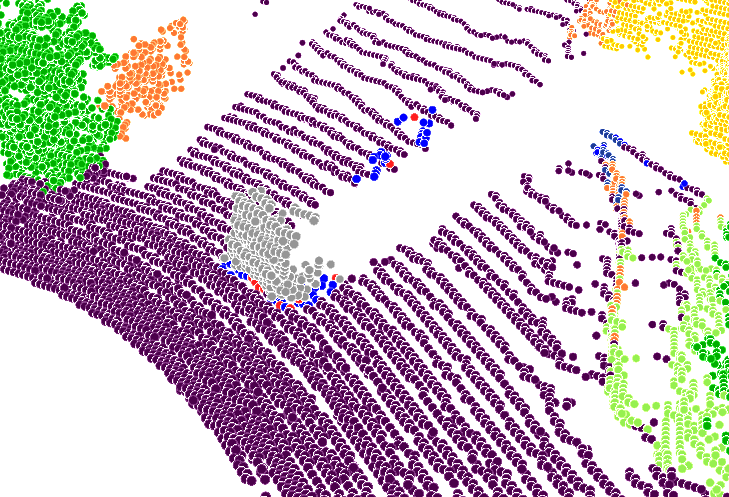}
        \put(-10, 20){\small\rotatebox{90}{\color{black}\footnotesize \textbf{KITTI-$5^2$}}}
        \end{overpic} &  
        \begin{overpic}[width=0.19\textwidth]{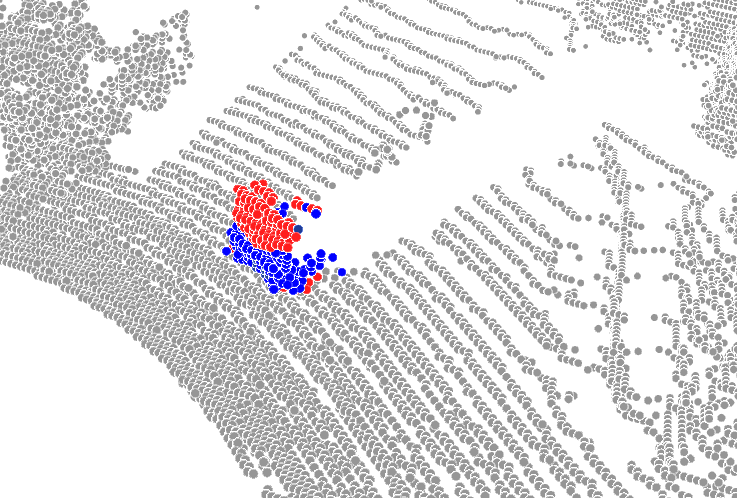}
        \end{overpic} &
        \begin{overpic}[width=0.19\textwidth]{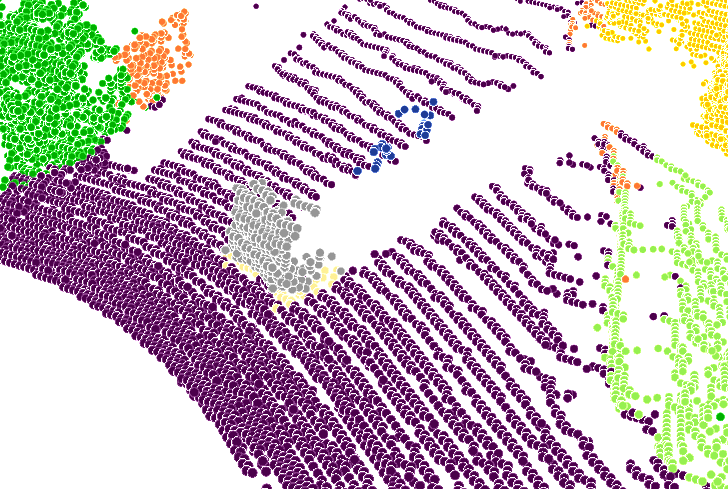}
        \end{overpic} &
        \begin{overpic}[width=0.19\textwidth]{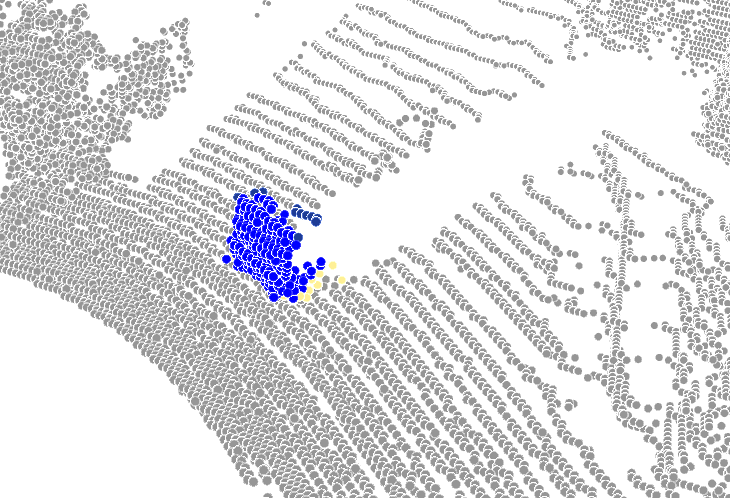}
        \end{overpic}
        \begin{overpic}[width=0.19\textwidth]{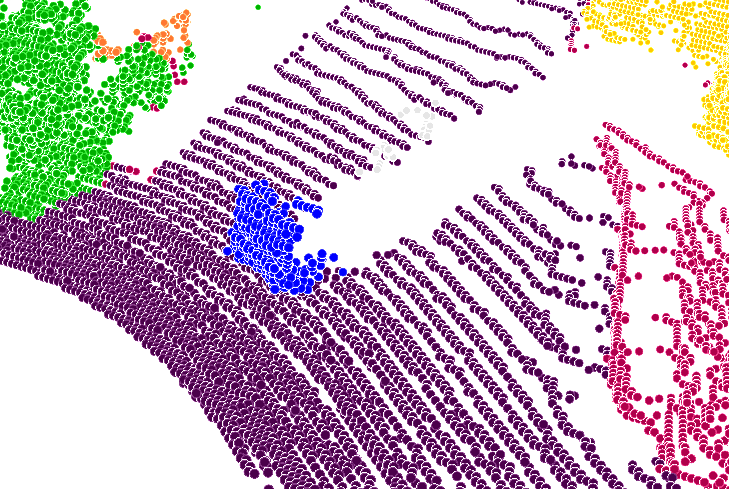}
        \end{overpic}
        \\
        \multicolumn{5}{c}{
        \begin{overpic}[width=0.99\textwidth]{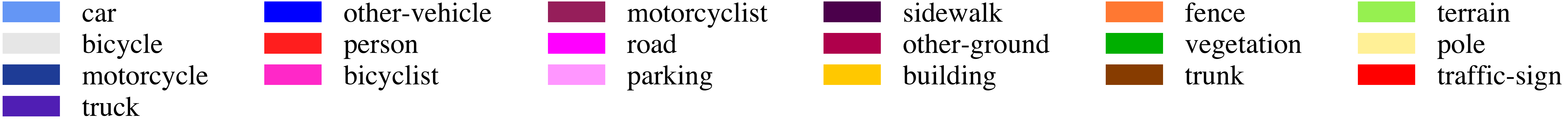}
        \end{overpic}}
    \end{tabular}
    \vspace{-.3cm}
    \caption{Qualitative comparisons on (top) SemanticPOSS and (bottom) SemanticKITTI. 
    We report results on both base and novel classes. 
    On novel classes, EUMS$^\dagger$ fails in recognising novel objects with mixed and cluttered predictions, e.g.~the facade of \textit{building} in POSS-$3^1$. 
    \ourmethod shows superior segmentation performance on all the novel classes.}
    \label{fig:qualitative}
\end{figure*}

Fig.~\ref{fig:qualitative} shows some segmentation results of \ourmethod and EUMS$^\dag$ on SemanticPOSS and SemaniticKITTI.
We can observe that the predictions of the base classes in the two datasets are correct for both the models, with just minor errors at the edges of the objects.
\ourmethod shows better segmentation capabilities also when dealing with the novel classes, being able to properly segment the correct class for the unknown objects. 
On the larger objects, such as buildings in the scenes, some mixing of labels can be observed.
Differently, EUMS nearly fails in correctly recognising the novel objects, resulting in the inclusion of different classes (either novel or base) into the same object, e.g.~the facade of the building in POSS-$3^1$.

\section{Ablation studies}\label{sec:ablation}
We study the main components of \ourmethod and its behaviour when varying its parameters on SemanticPOSS.

\begin{figure}[t]
\centering
    \begin{tabular}{@{}c@{}c}
        \begin{overpic}[width=0.48\columnwidth]{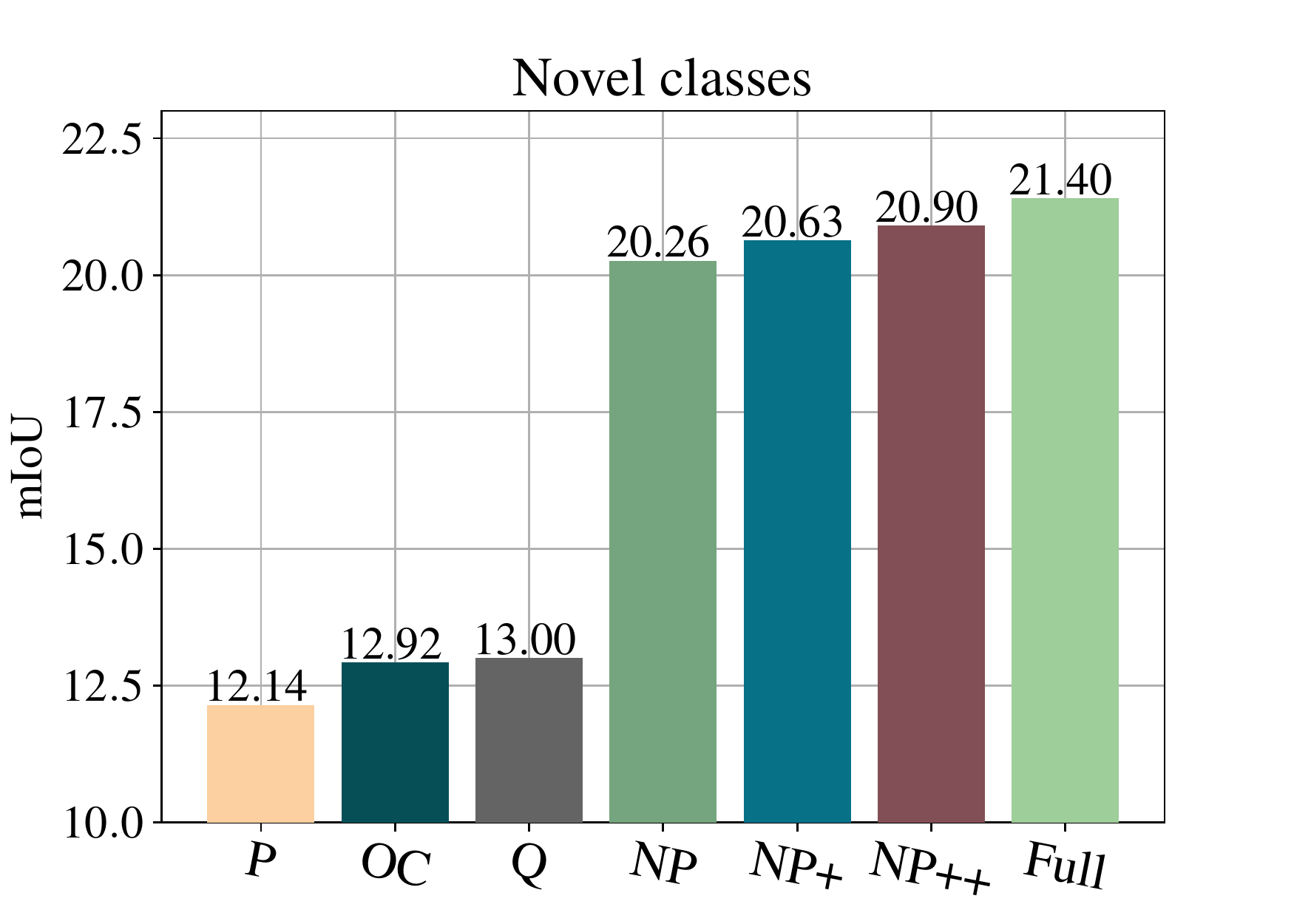}
        \end{overpic}& 
        \begin{overpic}[width=0.52\columnwidth]{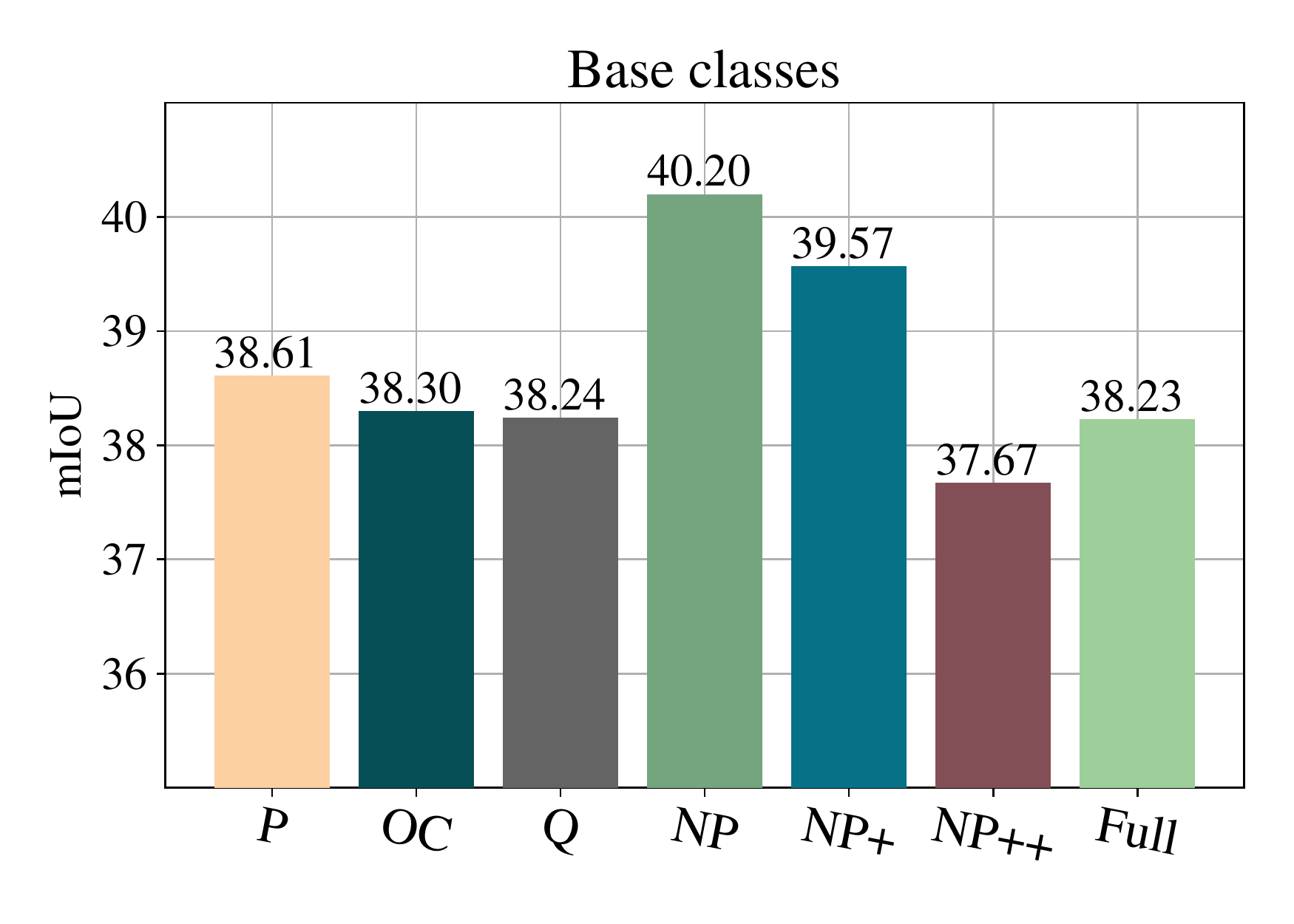}
        \end{overpic}
    \end{tabular}
    \vspace{-.3cm}
    \caption{Ablation study with different components and initialisation strategies on SemanticPOSS. 
    In $\mathsf{P}$, $\mathsf{OC}$ and $\mathsf{Q}$, we initialise the model after base pre-training, and use different configurations of the over-clustering heads and of our queue balancing. 
    In $\mathsf{NP}$ $\mathsf{NP}$+, $\mathsf{NP}$++ and $\mathsf{Full}$, we begin with $\mathsf{Q}$, we avoid pre-training, and we use $\phi$ and $\tau_c$ incrementally. 
    See Sec.~\ref{sec:ablation} for definition of methods.}
    \label{fig:ablation_components}
\end{figure}

\noindent\textbf{Method components.}
Fig.~\ref{fig:ablation_components} shows the performance on novel and base classes of seven versions of \ourmethod.
The first three versions use the pre-trained model on the base classes, while the last four versions use the model trained from scratch. 
Each version is defined as follows:

\begin{itemize}
    \item $\mathsf{P}$: we use a pre-trained model, and we remove $Z_q$, $\tau_c$ and the over-clustering heads.
    \item $\mathsf{OC}$: $\mathsf{P}$ + over-clustering heads.
    \item $\mathsf{Q}$: $\mathsf{OC}$ + $Z_q$, i.e.~our queue without class-balancing.
    \item $\mathsf{NP}$: $\mathsf{Q}$ without pre-training.
    \item $\mathsf{NP}$+: $\mathsf{NP}$ + our selection function $\phi$ on the queue.
    \item $\mathsf{NP}$++: $\mathsf{NP}$ + $\tau_c$ on the features used to derive the pseudo-labels.
    \item $\mathsf{Full}$: \ourmethod with all the components activated.
\end{itemize}

Pre-trained approaches generally underperform their trained-from-scratch counterparts on the novel classes.
This is visible in the low performance of $\mathsf{P}$, $\mathsf{OC}$ and $\mathsf{Q}$.
We have a significant improvement when pre-training is not used ($\mathsf{NP}$), i.e.~we achieve $20.26$ mIoU.
We can see that the queue both with and without pre-training is helpful.
When we add the feature selection for the queue and for the training, i.e.~$\mathsf{NP}$+ and $\mathsf{NP}$++, we have improvements, i.e.~$20.63$ mIoU and $20.90$ mIoU, respectively.
The best performance is achieved with $\mathsf{Full}$.
Although we can observe variations on the performance of the base classes, their information is retained by the network when we discover the novel ones.

\noindent \textbf{Parameter analysis.}
We study the behaviour of the percentile $p$ in our selection function $\phi$, when we apply it to the features both for pseudo-labelling and for the class-balanced queue $\mathtt{Z}_q$.
Tab.~\ref{tab:components_ablations} reports the results on each split of SemanticPOSS.
For each split, we can observe that the performance depends on the number of points and difficulty of the novel classes. In POSS$-4^0$ and POSS-$3^1$, lower values of $p$ result in less severe selection. 
We believe this is related to the class distribution within these splits.
This is in line with what observed in Tab.~\ref{tab:results_poss}.
In POSS$-3^2$ and POSS-$3^3$, we notice a different behaviour, a higher value of $p$ provides better results.
We relate this to the difficulty of the novel classes in these splits whose noisy pseudo-labels can benefit from a more rigorous selection of the features.

\begin{table}[t]
    \centering
    \caption{Ablation study showing how different values of $p$ affect the performance on SemanticPOSS. 
    The lower $p$ is, the less severe the selection of the features, resulting in better performances for POSS-$4^0$. Differently, POSS-$3^3$ benefits from an higher value of $p$, which leads to a more vigorous filtering of the features. POSS-$3^1$ and POSS-$3^2$ show the best performances with $p=0.5$}
    \label{tab:components_ablations}
    \vspace{-.2cm}
    \resizebox{.85\columnwidth}{!}{
    \begin{tabular}{lccccc}
        \toprule
        \multirow{2}{*}{Split} & \multicolumn{5}{c}{Percentile $p$}\\
        & 0.1 & 0.3 & 0.5 & 0.7 & 0.9 \\
        \midrule
        POSS-$4^0$ & 30.81 & \textbf{35.70} & 28.77 & 30.93 & 26.69 \\
        POSS-$3^1$ & 28.33 & 30.02 & \textbf{30.43} & 23.32 & 18.91 \\
        POSS-$3^2$ & 8.07 & 8.95 & \textbf{10.32} & 10.25 & 7.76 \\
        POSS-$3^3$ & 10.55 & 10.94 & 11.69 & \textbf{14.38} & 13.42 \\
        \midrule
        Avg. & 19.44 & \textbf{21.40} & 20.30 & 19.72 & 16.70 \\
        \bottomrule
    \end{tabular}
    }
\end{table}

\section{Conclusions}\label{sec:conclusions}

We explored the new problem of novel class discovery for 3D point cloud segmentation.
Firstly, we transposed the only NCD method for 2D semantic segmentation to 3D point cloud data, and experimentally found that it has several limitations.
We discussed that extending 2D NCD approaches to 3D data (point clouds) is not trivial because the assumptions made for 2D data are not easily transferable to 3D.
Secondly, we presented \ourmethod, to tackle NCD for point cloud segmentation by using online clustering and exploiting uncertainty quantification to produce pseudo-labels for the novel points.
Lastly, we introduced a new evaluation protocol to asses the performance of NCD for point cloud segmentation.
Experiments on two different segmentation dataset showed that \ourmethod outperforms the compared baselines by a large margin.
Future research directions could investigate the extension of our method when base annotations are fewer and/or weakly labelled.

\noindent \textbf{Limitations}
\ourmethod limitations include the prior knowledge on the number of novel classes $C_n$ to discover.
This could be a limitation when $C_n$ is not known a-priori and novel classes may appear in an incremental manner.
We believe that a solution may be to learn novel classes incrementally.
Another limitation is the loss we use to handle class unbalancing.
More recent techniques to handle this drawback could be further explored.


{\small
\bibliographystyle{ieee_fullname}
\bibliography{egbib}
}

\clearpage



\twocolumn[
\centering
\Large
\textbf{Novel Class Discovery for 3D Point Cloud Semantic Segmentation} \\
\vspace{0.5em}Supplementary Material \\
\vspace{1.0em}
] 
\appendix


\section{Introduction}

We provide some additional material in support of the main paper. 
The content is organised as follows:

\begin{itemize}
    \item Sec.~\ref{sec:dataset_splits} provides a description of the proposed datasets for the evaluation of NCD methods for point cloud semantic segmentation;
    \item Sec.~\ref{sec:eums_3D} reports the implementation details of EUMS$^\dag$, our adaptation for 3D point cloud data of the method proposed by Zhao et al.~\cite{zhao2022novel} (originally designed for NCD in 2D image semantic segmentation);
    \item Sec.~\ref{sec:supplementary_qualitative} shows a collection of additional qualitative results produced with \ourmethod on all the different splits of  SemanticPOSS-$n^i$ and SemanticKITTI-$n^i$.
\end{itemize}

\section{Dataset splits for 3D NCD} \label{sec:dataset_splits}

\begin{figure}[t]
\centering
    \begin{tabular}{c}
        \begin{overpic}[width=0.95\columnwidth]{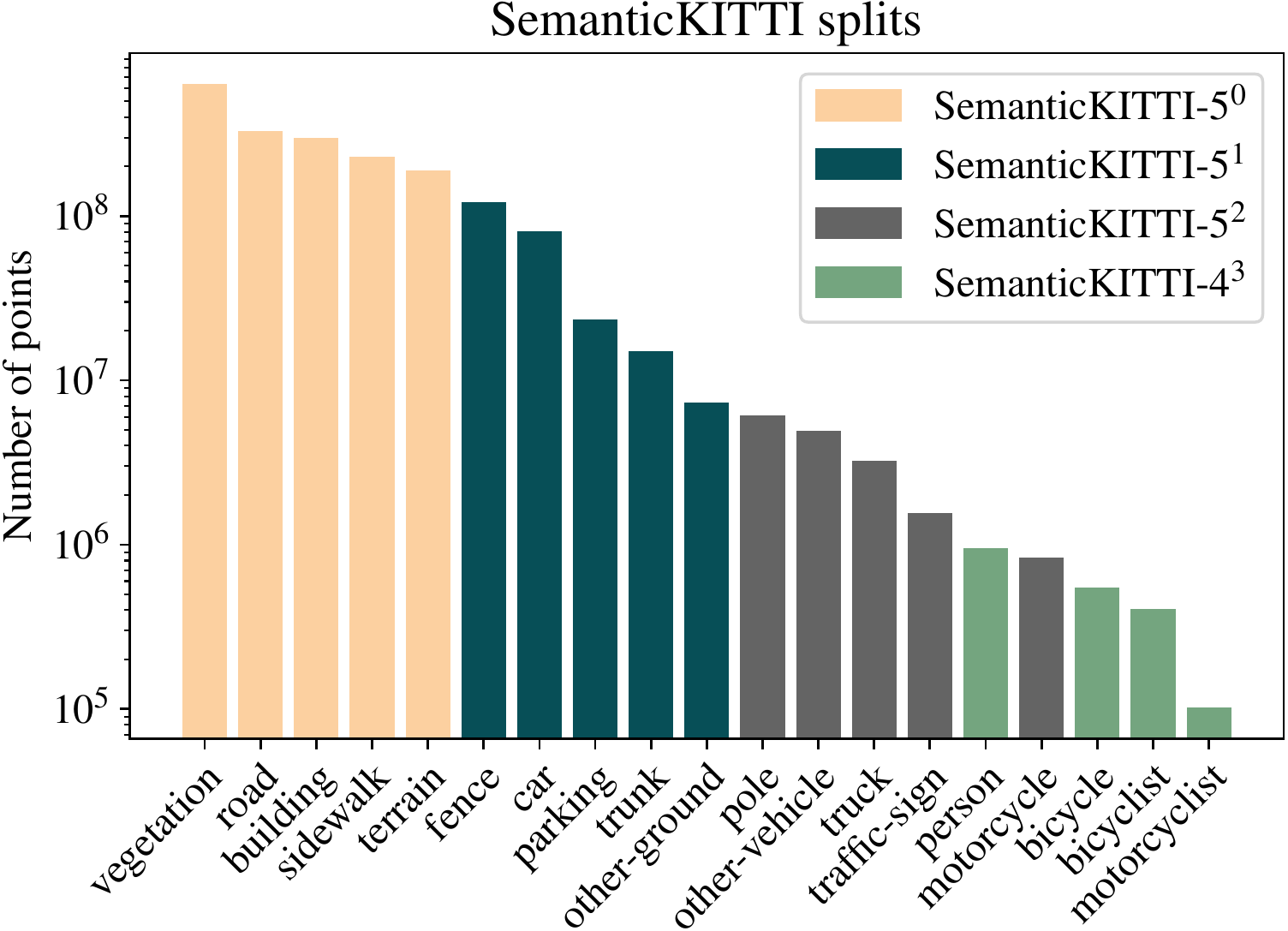}
        \end{overpic} \\
        \begin{overpic}[width=0.95\columnwidth]{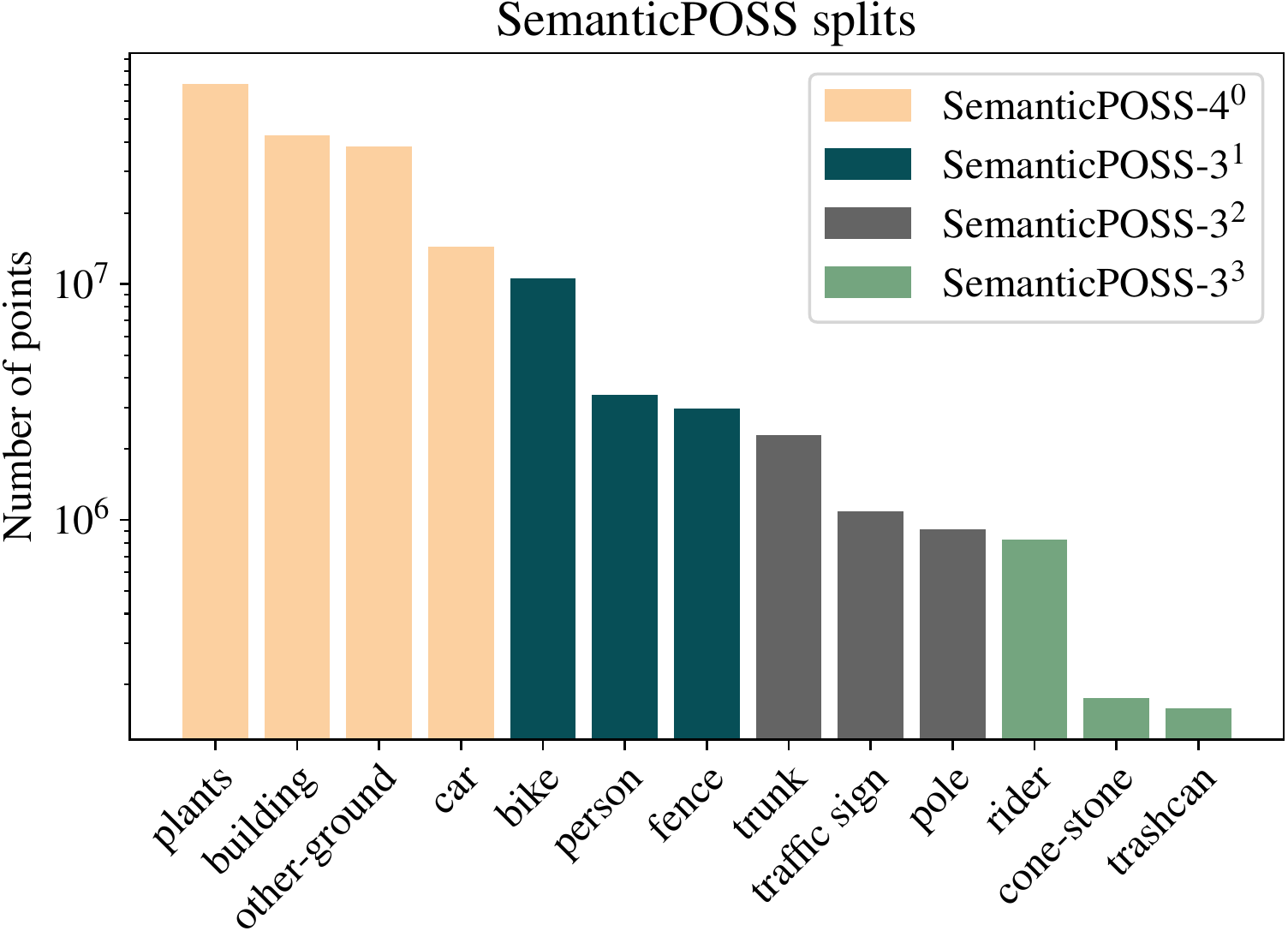}
        \end{overpic}\\
    \end{tabular}
    \vspace{-.3cm}
    \caption{Histograms representing the number of points belonging to each class in SemanticKITTI \cite{behley2019semantickitti} and SemanticPOSS \cite{pan2020semanticposs}. Each class has been assigned the colour of the split in which it has to be considered novel (unlabelled).}
    \label{fig:dataset_splits}
\end{figure}

To evaluate the performances of NOPS, we divide SemanticKITTI \cite{behley2019semantickitti} and SemanticPOSS \cite{pan2020semanticposs} into four different splits. We name these splits as SemanticKITTI-$n^i$ and SemanticPOSS-$n^i$, respectively, where $n$ is the number of novel classes contained in each split and $i$ indexes the split. 
In each set, the novel classes and the base classes correspond to unlabelled and labelled points.  
These splits are selected based on two principles, i.e.~balancing the distribution of the novel classes in each split, and including semantic relationships between base and novel classes within the same split.
See details about the splits in Fig.~\ref{fig:dataset_splits}.
The first principle allows us to avoid the case in which the most frequent novel class affects the other classes, thus in turn affecting the learning of the unsupervised points. 
The second principle encourages the model to exploit the supervised knowledge over some base classes to discover the unsupervised novel classes, as in the case of the novel class \textit{rider} in SemanticPOSS-$3^3$, whose discovery can be facilitated by the presence of the class \textit{bike}, that is considered as base class in this specific split.

\section{Adapting NCD for 2D images to 3D} \label{sec:eums_3D}

\begin{figure*}[t]
    \centering
    \includegraphics[width=\textwidth]{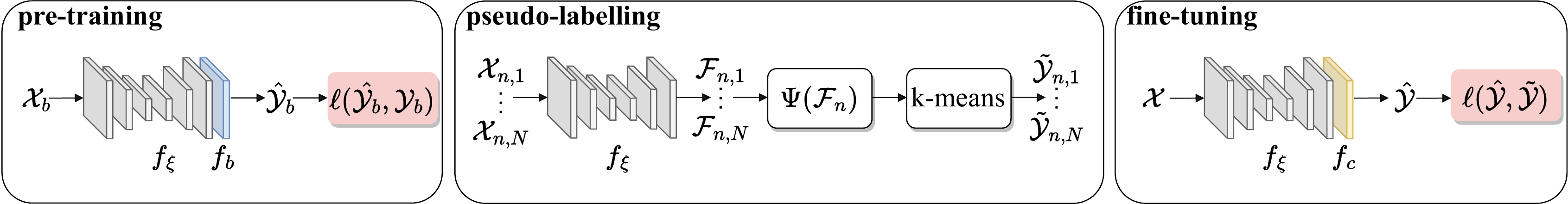}
    \caption{Overview of EUMS$^\dag$, our adaptation of the method proposed by Zhao et al.~\cite{zhao2022novel}. We first pre-train $f_\xi$ and $f_b$ considering only the base points in each point cloud. Using $f_\xi$, we extract the features of the novel points in each scene, that are filtered with the selection function $\Psi(\cdot)$. Then, we produce the pseudo-labels for the selected novel points by using the k-means algorithm. Lastly, we plug a new segmentation head $f_c$ into $f_\xi$ and fine-tune the complete model on both novel and base points, considering pseudo-labels and ground-truth labels respectively.}
    \label{fig:eums_chart}
\end{figure*}

One of our contribution is the adaptation of EUMS~\cite{zhao2022novel}, proposed for NCD in 2D image semantic segmentation, to 3D data. 
As some of the EUMS assumptions for the 2D case do not hold in the 3D point cloud domain, we introduce some changes in the proposed baseline. 
We name this adapted version as EUMS$^\dag$.

As in the original implementation, EUMS$^\dag$ consists of three consecutive steps: i) pre-training, ii) pseudo-labelling and iii) fine-tuning. 
The block diagram of EUMS$^\dag$ is illustrated in Fig.~\ref{fig:eums_chart}.
We first pre-train our model $f_\xi \circ f_b$ on the base classes only, where $f_\xi$ is the feature extractor, $f_b$ is the segmentation head for the base classes and $\circ$ is the composition operator.
Then, we generate the pseudo-labels considering the features extracted by $f_\xi$ and filtered with the selection function $\Psi(\cdot)$ working on the whole dataset, where $\Psi$ is a random selection function.
Lastly, we fine-tune the architecture $f_\xi \circ f_c$ jointly on the labelled base points and on the pseudo-labelled novel points, where $f_c$ is the segmentation head for both base and novel classes.
Here below each step is described in detail.

\noindent \textbf{Pre-training.} EUMS assumes that the novel classes belong to the foreground. Then, the novel classes are merged with the background class (considered as base in all the dataset splits) during the pre-training phase. The foreground pixels are obtained by an auxiliary saliency detection model~\cite{qin2019basnet}. The background pixels are just the output of the pre-trained model. 
The portion of the image belonging to both the foreground and the background masks contains the novel pixels.
Because in point clouds there is no concept of foreground/background and saliency for 3D data cannot be leveraged as easily as for 2D data \cite{Ran2021}, we consider the novel points as the unlabelled points and we discard them during the pre-training phase. 
Therefore, the pre-training stage of EUMS$^\dag$ considers only the base points in each scene $\mathcal{X}_b$ and optimises $f_\xi \circ f_b$ by considering the objective function $\ell(\hat{\mathcal{Y}}_b, \mathcal{Y}_b)$, where $\hat{\mathcal{Y}}_b$ are the network predictions $ \hat{\mathcal{Y}}_b = (f_\xi \circ f_b) (\mathcal{X}_b)$ and $\mathcal{Y}_b$ are the ground-truth annotations for the base points.

\noindent \textbf{Pseudo-labelling.} EUMS assumes that each image contains at most one novel class, allowing to compute a unique pseudo-label for each image. 
Authors in \cite{zhao2022novel} propose to first average pool the features of the novel pixels in each image and then collect the image-level representations for the whole dataset. Finally, the hard pseudo-labels for all the novel points in each image are obtained by propagating the clustering affiliation of each image-level feature vector, determined by using the k-means algorithm.

In semantic segmentation for 3D point clouds, multiple novel classes usually occur in the same scene. Therefore, in EUMS$^\dag$ we propose to extract the per-point features $\mathcal{F}_{n, i}$ with $f_\xi$ for all the novel points $\mathcal{X}_{n, i}$ contained in the $i$-th scene of the dataset. 
However, a large amount of novel points is difficult to handle due to hardware constrains.
We randomly select a subset of point-level vectors using $\Psi$ from each scene by setting a ratio (i.e.~$30\%$) with an upper bound (i.e.~1K) on the number of points to select. 
Finally, we apply k-means clustering on the set of features collected over the whole dataset and obtain the pseudo-labels $\tilde{\mathcal{Y}}_{n, i}$ for the selected novel points in $\mathcal{X}_{n, i}$. To further enrich the pseudo-labels, we propagate the pseudo-label of each novel point to its nearest neighbour in the coordinate space. This allows us to increase the number of pseudo-labelled novel points.

\noindent \textbf{Fine-tuning.} During the last step of the EUMS$^\dag$, we fine-tune the complete model following the same strategy used in \cite{zhao2022novel}: given a point cloud $\mathcal{X}$, we compute the class predictions $\hat{\mathcal{Y}}$ as  $\hat{\mathcal{Y}} = (f_\xi \circ f_c)(\mathcal{X})$ and we optimise the network considering the loss $\ell(\hat{\mathcal{Y}}, \tilde{\mathcal{Y}})$, where $\tilde{\mathcal{Y}} = \mathcal{Y}_b \cup \tilde{\mathcal{Y}}_n$.

\section{Qualitative results} \label{sec:supplementary_qualitative}

In this section, we report additional qualitative results by comparing \ourmethod with EUMS$^\dag$~\cite{zhao2022novel} predictions.

\begin{figure*}[t]
\centering
    \setlength\tabcolsep{2.5pt}
    \begin{tabular}{ccccc}
    \raggedright
        \begin{overpic}[width=0.19\textwidth]{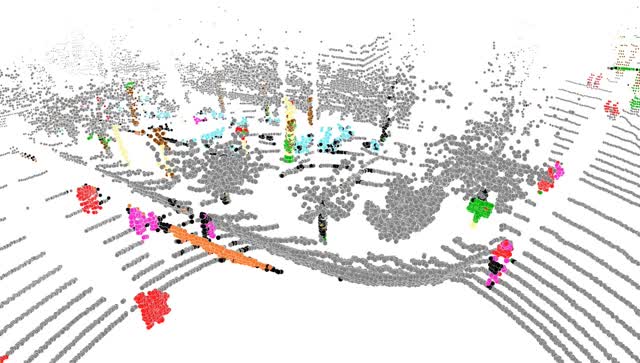}
        \put(20,60){\color{black}\footnotesize \textbf{EUMS$^\dag$~\cite{zhao2022novel} base}}
        \put(130,60){\color{black}\footnotesize \textbf{EUMS$^\dag$~\cite{zhao2022novel} novel}}
        \put(225,60){\color{black}\footnotesize \textbf{\ourmethod base (Ours)}}
        \put(330,60){\color{black}\footnotesize \textbf{\ourmethod novel (Ours)}}
        \put(460,60){\color{black}\footnotesize \textbf{GT}}
        \end{overpic} &  
        \begin{overpic}[width=0.19\textwidth]{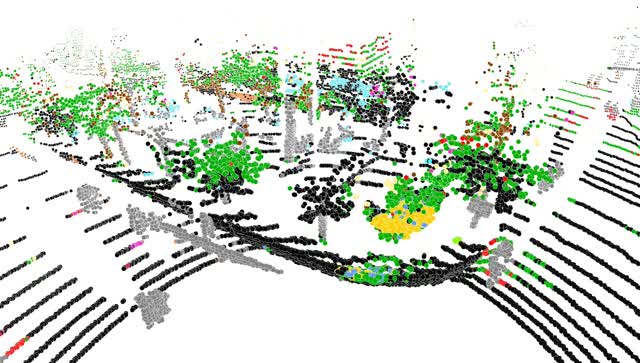}
        \end{overpic} &
        \begin{overpic}[width=0.19\textwidth]{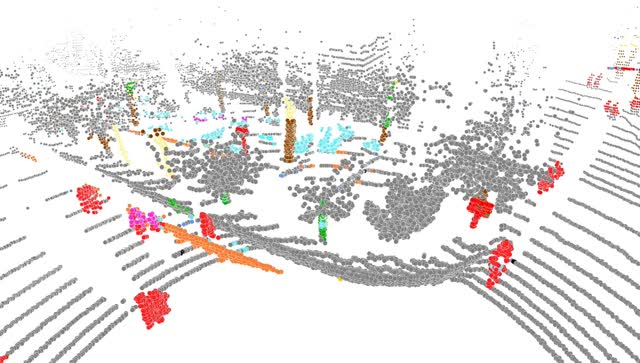}
        \end{overpic} &
        \begin{overpic}[width=0.19\textwidth]{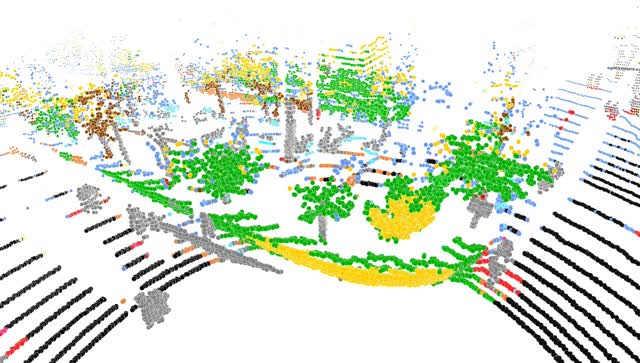}
        \end{overpic}
        \begin{overpic}[width=0.19\textwidth]{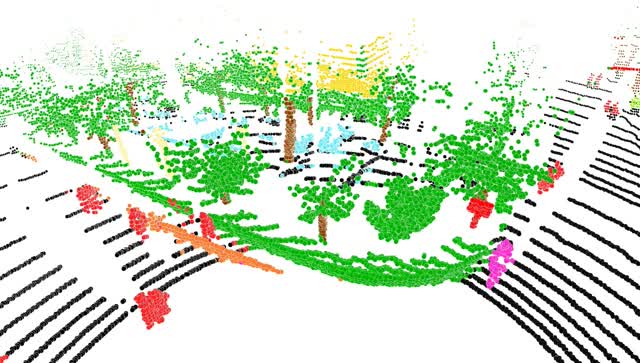}
        \end{overpic}
        \\
        \begin{overpic}[width=0.19\textwidth]{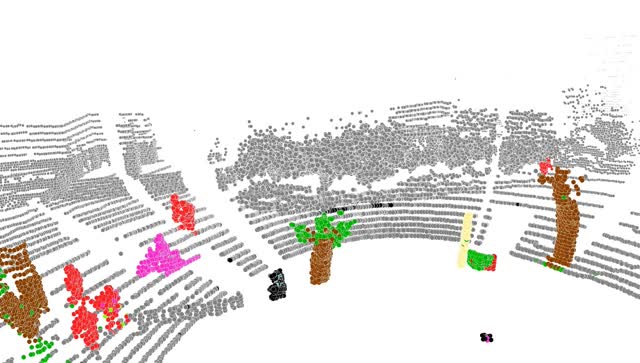}
        \end{overpic} &  
        \begin{overpic}[width=0.19\textwidth]{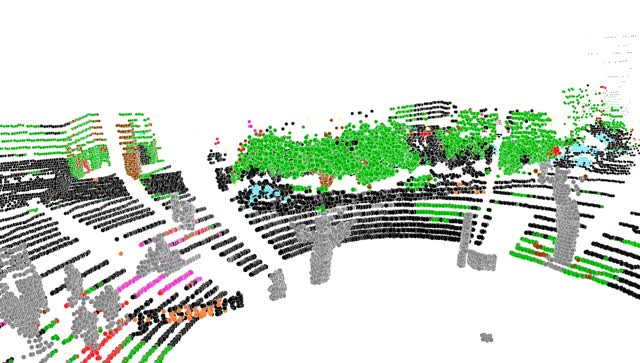}
        \end{overpic} &
        \begin{overpic}[width=0.19\textwidth]{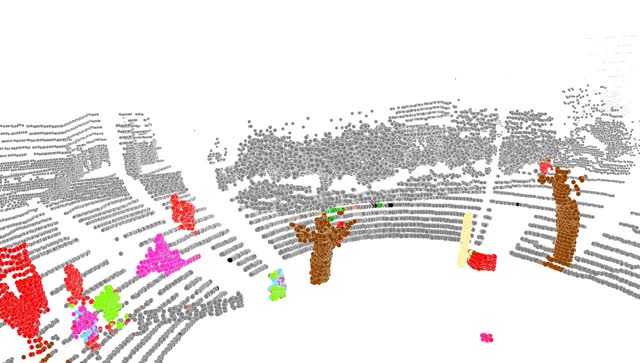}
        \end{overpic} &
        \begin{overpic}[width=0.19\textwidth]{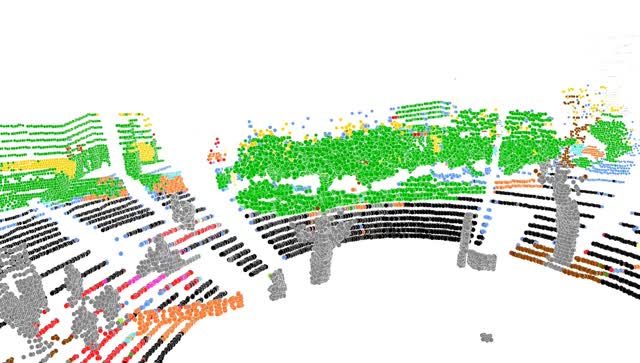}
        \end{overpic}
        \begin{overpic}[width=0.19\textwidth]{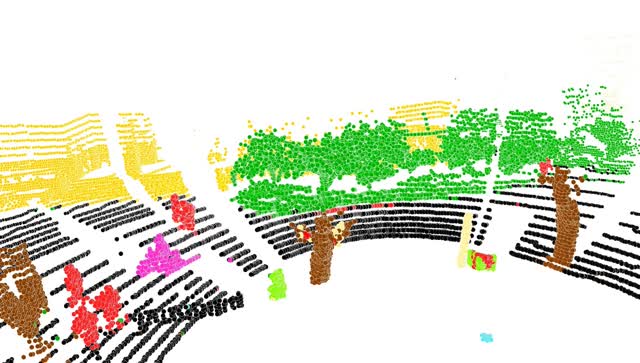}
        \end{overpic}
        \\
        \begin{overpic}[width=0.19\textwidth]{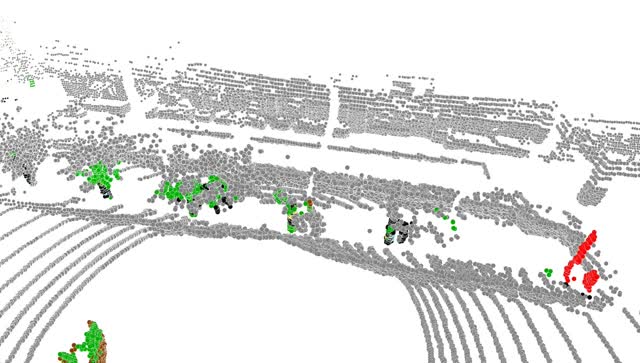}
        \end{overpic} &  
        \begin{overpic}[width=0.19\textwidth]{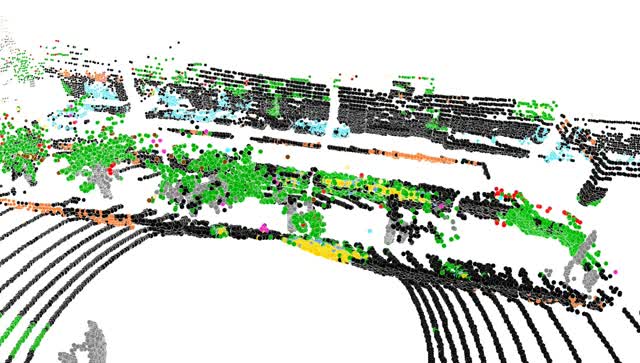}
        \end{overpic} &
        \begin{overpic}[width=0.19\textwidth]{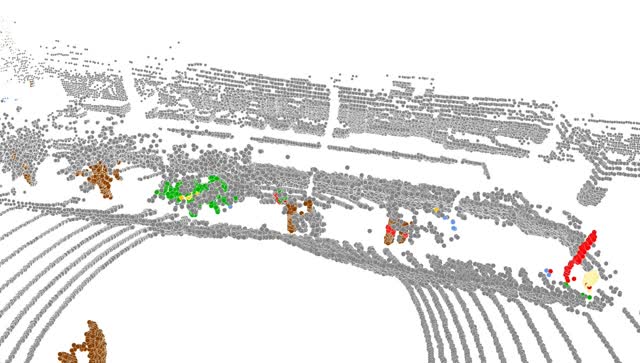}
        \end{overpic} &
        \begin{overpic}[width=0.19\textwidth]{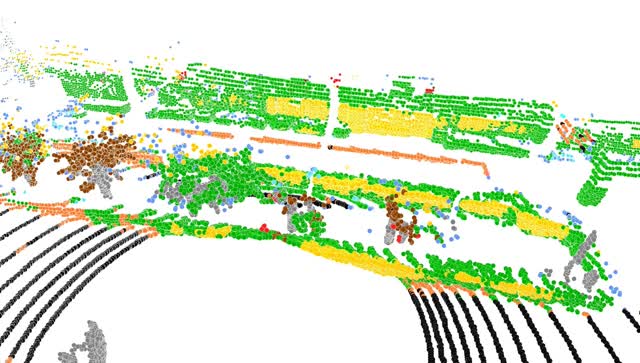}
        \end{overpic}
        \begin{overpic}[width=0.19\textwidth]{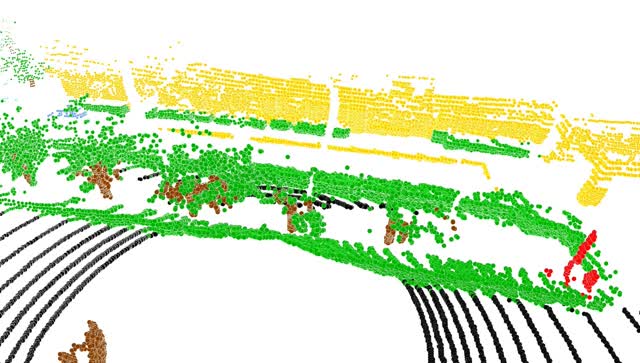}
        \end{overpic}
        \\
        \begin{overpic}[width=0.19\textwidth]{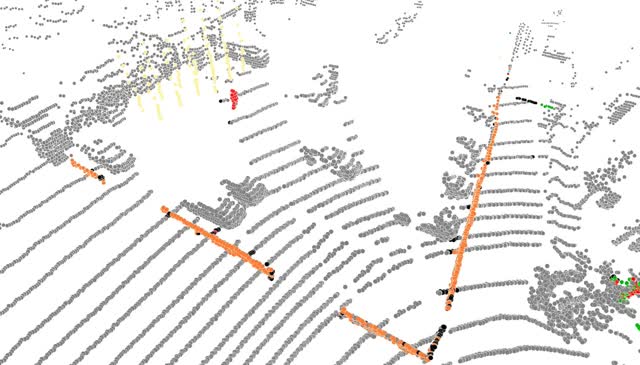}
        \end{overpic} &  
        \begin{overpic}[width=0.19\textwidth]{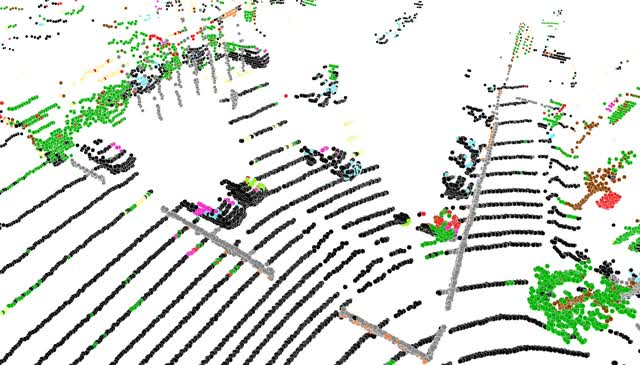}
        \end{overpic} &
        \begin{overpic}[width=0.19\textwidth]{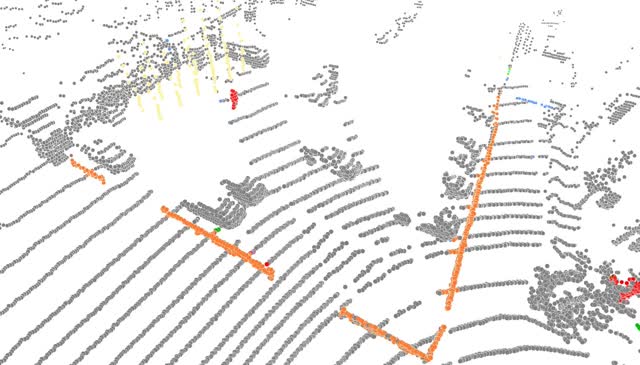}
        \end{overpic} &
        \begin{overpic}[width=0.19\textwidth]{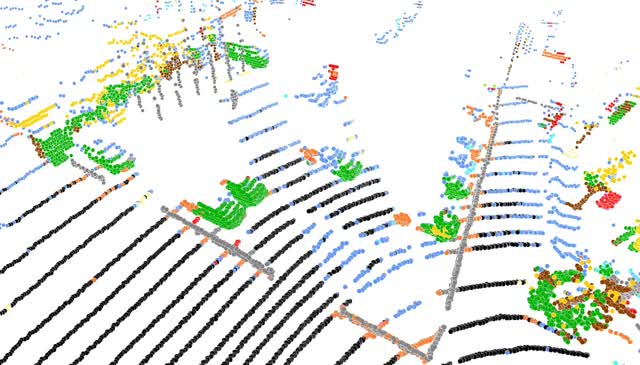}
        \end{overpic}
        \begin{overpic}[width=0.19\textwidth]{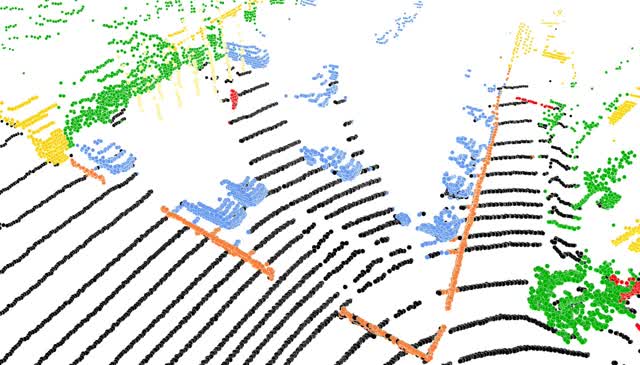}
        \end{overpic}
        \\
        \multicolumn{5}{c}{
        \begin{overpic}[width=0.99\textwidth]{images/qualitative/poss/legend_poss.pdf}
        \end{overpic}}
    \end{tabular}
    \caption{Qualitative comparison on SemanticPOSS from POSS-$4^0$. EUMS$^\dag$~\cite{zhao2022novel} predicts wrong and cluttered predictions on the novel classes. \ourmethod provides improved predictions by assigning the correct classes to the majority of the points and only a minority are misclassified.}
    \label{fig:supp_qualitative_poss0}
\end{figure*}

\begin{figure*}[t]
\centering
    \setlength\tabcolsep{2.5pt}
    \begin{tabular}{ccccc}
    \raggedright
        \begin{overpic}[width=0.19\textwidth]{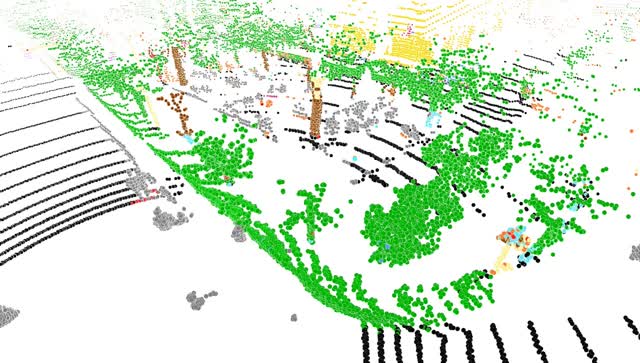}
        \put(20,60){\color{black}\footnotesize \textbf{EUMS$^\dag$~\cite{zhao2022novel} base}}
        \put(130,60){\color{black}\footnotesize \textbf{EUMS$^\dag$~\cite{zhao2022novel} novel}}
        \put(225,60){\color{black}\footnotesize \textbf{\ourmethod base (Ours)}}
        \put(330,60){\color{black}\footnotesize \textbf{\ourmethod novel (Ours)}}
        \put(460,60){\color{black}\footnotesize \textbf{GT}}
        \end{overpic} &  
        \begin{overpic}[width=0.19\textwidth]{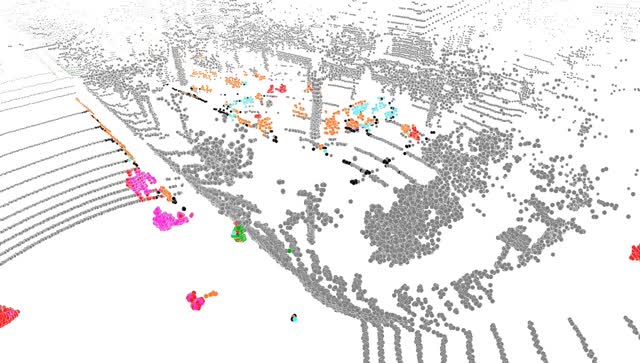}
        \end{overpic} &
        \begin{overpic}[width=0.19\textwidth]{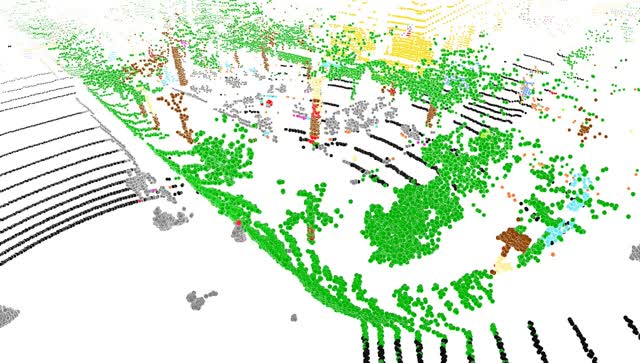}
        \end{overpic} &
        \begin{overpic}[width=0.19\textwidth]{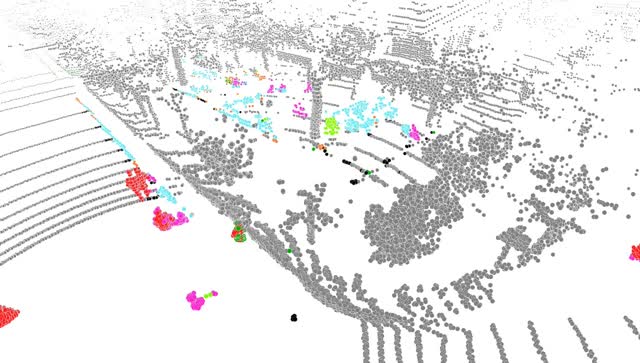}
        \end{overpic}
        \begin{overpic}[width=0.19\textwidth]{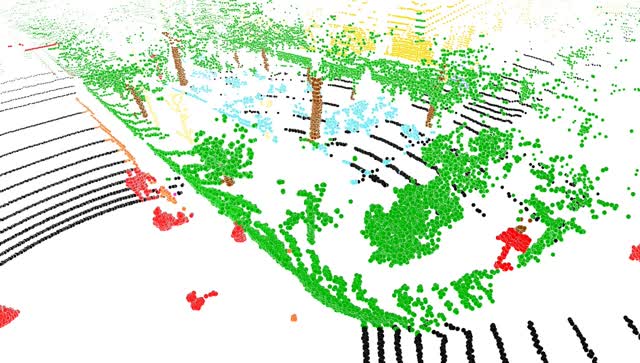}
        \end{overpic}
        \\
        \begin{overpic}[width=0.19\textwidth]{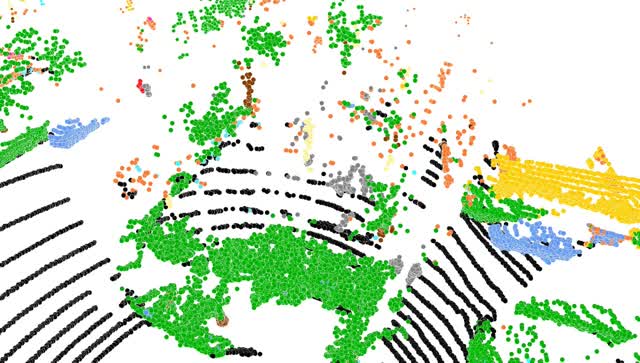}
        \end{overpic} &  
        \begin{overpic}[width=0.19\textwidth]{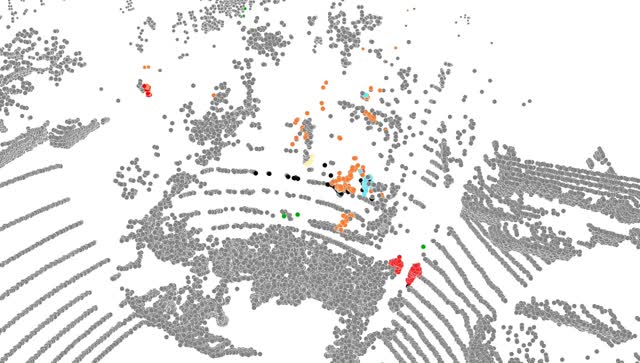}
        \end{overpic} &
        \begin{overpic}[width=0.19\textwidth]{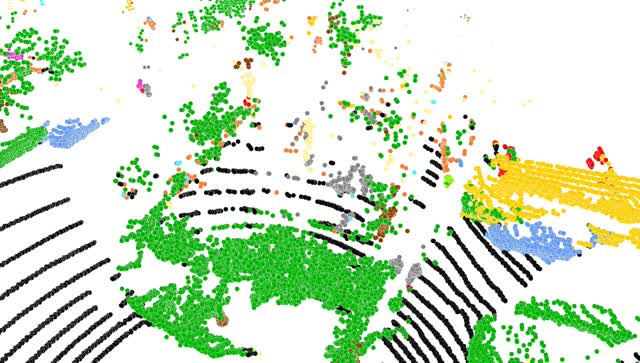}
        \end{overpic} &
        \begin{overpic}[width=0.19\textwidth]{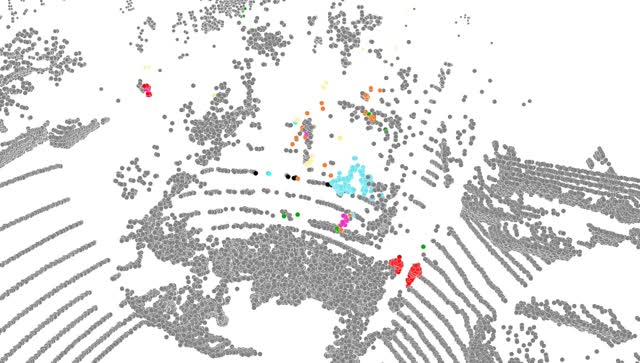}
        \end{overpic}
        \begin{overpic}[width=0.19\textwidth]{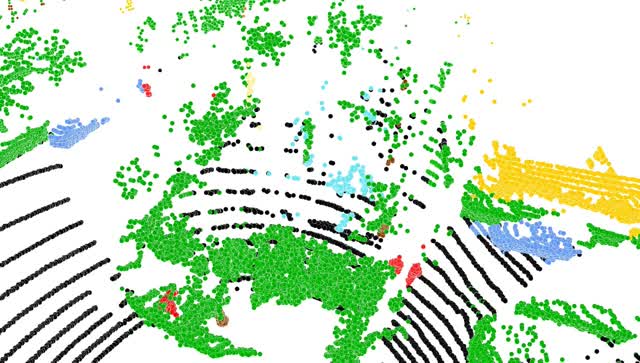}
        \end{overpic}
        \\
        \begin{overpic}[width=0.19\textwidth]{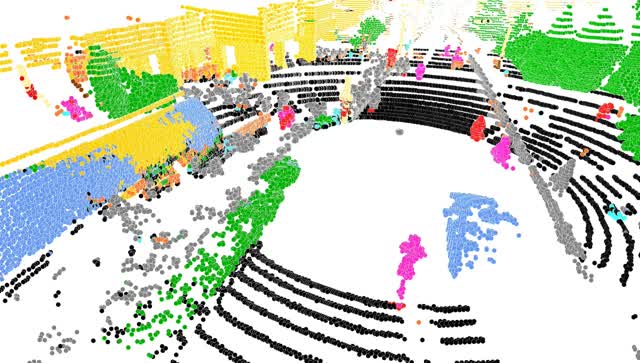}
        \end{overpic} &  
        \begin{overpic}[width=0.19\textwidth]{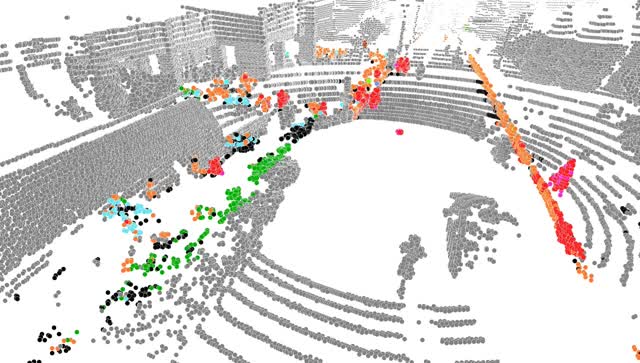}
        \end{overpic} &
        \begin{overpic}[width=0.19\textwidth]{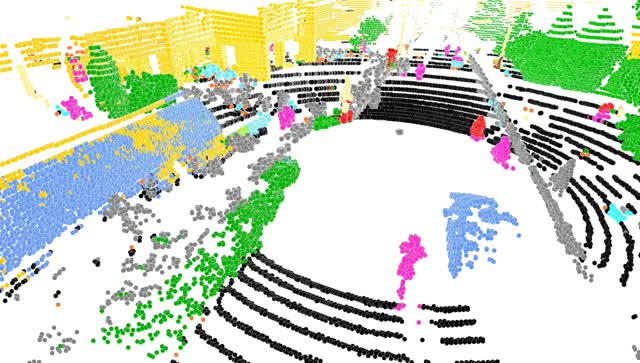}
        \end{overpic} &
        \begin{overpic}[width=0.19\textwidth]{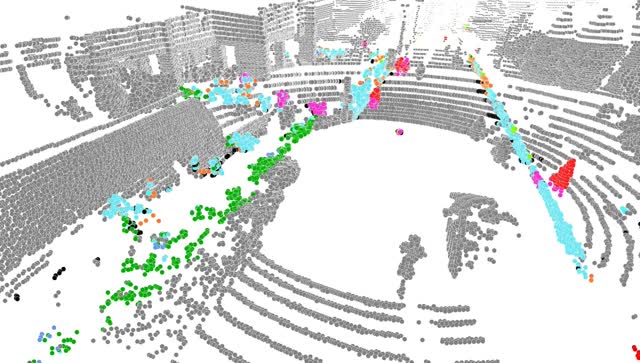}
        \end{overpic}
        \begin{overpic}[width=0.19\textwidth]{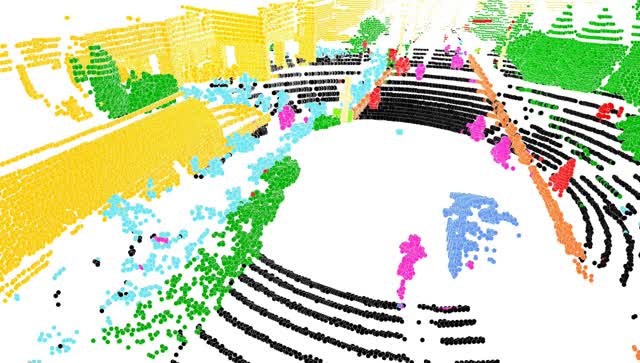}
        \end{overpic}
        \\
        \begin{overpic}[width=0.19\textwidth]{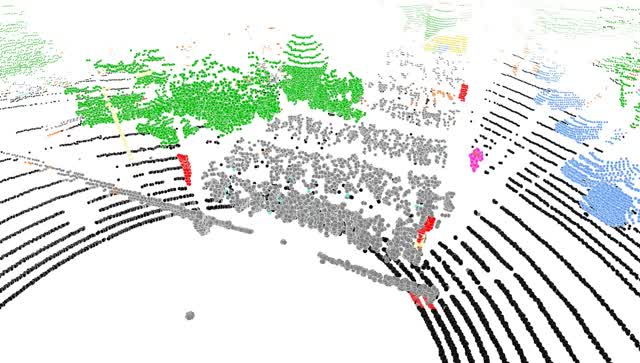}
        \end{overpic} &  
        \begin{overpic}[width=0.19\textwidth]{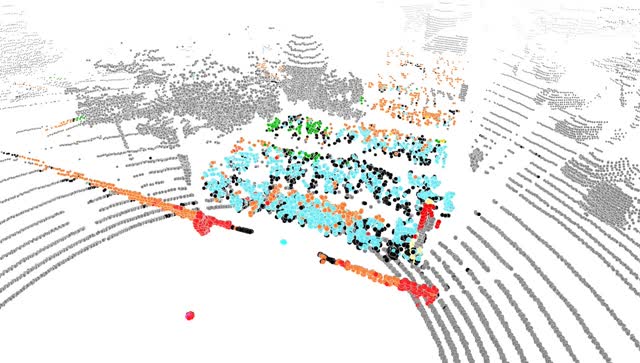}
        \end{overpic} &
        \begin{overpic}[width=0.19\textwidth]{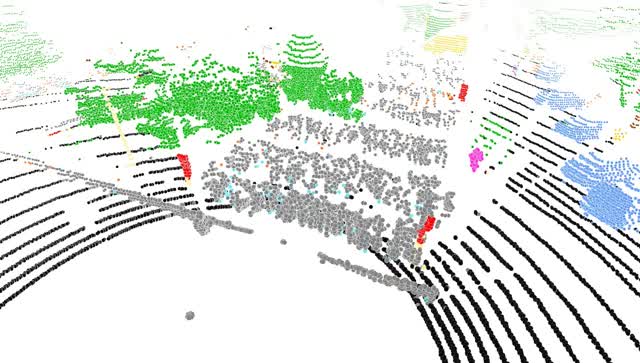}
        \end{overpic} &
        \begin{overpic}[width=0.19\textwidth]{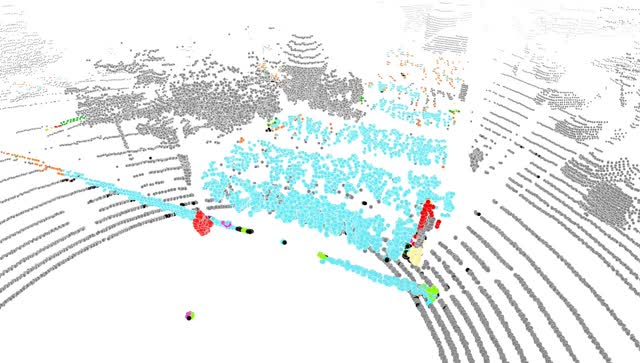}
        \end{overpic}
        \begin{overpic}[width=0.19\textwidth]{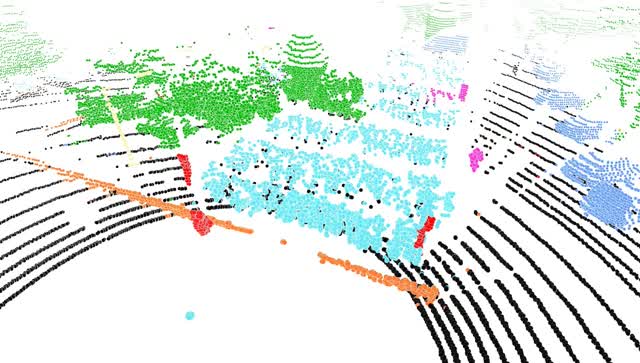}
        \end{overpic}
        \\
        \multicolumn{5}{c}{
        \begin{overpic}[width=0.99\textwidth]{images/qualitative/poss/legend_poss.pdf}
        \end{overpic}}
    \end{tabular}
    \caption{Qualitative comparison on SemanticPOSS from POSS-$3^1$. EUMS$^\dag$~\cite{zhao2022novel} predicts wrong and cluttered predictions on the novel classes. \ourmethod provides improved predictions by assigning the correct classes to the majority of the points and only a minority are misclassified.}
    \label{fig:supp_qualitative_poss1}
\end{figure*}

\begin{figure*}[t]
\centering
    \setlength\tabcolsep{2.5pt}
    \begin{tabular}{ccccc}
    \raggedright
        \begin{overpic}[width=0.19\textwidth]{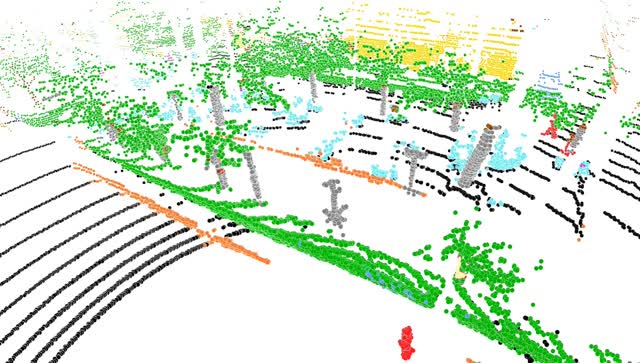}
        \put(20,60){\color{black}\footnotesize \textbf{EUMS$^\dag$~\cite{zhao2022novel} base}}
        \put(130,60){\color{black}\footnotesize \textbf{EUMS$^\dag$~\cite{zhao2022novel} novel}}
        \put(225,60){\color{black}\footnotesize \textbf{\ourmethod base (Ours)}}
        \put(330,60){\color{black}\footnotesize \textbf{\ourmethod novel (Ours)}}
        \put(460,60){\color{black}\footnotesize \textbf{GT}}
        \end{overpic} &  
        \begin{overpic}[width=0.19\textwidth]{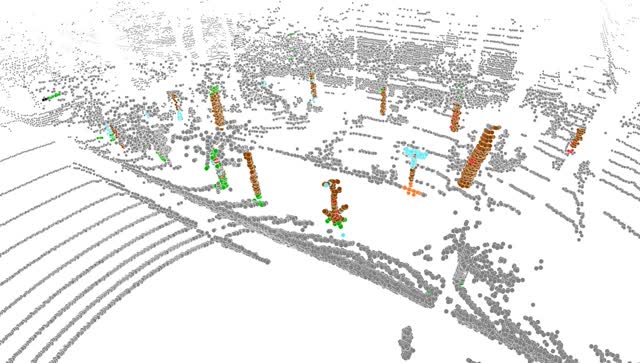}
        \end{overpic} &
        \begin{overpic}[width=0.19\textwidth]{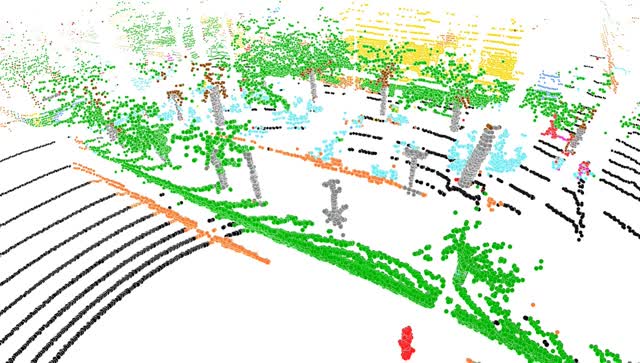}
        \end{overpic} &
        \begin{overpic}[width=0.19\textwidth]{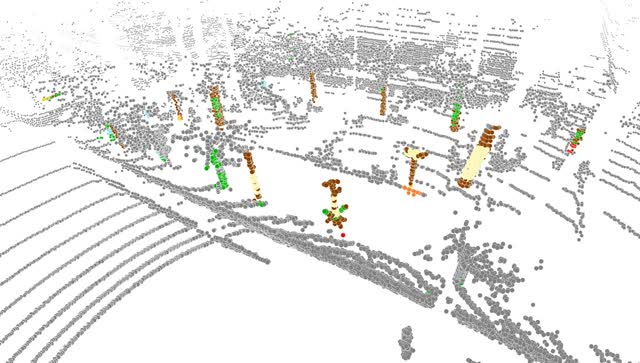}
        \end{overpic}
        \begin{overpic}[width=0.19\textwidth]{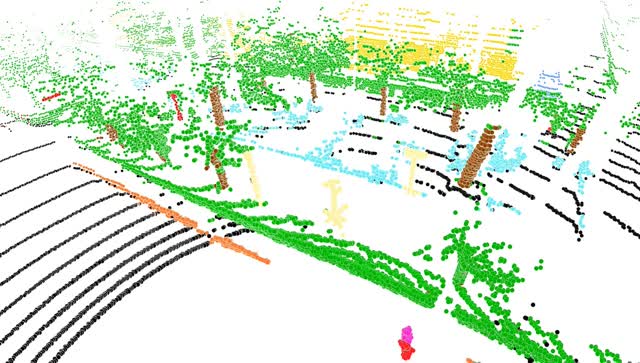}
        \end{overpic}
        \\
        \begin{overpic}[width=0.19\textwidth]{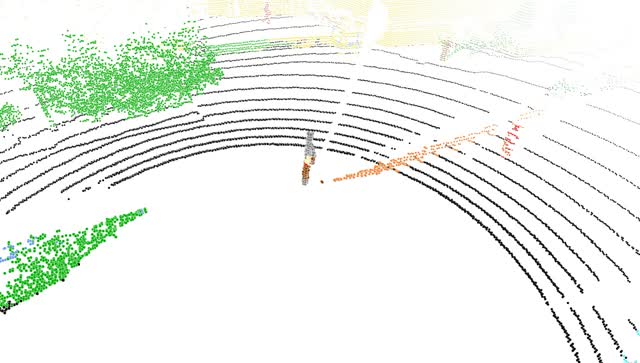}
        \end{overpic} &  
        \begin{overpic}[width=0.19\textwidth]{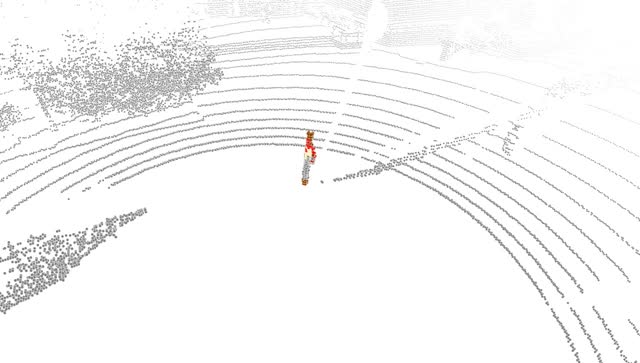}
        \end{overpic} &
        \begin{overpic}[width=0.19\textwidth]{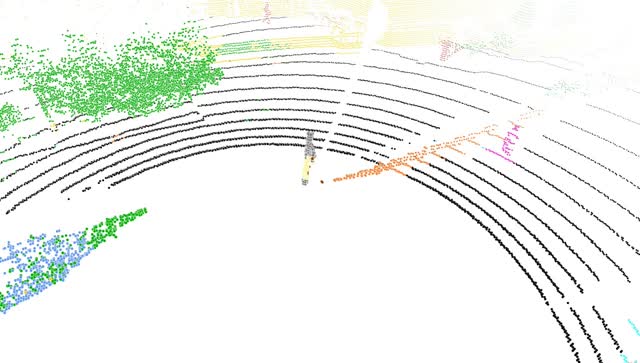}
        \end{overpic} &
        \begin{overpic}[width=0.19\textwidth]{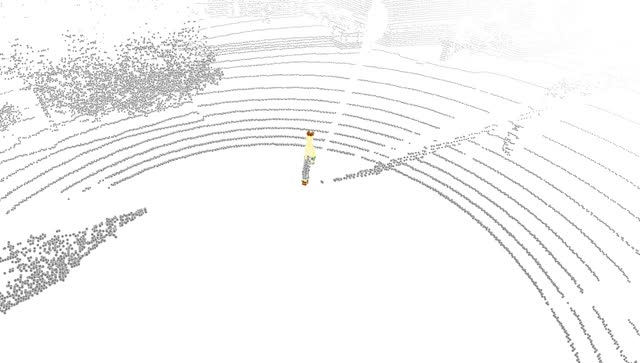}
        \end{overpic}
        \begin{overpic}[width=0.19\textwidth]{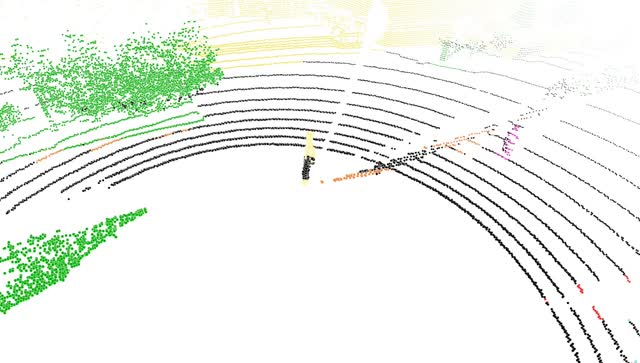}
        \end{overpic}
        \\
        \begin{overpic}[width=0.19\textwidth]{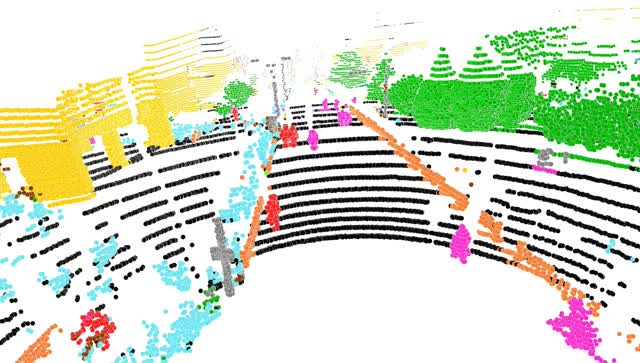}
        \end{overpic} &  
        \begin{overpic}[width=0.19\textwidth]{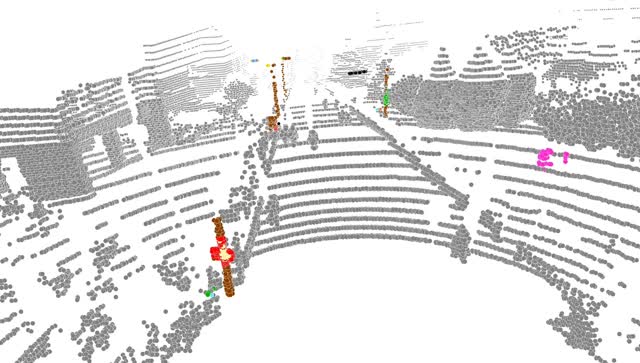}
        \end{overpic} &
        \begin{overpic}[width=0.19\textwidth]{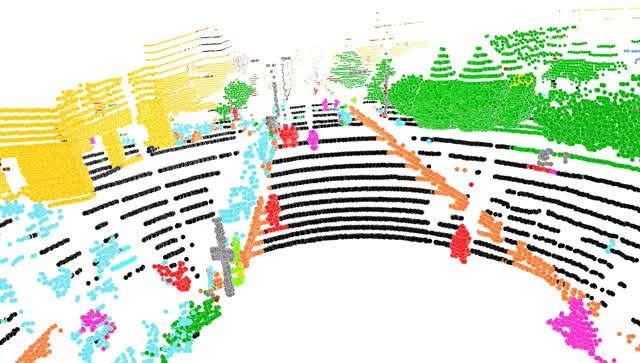}
        \end{overpic} &
        \begin{overpic}[width=0.19\textwidth]{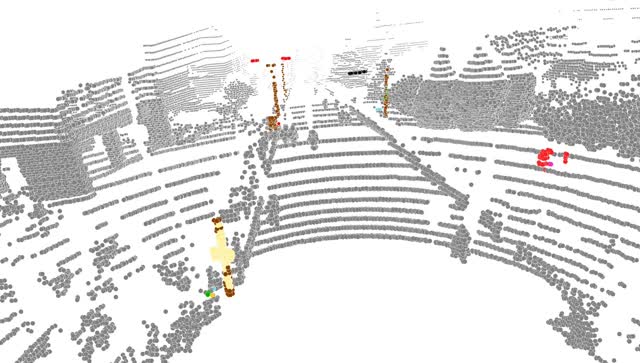}
        \end{overpic}
        \begin{overpic}[width=0.19\textwidth]{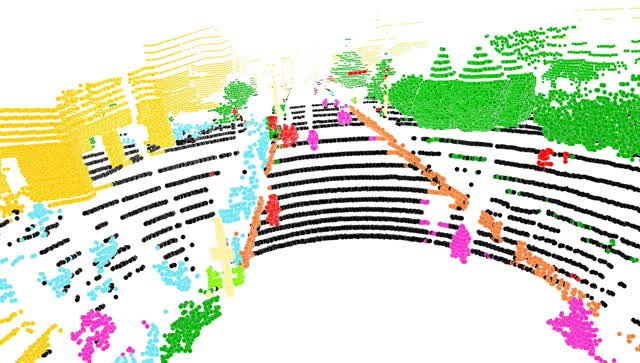}
        \end{overpic}
        \\
        \begin{overpic}[width=0.19\textwidth]{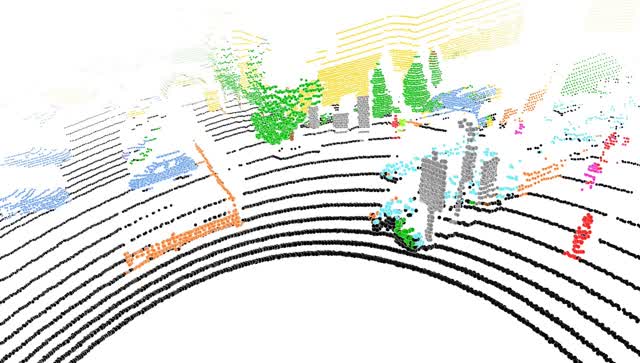}
        \end{overpic} &  
        \begin{overpic}[width=0.19\textwidth]{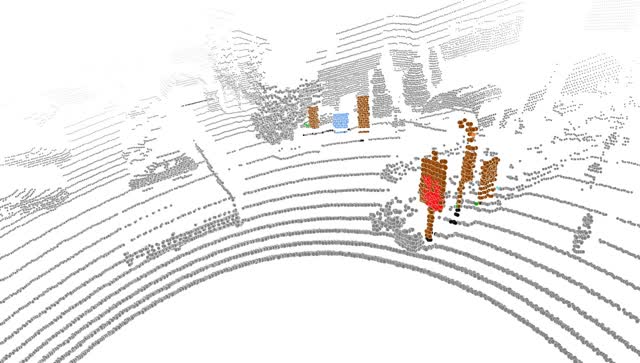}
        \end{overpic} &
        \begin{overpic}[width=0.19\textwidth]{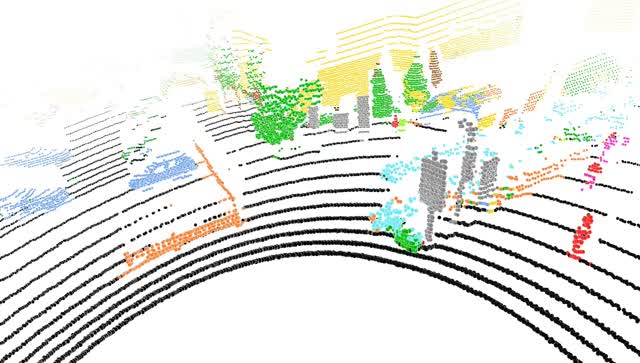}
        \end{overpic} &
        \begin{overpic}[width=0.19\textwidth]{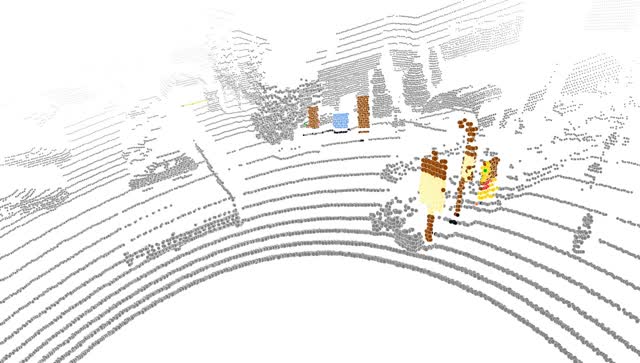}
        \end{overpic}
        \begin{overpic}[width=0.19\textwidth]{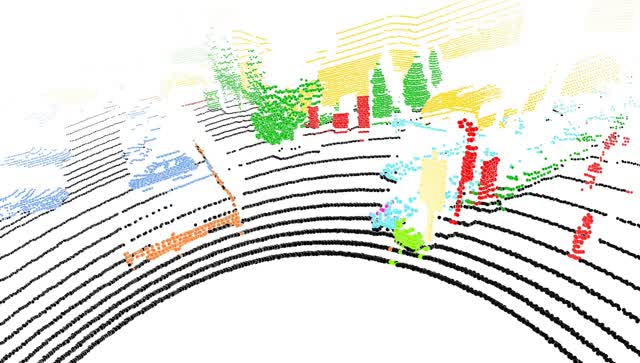}
        \end{overpic}
        \\
        \multicolumn{5}{c}{
        \begin{overpic}[width=0.99\textwidth]{images/qualitative/poss/legend_poss.pdf}
        \end{overpic}}
    \end{tabular}
    \caption{Qualitative comparison on SemanticPOSS from POSS-$3^2$. EUMS$^\dag$~\cite{zhao2022novel} predicts wrong and cluttered predictions on the novel classes. \ourmethod provides improved predictions by assigning the correct classes to the majority of the points and only a minority are misclassified.}
    \label{fig:supp_qualitative_poss2}
\end{figure*}

\begin{figure*}[t]
\centering
    \setlength\tabcolsep{2.5pt}
    \begin{tabular}{ccccc}
    \raggedright
        \begin{overpic}[width=0.19\textwidth]{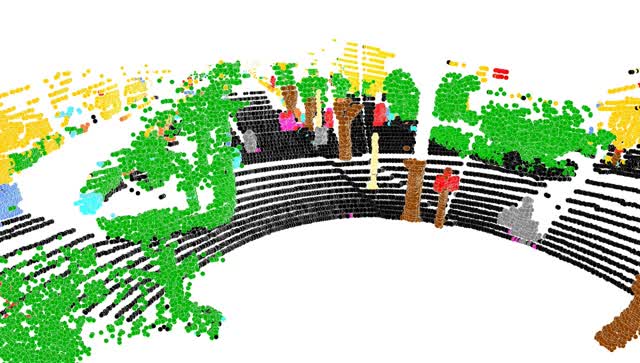}
        \put(20,60){\color{black}\footnotesize \textbf{EUMS$^\dag$~\cite{zhao2022novel} base}}
        \put(130,60){\color{black}\footnotesize \textbf{EUMS$^\dag$~\cite{zhao2022novel} novel}}
        \put(225,60){\color{black}\footnotesize \textbf{\ourmethod base (Ours)}}
        \put(330,60){\color{black}\footnotesize \textbf{\ourmethod novel (Ours)}}
        \put(460,60){\color{black}\footnotesize \textbf{GT}}
        \end{overpic} &  
        \begin{overpic}[width=0.19\textwidth]{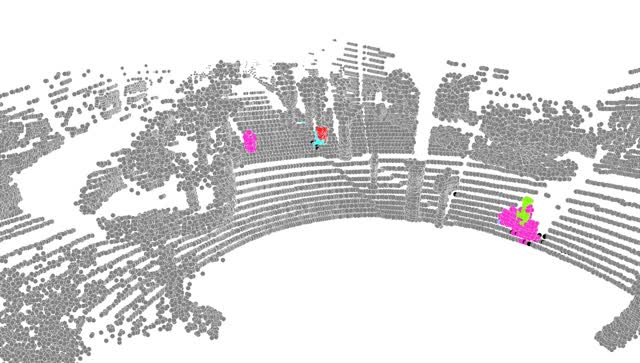}
        \end{overpic} &
        \begin{overpic}[width=0.19\textwidth]{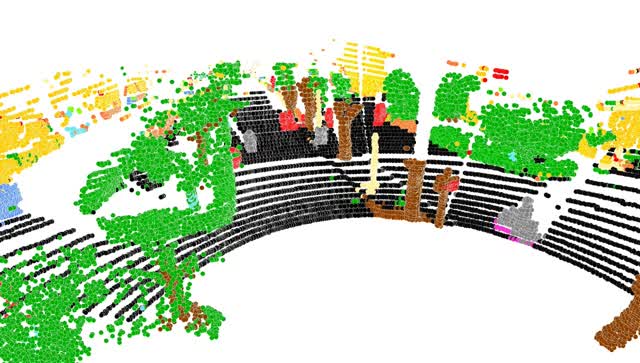}
        \end{overpic} &
        \begin{overpic}[width=0.19\textwidth]{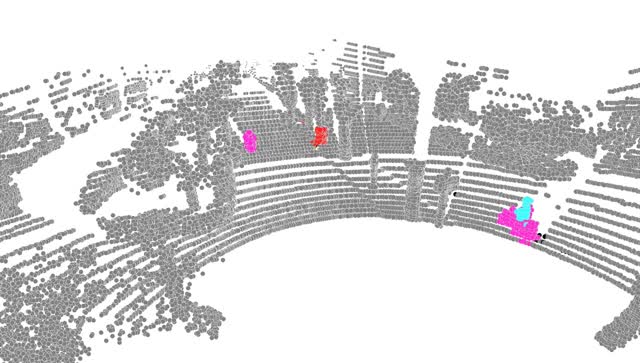}
        \end{overpic}
        \begin{overpic}[width=0.19\textwidth]{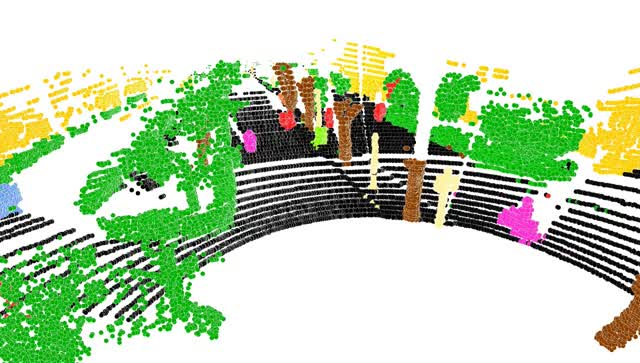}
        \end{overpic}
        \\
        \begin{overpic}[width=0.19\textwidth]{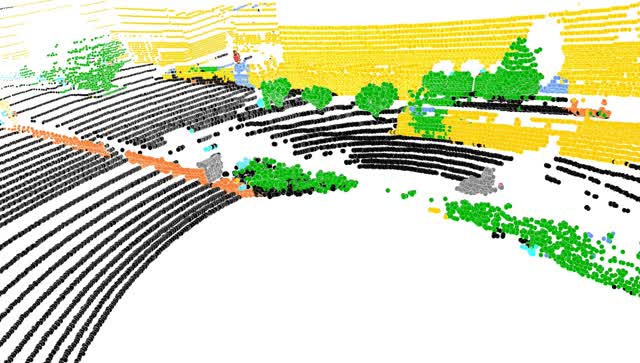}
        \end{overpic} &  
        \begin{overpic}[width=0.19\textwidth]{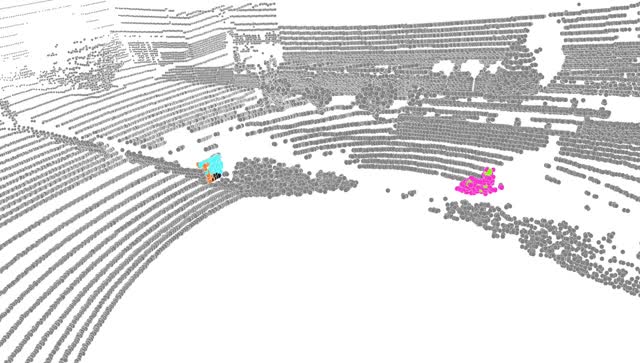}
        \end{overpic} &
        \begin{overpic}[width=0.19\textwidth]{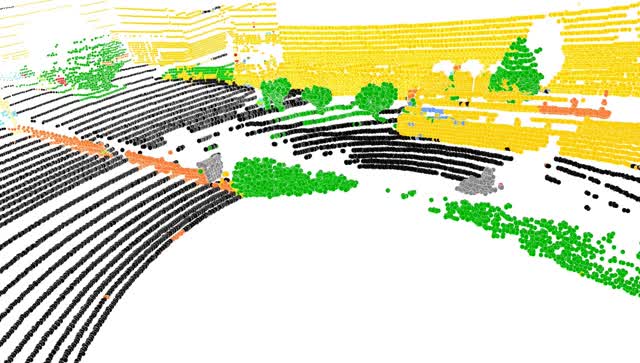}
        \end{overpic} &
        \begin{overpic}[width=0.19\textwidth]{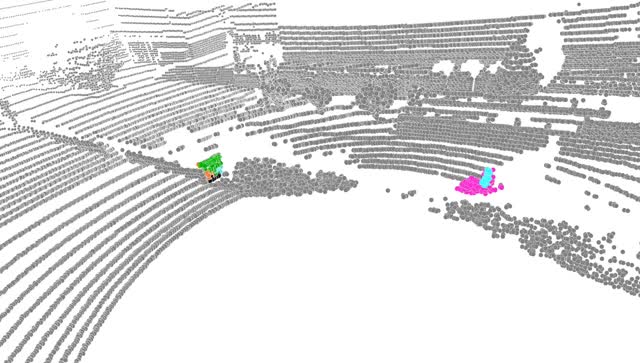}
        \end{overpic}
        \begin{overpic}[width=0.19\textwidth]{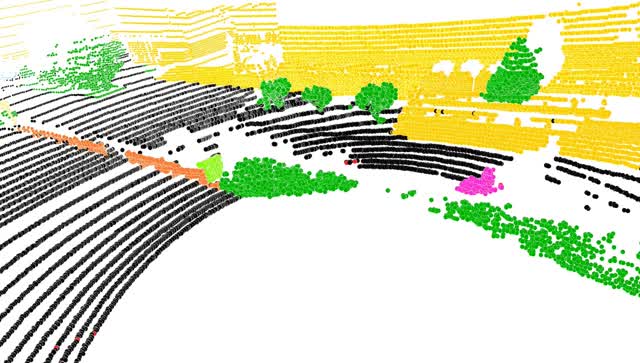}
        \end{overpic}
        \\
        \begin{overpic}[width=0.19\textwidth]{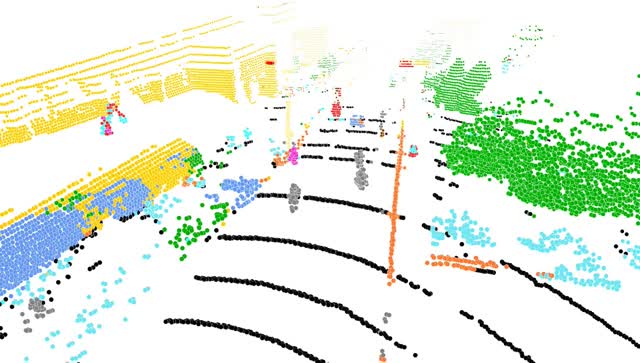}
        \end{overpic} &  
        \begin{overpic}[width=0.19\textwidth]{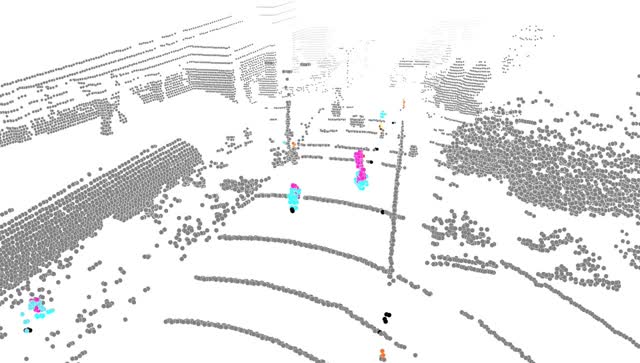}
        \end{overpic} &
        \begin{overpic}[width=0.19\textwidth]{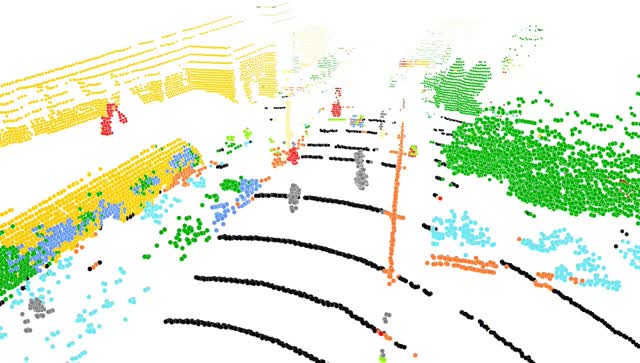}
        \end{overpic} &
        \begin{overpic}[width=0.19\textwidth]{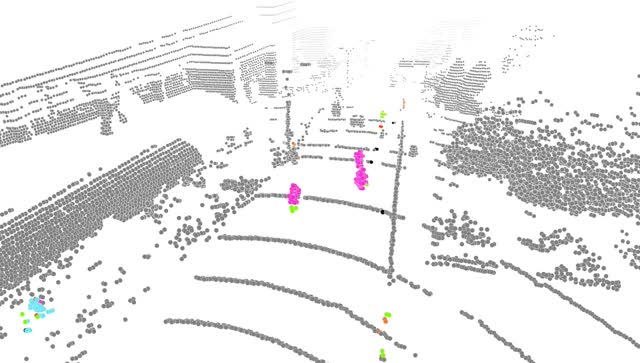}
        \end{overpic}
        \begin{overpic}[width=0.19\textwidth]{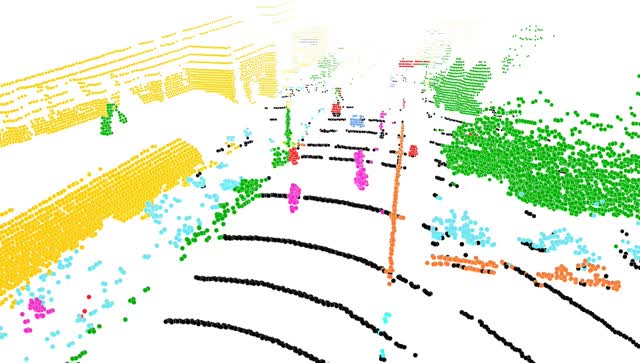}
        \end{overpic}
        \\
        \begin{overpic}[width=0.19\textwidth]{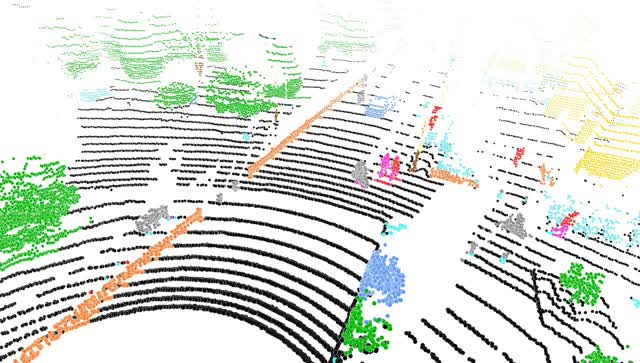}
        \end{overpic} &  
        \begin{overpic}[width=0.19\textwidth]{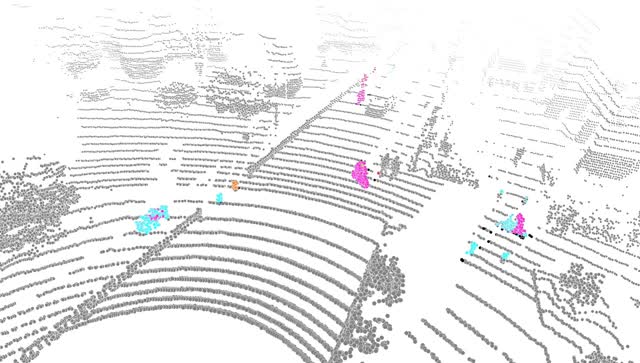}
        \end{overpic} &
        \begin{overpic}[width=0.19\textwidth]{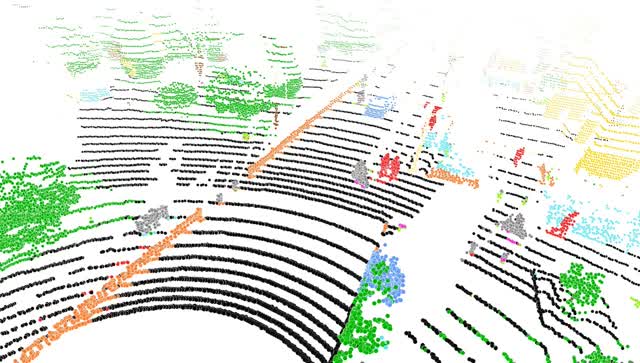}
        \end{overpic} &
        \begin{overpic}[width=0.19\textwidth]{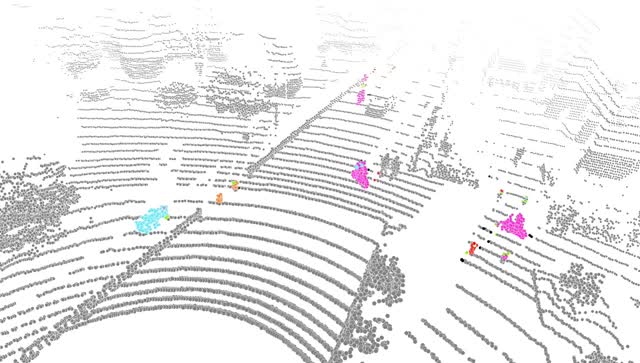}
        \end{overpic}
        \begin{overpic}[width=0.19\textwidth]{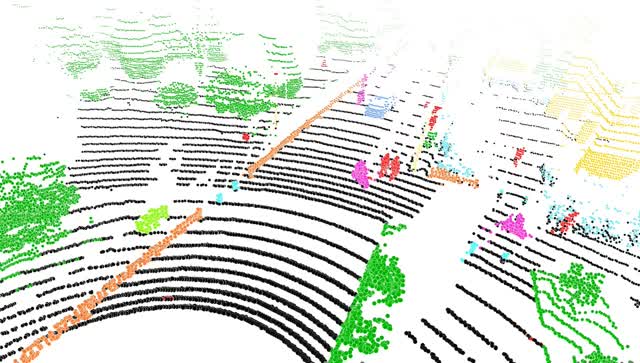}
        \end{overpic}
        \\
        \multicolumn{5}{c}{
        \begin{overpic}[width=0.99\textwidth]{images/qualitative/poss/legend_poss.pdf}
        \end{overpic}}
    \end{tabular}
    \caption{Qualitative comparison on SemanticPOSS from POSS-$3^3$. EUMS$^\dag$~\cite{zhao2022novel} predicts wrong and cluttered predictions on the novel classes. \ourmethod provides improved predictions by assigning the correct classes to the majority of the points and only a minority are misclassified.}
    \label{fig:supp_qualitative_poss3}
\end{figure*}

Figs.~\ref{fig:supp_qualitative_poss0}-\ref{fig:supp_qualitative_poss3}, show qualitative results for SemanticPOSS from the split POSS-$4^0$ (Fig.~\ref{fig:supp_qualitative_poss0}), POSS-$3^1$ (Fig.~\ref{fig:supp_qualitative_poss1}), POSS-$3^2$ (Fig.~\ref{fig:supp_qualitative_poss2}) and POSS-$3^3$ (Fig.~\ref{fig:supp_qualitative_poss3}).

\begin{figure*}[t]
\centering
    \setlength\tabcolsep{2.5pt}
    \begin{tabular}{ccccc}
    \raggedright
        \begin{overpic}[width=0.19\textwidth]{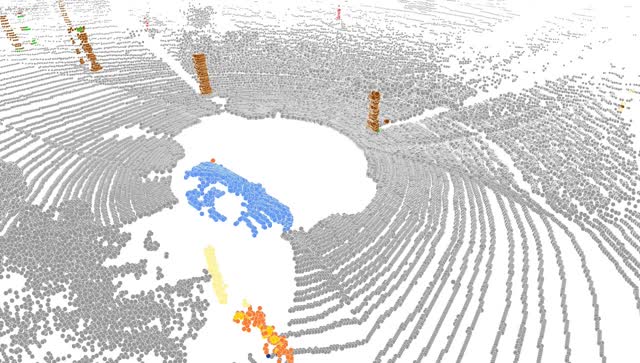}
        \put(20,60){\color{black}\footnotesize \textbf{EUMS$^\dag$~\cite{zhao2022novel} base}}
        \put(130,60){\color{black}\footnotesize \textbf{EUMS$^\dag$~\cite{zhao2022novel} novel}}
        \put(225,60){\color{black}\footnotesize \textbf{\ourmethod base (Ours)}}
        \put(330,60){\color{black}\footnotesize \textbf{\ourmethod novel (Ours)}}
        \put(460,60){\color{black}\footnotesize \textbf{GT}}
        \end{overpic} &  
        \begin{overpic}[width=0.19\textwidth]{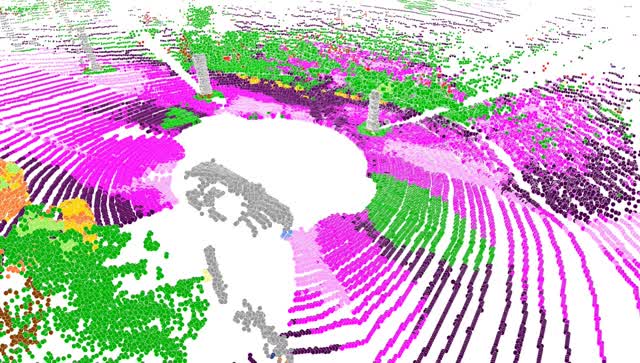}
        \end{overpic} &
        \begin{overpic}[width=0.19\textwidth]{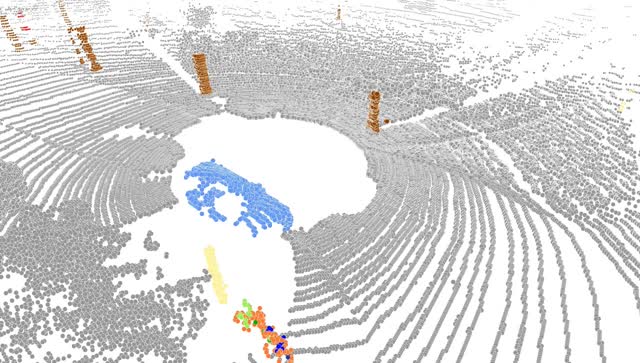}
        \end{overpic} &
        \begin{overpic}[width=0.19\textwidth]{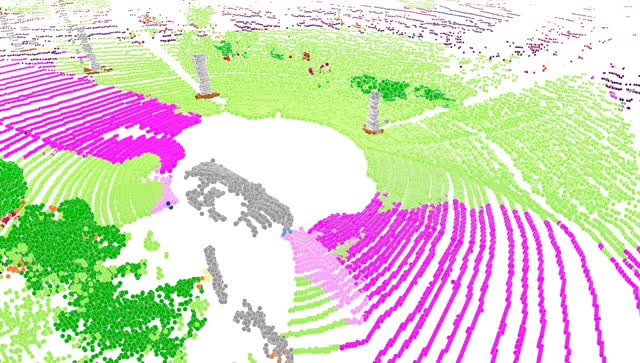}
        \end{overpic}
        \begin{overpic}[width=0.19\textwidth]{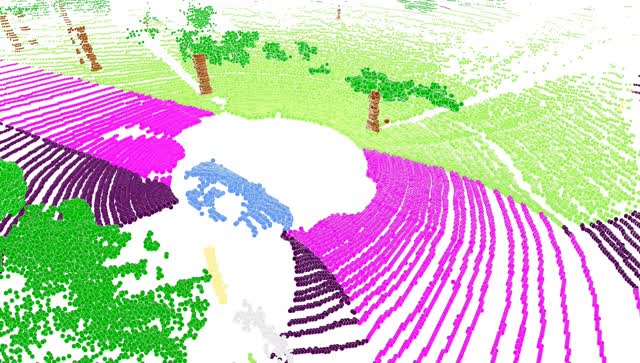}
        \end{overpic}
        \\
        \begin{overpic}[width=0.19\textwidth]{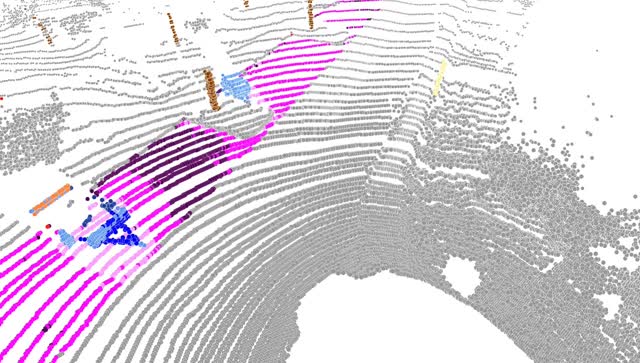}
        \end{overpic} &  
        \begin{overpic}[width=0.19\textwidth]{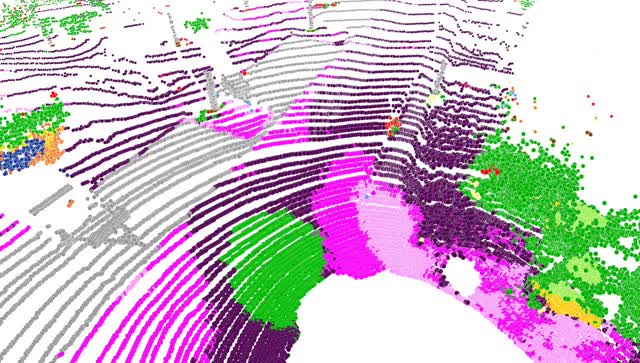}
        \end{overpic} &
        \begin{overpic}[width=0.19\textwidth]{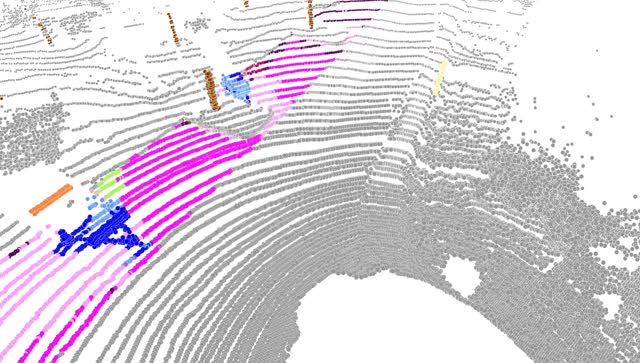}
        \end{overpic} &
        \begin{overpic}[width=0.19\textwidth]{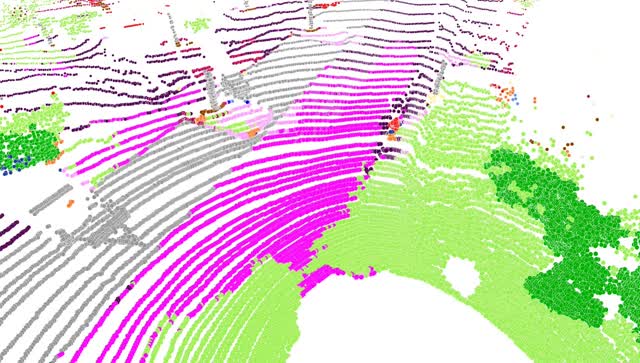}
        \end{overpic}
        \begin{overpic}[width=0.19\textwidth]{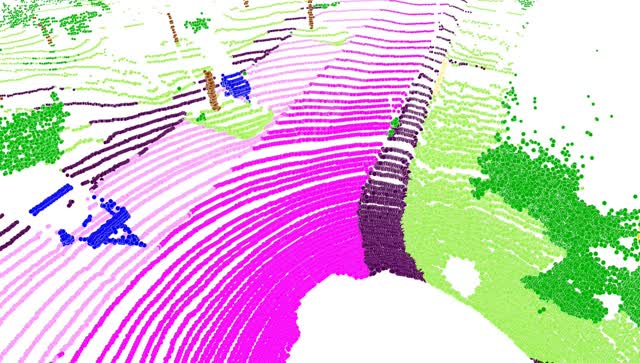}
        \end{overpic}
        \\
        \begin{overpic}[width=0.19\textwidth]{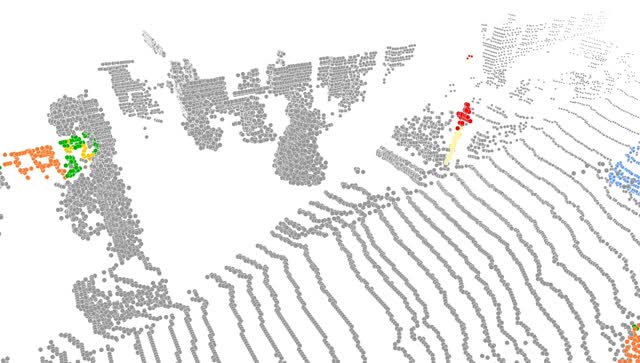}
        \end{overpic} &  
        \begin{overpic}[width=0.19\textwidth]{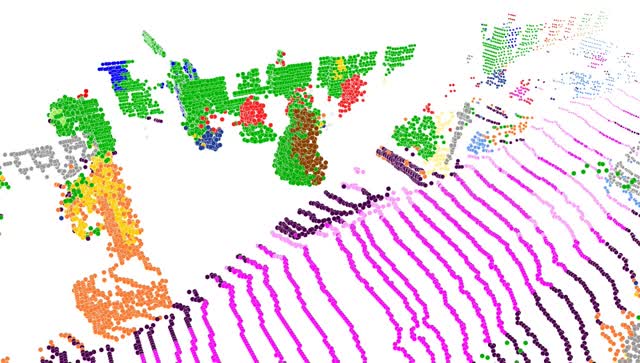}
        \end{overpic} &
        \begin{overpic}[width=0.19\textwidth]{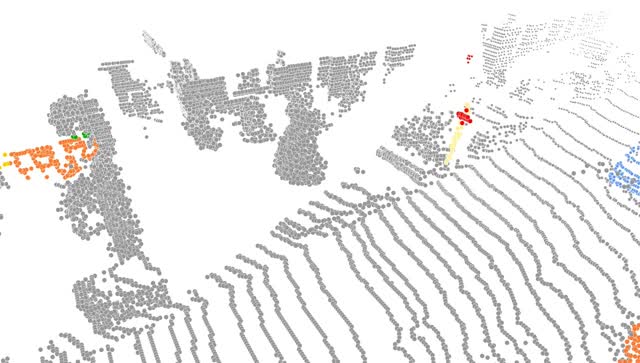}
        \end{overpic} &
        \begin{overpic}[width=0.19\textwidth]{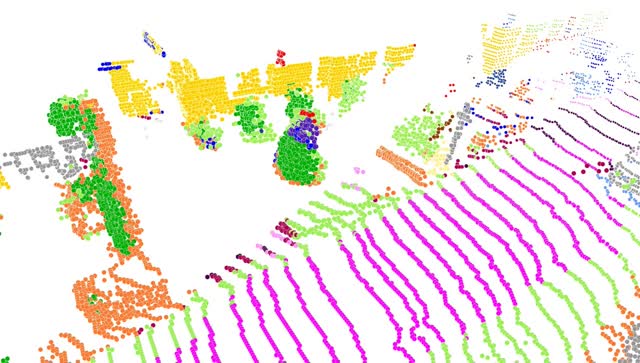}
        \end{overpic}
        \begin{overpic}[width=0.19\textwidth]{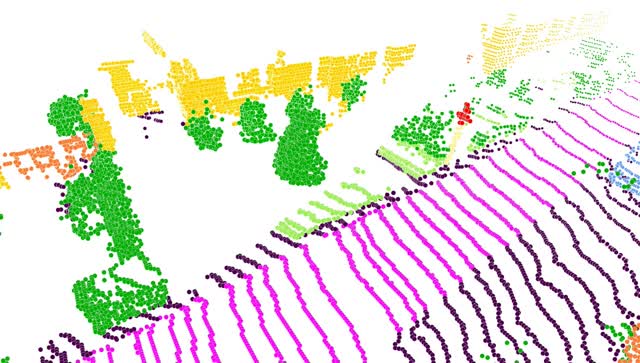}
        \end{overpic}
        \\
        \begin{overpic}[width=0.19\textwidth]{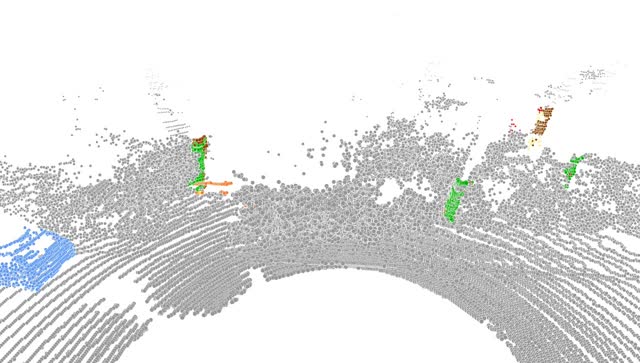}
        \end{overpic} &  
        \begin{overpic}[width=0.19\textwidth]{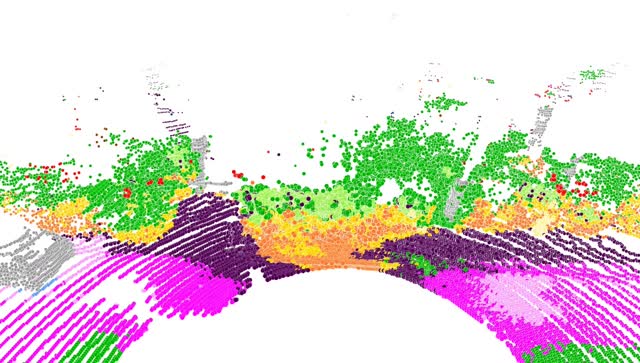}
        \end{overpic} &
        \begin{overpic}[width=0.19\textwidth]{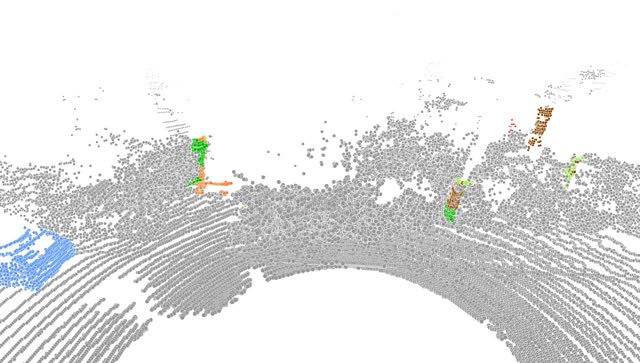}
        \end{overpic} &
        \begin{overpic}[width=0.19\textwidth]{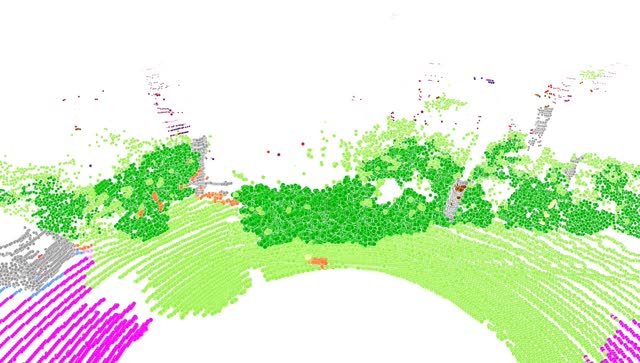}
        \end{overpic}
        \begin{overpic}[width=0.19\textwidth]{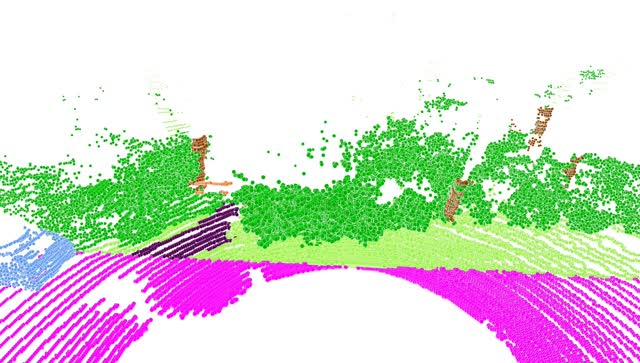}
        \end{overpic}
        \\
        \multicolumn{5}{c}{
        \begin{overpic}[width=0.99\textwidth]{images/qualitative/kitti/legend_kitti.pdf}
        \end{overpic}}
    \end{tabular}
    \caption{Qualitative comparison on SemanticKITTI from KITTI-$5^0$. EUMS$^\dag$~\cite{zhao2022novel} outputs are completely or partially wrong for the novel classes. \ourmethod improves the performance by providing correct and more homogeneous predictions.}
    \label{fig:supp_qualitative_kitti0}
\end{figure*}

\begin{figure*}[t]
\centering
    \setlength\tabcolsep{2.5pt}
    \begin{tabular}{ccccc}
    \raggedright
        \begin{overpic}[width=0.19\textwidth]{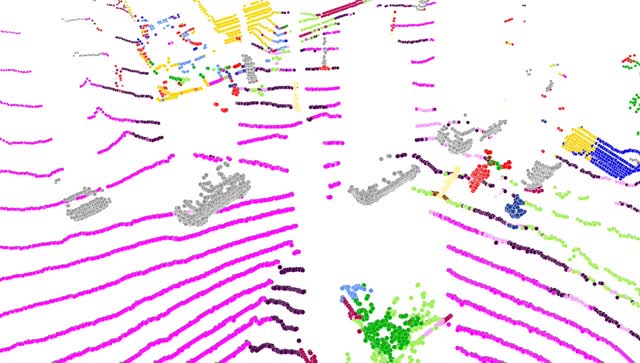}
        \put(20,60){\color{black}\footnotesize \textbf{EUMS$^\dag$~\cite{zhao2022novel} base}}
        \put(130,60){\color{black}\footnotesize \textbf{EUMS$^\dag$~\cite{zhao2022novel} novel}}
        \put(225,60){\color{black}\footnotesize \textbf{\ourmethod base (Ours)}}
        \put(330,60){\color{black}\footnotesize \textbf{\ourmethod novel (Ours)}}
        \put(460,60){\color{black}\footnotesize \textbf{GT}}
        \end{overpic} &  
        \begin{overpic}[width=0.19\textwidth]{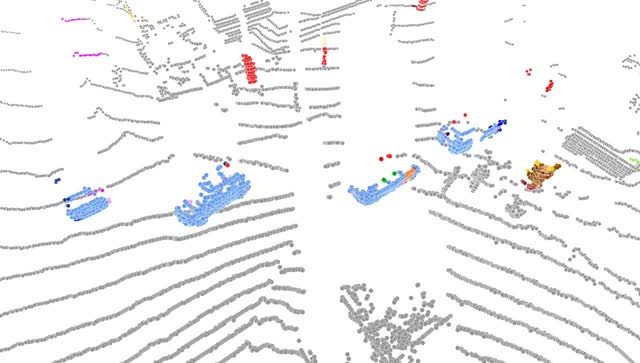}
        \end{overpic} &
        \begin{overpic}[width=0.19\textwidth]{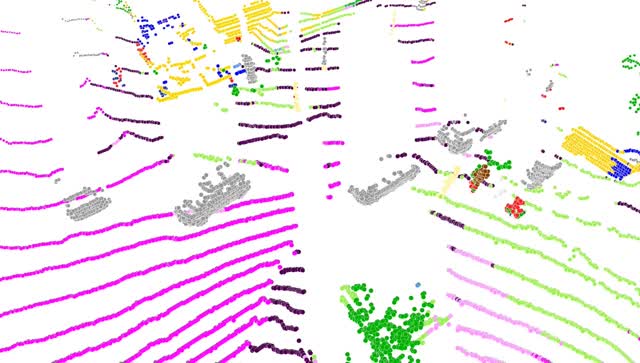}
        \end{overpic} &
        \begin{overpic}[width=0.19\textwidth]{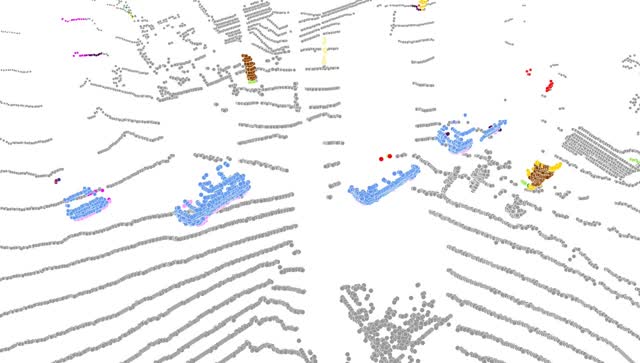}
        \end{overpic}
        \begin{overpic}[width=0.19\textwidth]{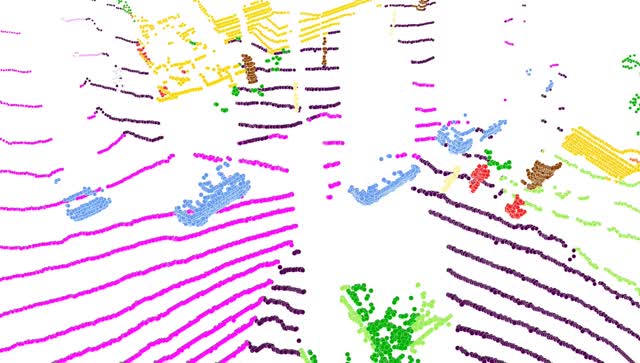}
        \end{overpic}
        \\
        \begin{overpic}[width=0.19\textwidth]{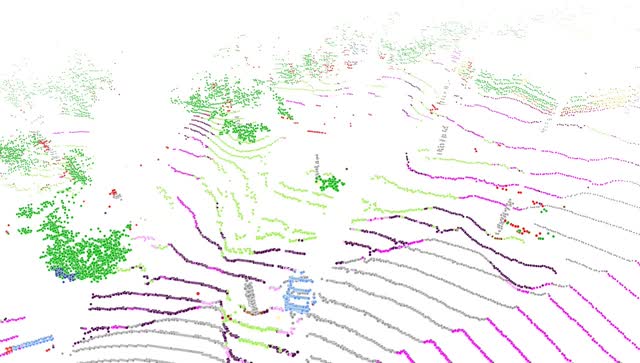}
        \end{overpic} &  
        \begin{overpic}[width=0.19\textwidth]{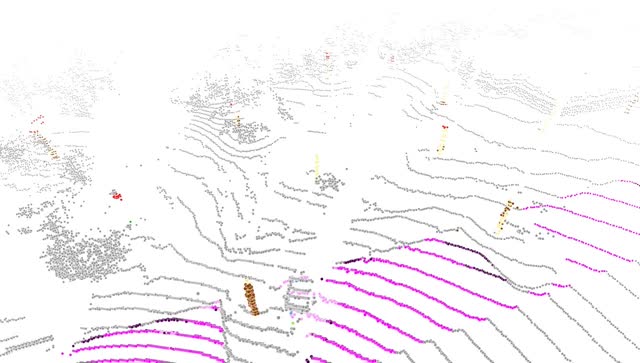}
        \end{overpic} &
        \begin{overpic}[width=0.19\textwidth]{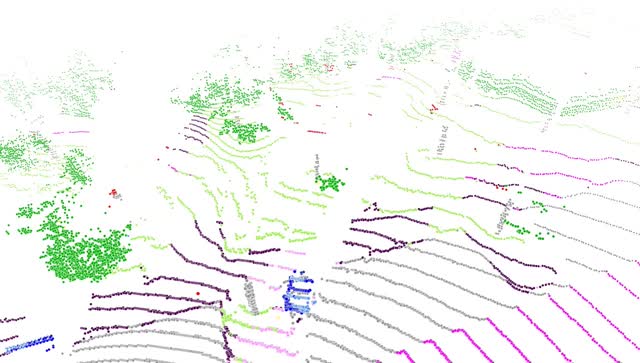}
        \end{overpic} &
        \begin{overpic}[width=0.19\textwidth]{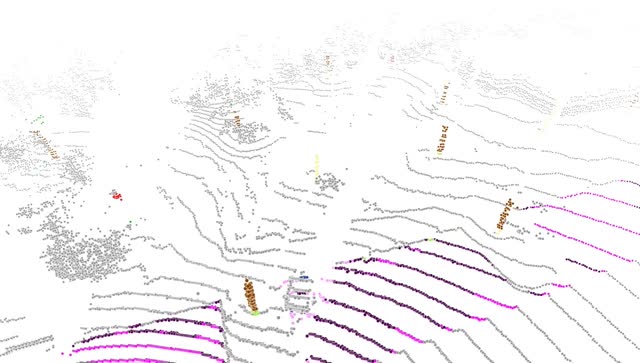}
        \end{overpic}
        \begin{overpic}[width=0.19\textwidth]{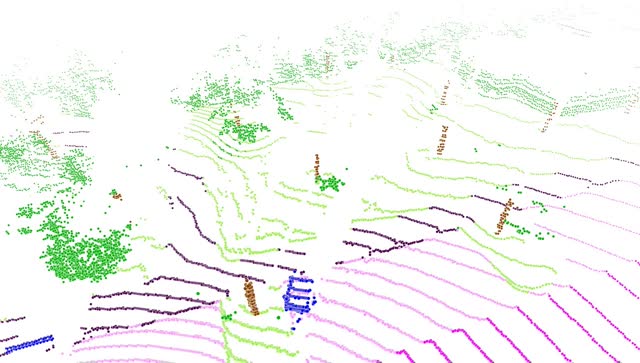}
        \end{overpic}
        \\
        \begin{overpic}[width=0.19\textwidth]{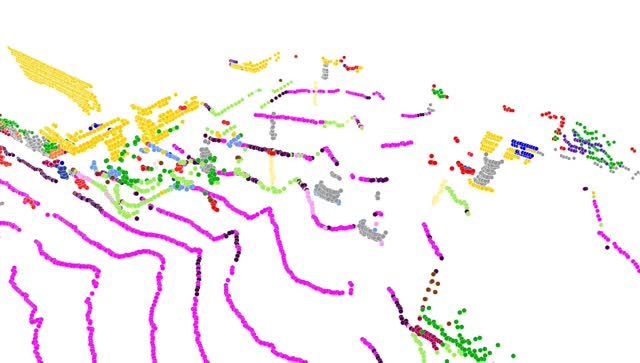}
        \end{overpic} &  
        \begin{overpic}[width=0.19\textwidth]{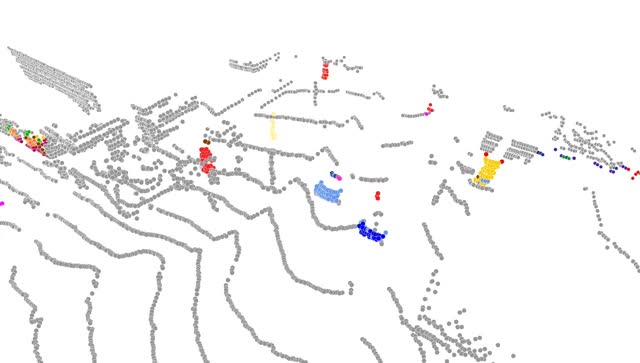}
        \end{overpic} &
        \begin{overpic}[width=0.19\textwidth]{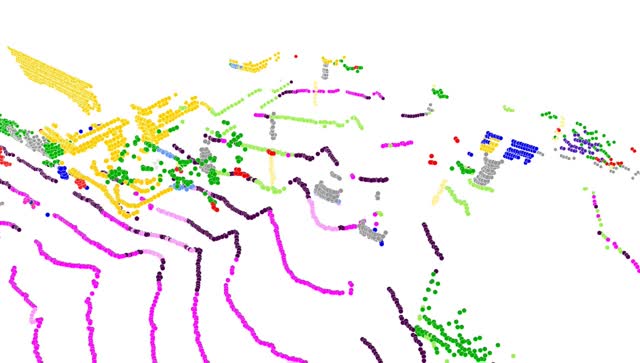}
        \end{overpic} &
        \begin{overpic}[width=0.19\textwidth]{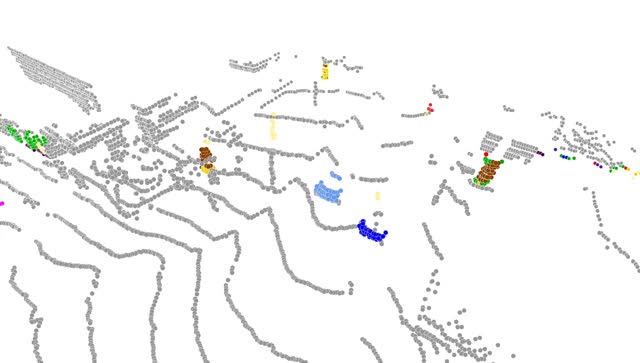}
        \end{overpic}
        \begin{overpic}[width=0.19\textwidth]{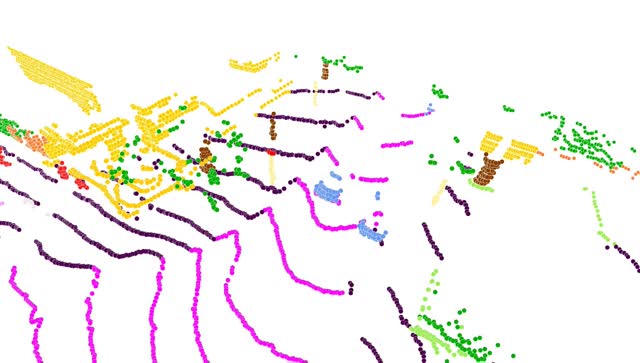}
        \end{overpic}
        \\
        \begin{overpic}[width=0.19\textwidth]{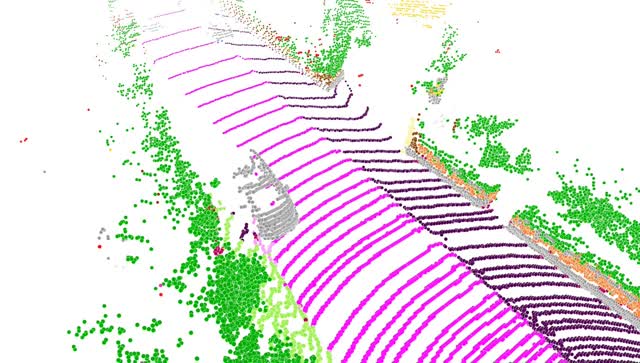}
        \end{overpic} &  
        \begin{overpic}[width=0.19\textwidth]{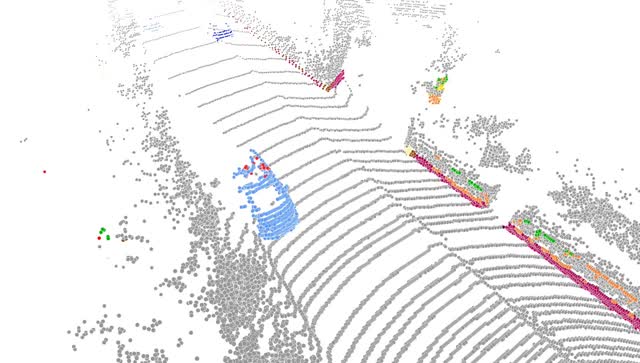}
        \end{overpic} &
        \begin{overpic}[width=0.19\textwidth]{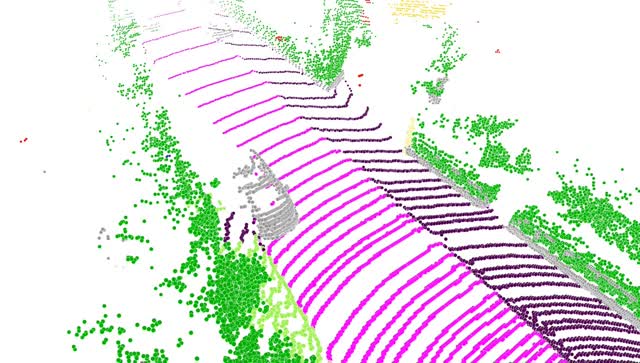}
        \end{overpic} &
        \begin{overpic}[width=0.19\textwidth]{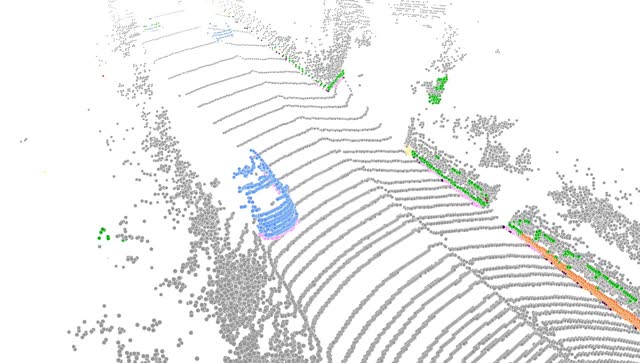}
        \end{overpic}
        \begin{overpic}[width=0.19\textwidth]{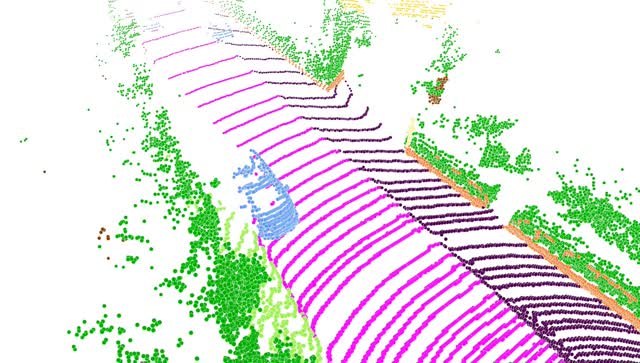}
        \end{overpic}
        \\
        \multicolumn{5}{c}{
        \begin{overpic}[width=0.99\textwidth]{images/qualitative/kitti/legend_kitti.pdf}
        \end{overpic}}
    \end{tabular}
    \caption{Qualitative comparison on SemanticKITTI from KITTI-$5^1$. EUMS$^\dag$~\cite{zhao2022novel} outputs are completely or partially wrong for the novel classes. \ourmethod improves the performance by providing correct and more homogeneous predictions.}
    \label{fig:supp_qualitative_kitti1}
\end{figure*}

\begin{figure*}[t]
\centering
    \setlength\tabcolsep{2.5pt}
    \begin{tabular}{ccccc}
    \raggedright
        \begin{overpic}[width=0.19\textwidth]{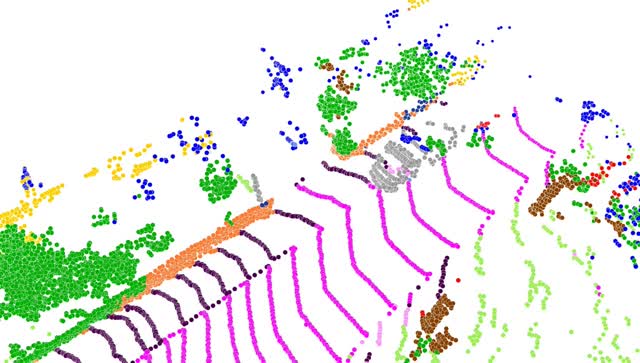}
        \put(20,60){\color{black}\footnotesize \textbf{EUMS$^\dag$~\cite{zhao2022novel} base}}
        \put(130,60){\color{black}\footnotesize \textbf{EUMS$^\dag$~\cite{zhao2022novel} novel}}
        \put(225,60){\color{black}\footnotesize \textbf{\ourmethod base (Ours)}}
        \put(330,60){\color{black}\footnotesize \textbf{\ourmethod novel (Ours)}}
        \put(460,60){\color{black}\footnotesize \textbf{GT}}
        \end{overpic} &  
        \begin{overpic}[width=0.19\textwidth]{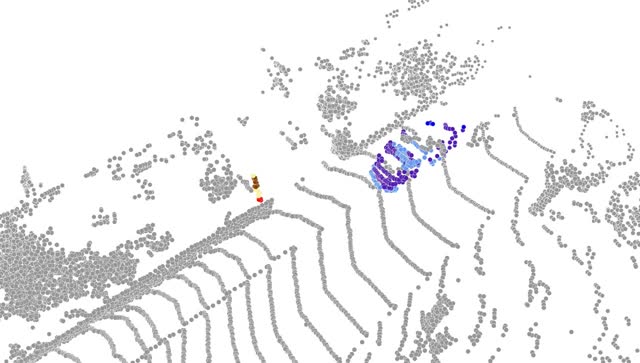}
        \end{overpic} &
        \begin{overpic}[width=0.19\textwidth]{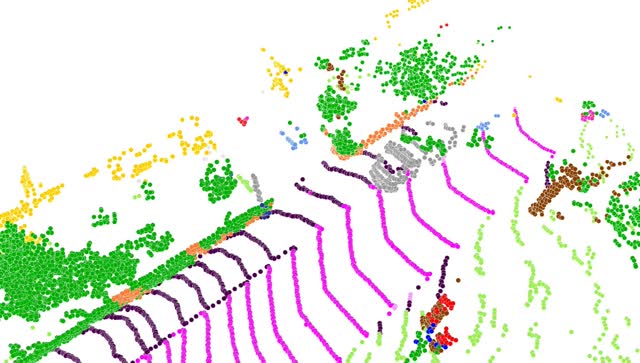}
        \end{overpic} &
        \begin{overpic}[width=0.19\textwidth]{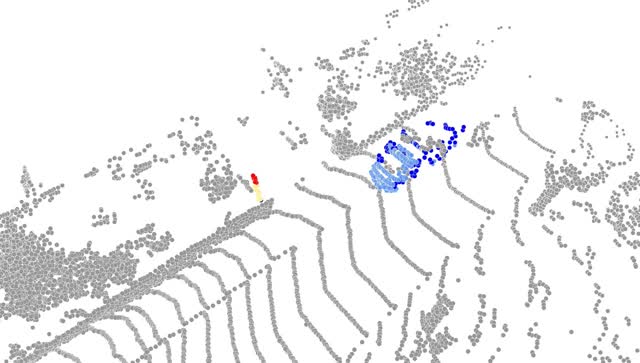}
        \end{overpic}
        \begin{overpic}[width=0.19\textwidth]{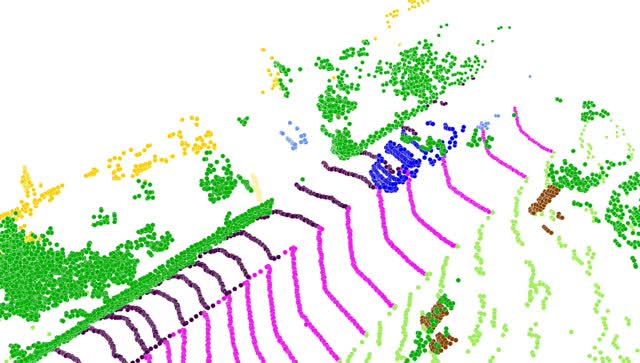}
        \end{overpic}
        \\
        \begin{overpic}[width=0.19\textwidth]{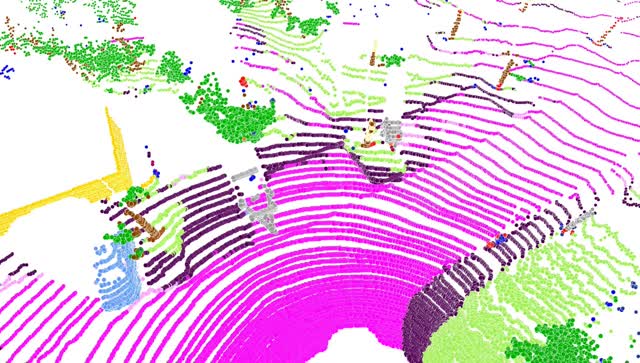}
        \end{overpic} &  
        \begin{overpic}[width=0.19\textwidth]{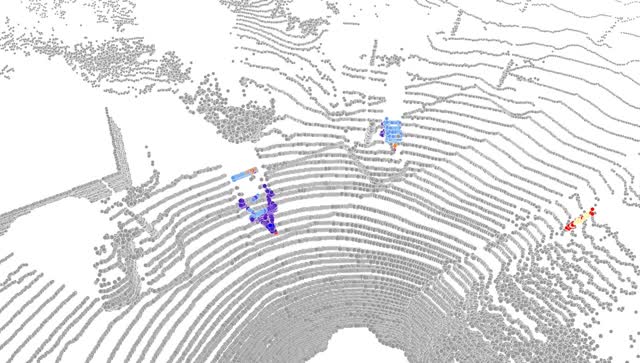}
        \end{overpic} &
        \begin{overpic}[width=0.19\textwidth]{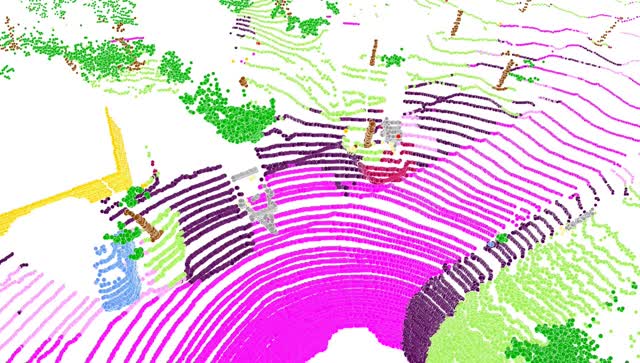}
        \end{overpic} &
        \begin{overpic}[width=0.19\textwidth]{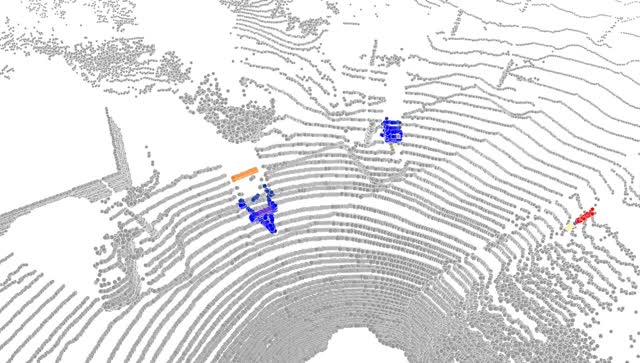}
        \end{overpic}
        \begin{overpic}[width=0.19\textwidth]{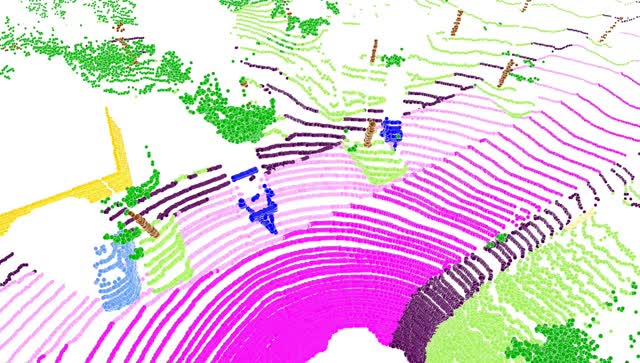}
        \end{overpic}
        \\
        \begin{overpic}[width=0.19\textwidth]{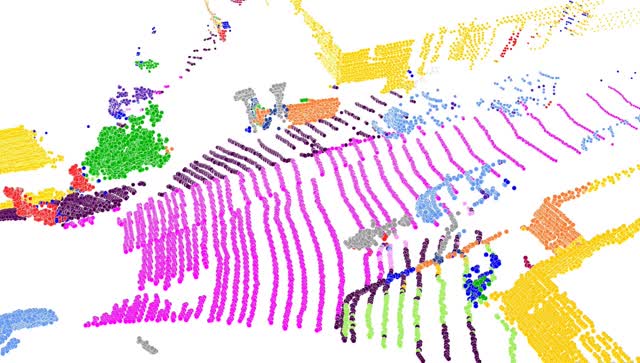}
        \end{overpic} &  
        \begin{overpic}[width=0.19\textwidth]{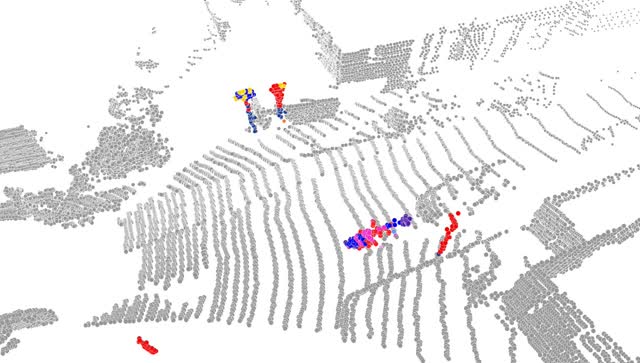}
        \end{overpic} &
        \begin{overpic}[width=0.19\textwidth]{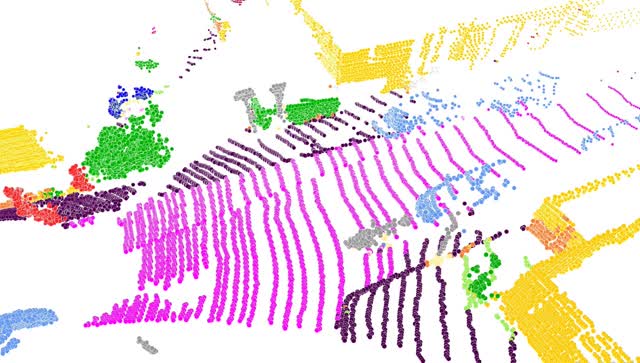}
        \end{overpic} &
        \begin{overpic}[width=0.19\textwidth]{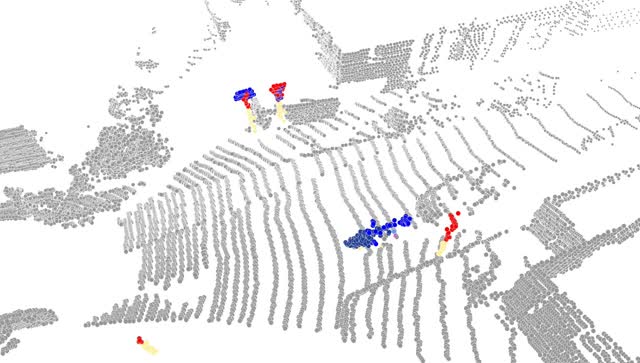}
        \end{overpic}
        \begin{overpic}[width=0.19\textwidth]{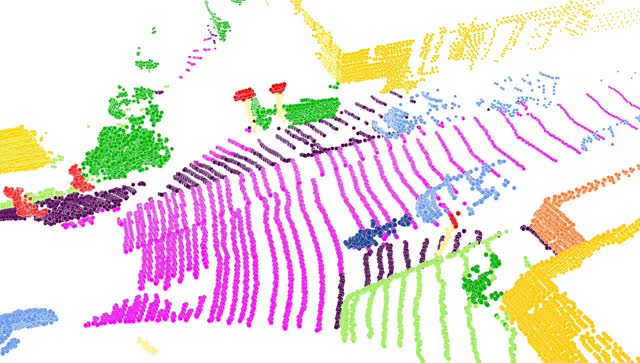}
        \end{overpic}
        \\
        \begin{overpic}[width=0.19\textwidth]{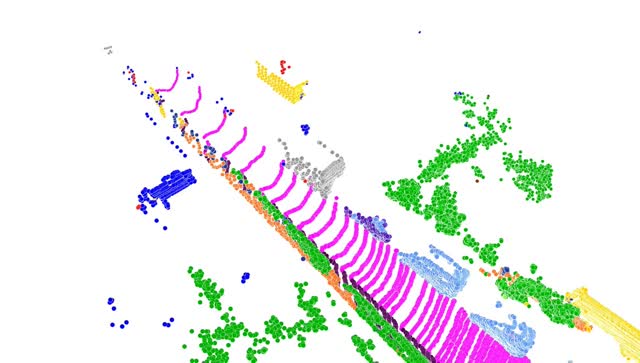}
        \end{overpic} &  
        \begin{overpic}[width=0.19\textwidth]{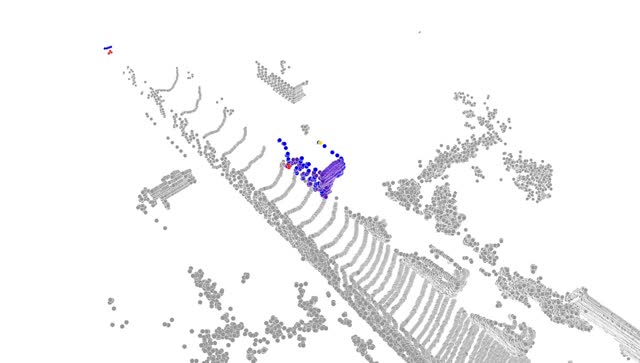}
        \end{overpic} &
        \begin{overpic}[width=0.19\textwidth]{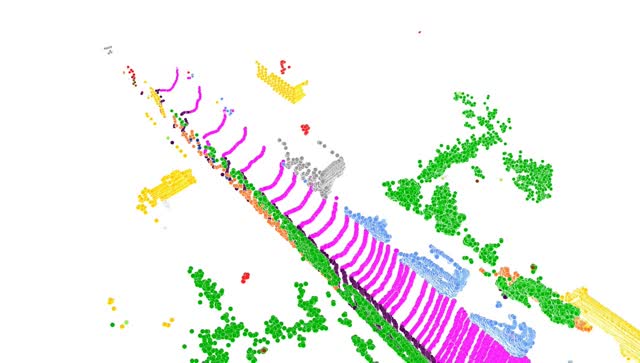}
        \end{overpic} &
        \begin{overpic}[width=0.19\textwidth]{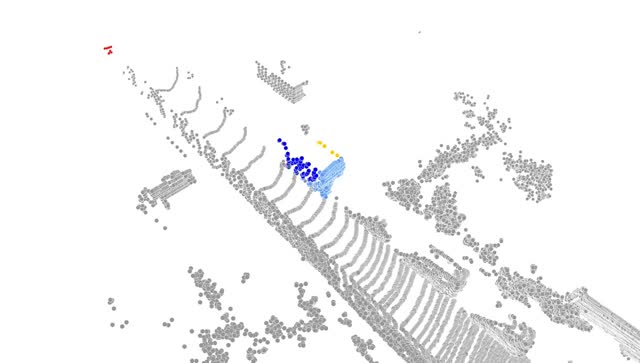}
        \end{overpic}
        \begin{overpic}[width=0.19\textwidth]{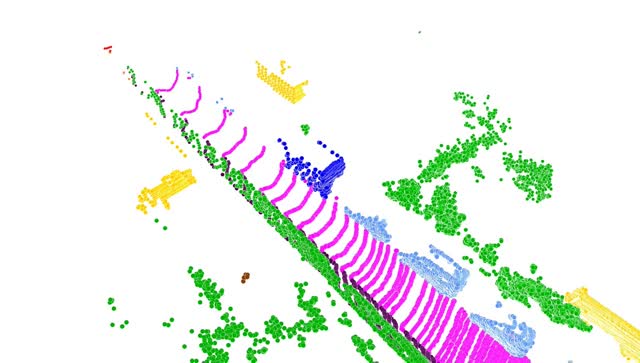}
        \end{overpic}
        \\
        \multicolumn{5}{c}{
        \begin{overpic}[width=0.99\textwidth]{images/qualitative/kitti/legend_kitti.pdf}
        \end{overpic}}
    \end{tabular}
    \caption{Qualitative comparison on SemanticKITTI from KITTI-$5^2$. EUMS$^\dag$~\cite{zhao2022novel} outputs are completely or partially wrong for the novel classes. \ourmethod improves the performance by providing correct and more homogeneous predictions.}
    \label{fig:supp_qualitative_kitti2}
\end{figure*}

\begin{figure*}[t]
\centering
    \setlength\tabcolsep{2.5pt}
    \begin{tabular}{ccccc}
    \raggedright
        \begin{overpic}[width=0.19\textwidth]{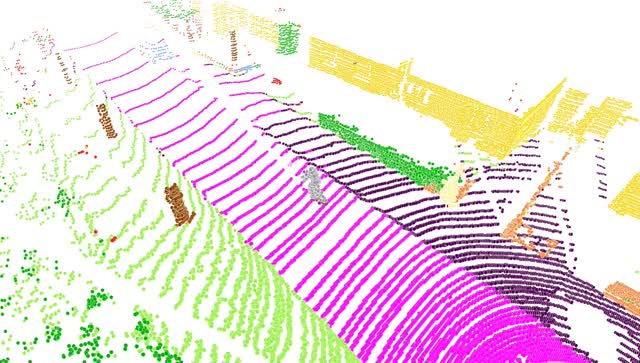}
        \put(20,60){\color{black}\footnotesize \textbf{EUMS$^\dag$~\cite{zhao2022novel} base}}
        \put(130,60){\color{black}\footnotesize \textbf{EUMS$^\dag$~\cite{zhao2022novel} novel}}
        \put(225,60){\color{black}\footnotesize \textbf{\ourmethod base (Ours)}}
        \put(330,60){\color{black}\footnotesize \textbf{\ourmethod novel (Ours)}}
        \put(460,60){\color{black}\footnotesize \textbf{GT}}
        \end{overpic} &  
        \begin{overpic}[width=0.19\textwidth]{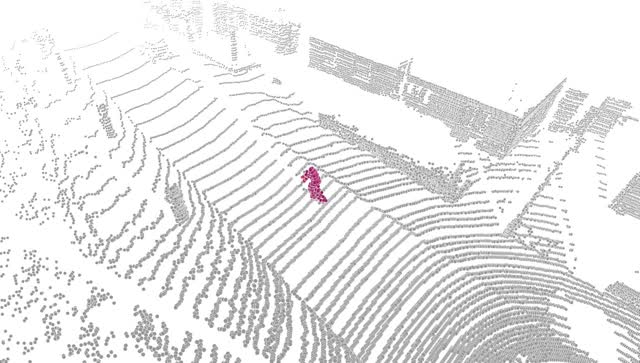}
        \end{overpic} &
        \begin{overpic}[width=0.19\textwidth]{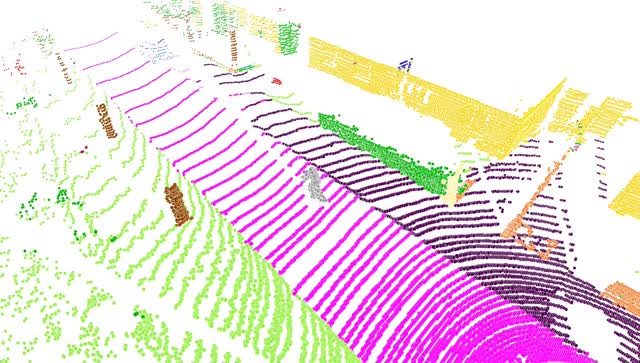}
        \end{overpic} &
        \begin{overpic}[width=0.19\textwidth]{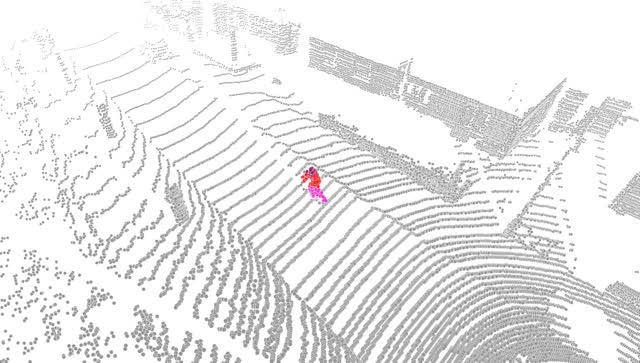}
        \end{overpic}
        \begin{overpic}[width=0.19\textwidth]{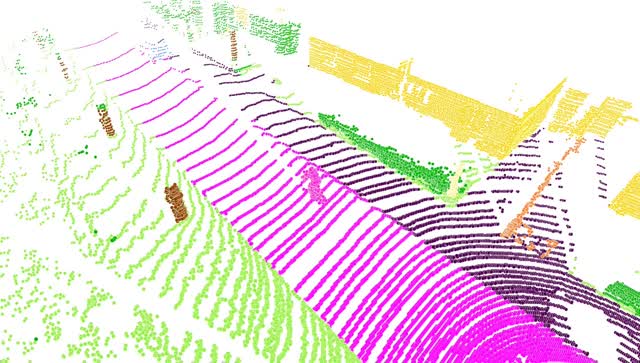}
        \end{overpic}
        \\
        \begin{overpic}[width=0.19\textwidth]{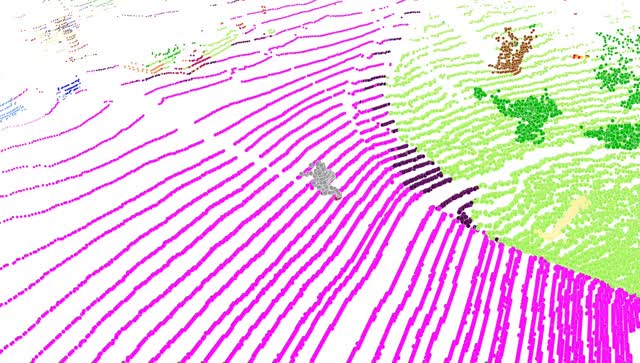}
        \end{overpic} &  
        \begin{overpic}[width=0.19\textwidth]{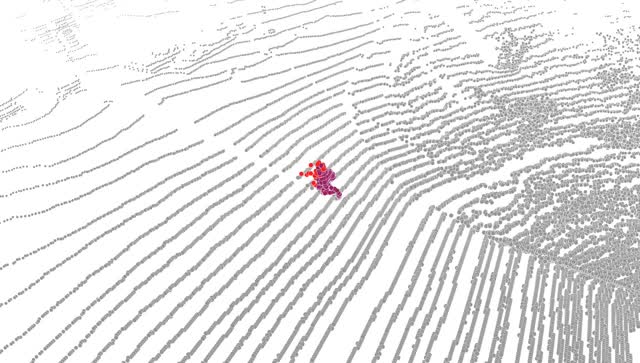}
        \end{overpic} &
        \begin{overpic}[width=0.19\textwidth]{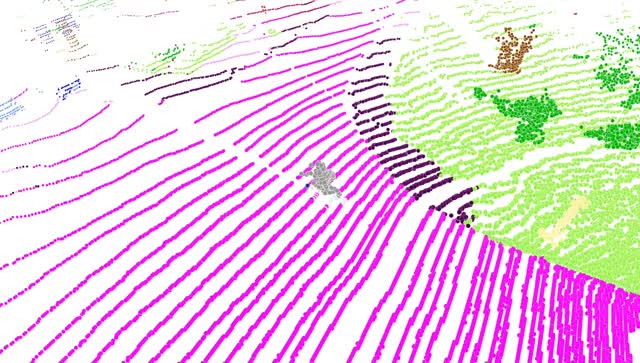}
        \end{overpic} &
        \begin{overpic}[width=0.19\textwidth]{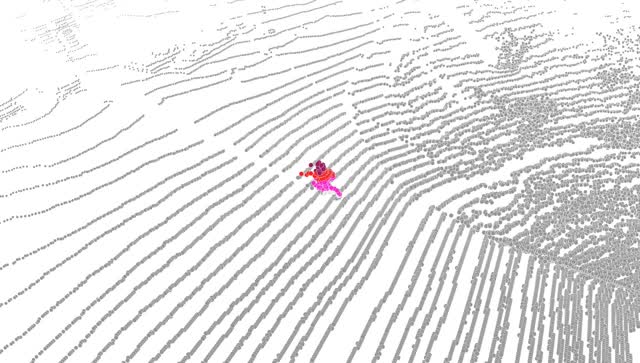}
        \end{overpic}
        \begin{overpic}[width=0.19\textwidth]{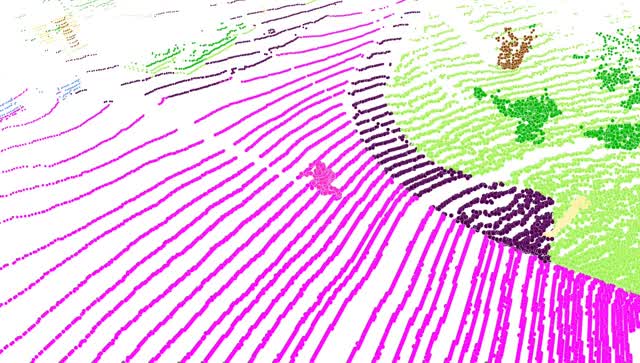}
        \end{overpic}
        \\
        \begin{overpic}[width=0.19\textwidth]{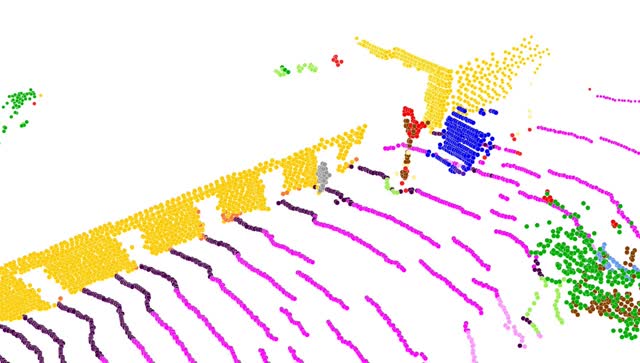}
        \end{overpic} &  
        \begin{overpic}[width=0.19\textwidth]{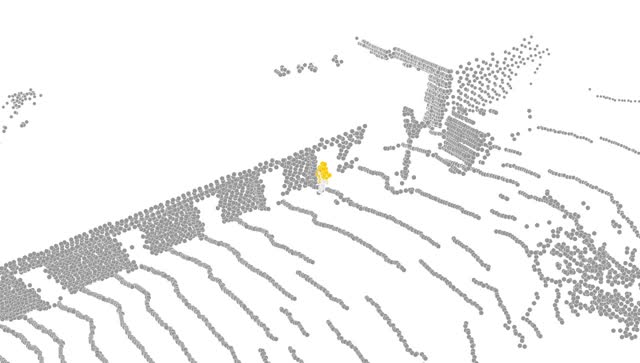}
        \end{overpic} &
        \begin{overpic}[width=0.19\textwidth]{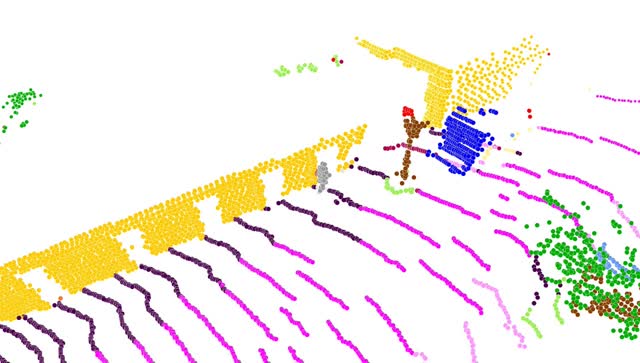}
        \end{overpic} &
        \begin{overpic}[width=0.19\textwidth]{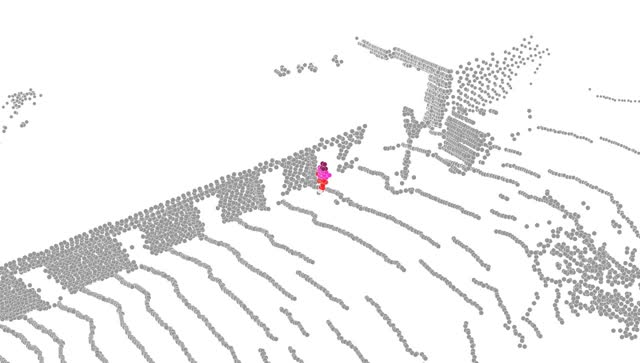}
        \end{overpic}
        \begin{overpic}[width=0.19\textwidth]{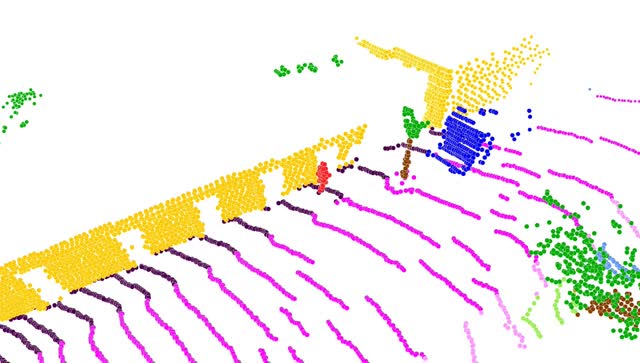}
        \end{overpic}
        \\
        \begin{overpic}[width=0.19\textwidth]{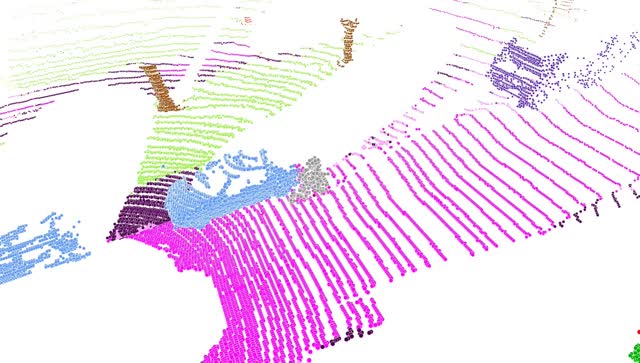}
        \end{overpic} &  
        \begin{overpic}[width=0.19\textwidth]{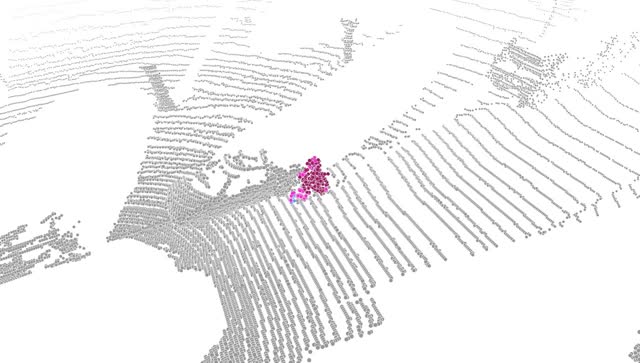}
        \end{overpic} &
        \begin{overpic}[width=0.19\textwidth]{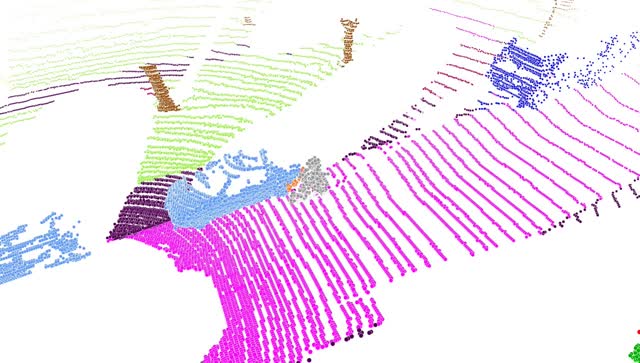}
        \end{overpic} &
        \begin{overpic}[width=0.19\textwidth]{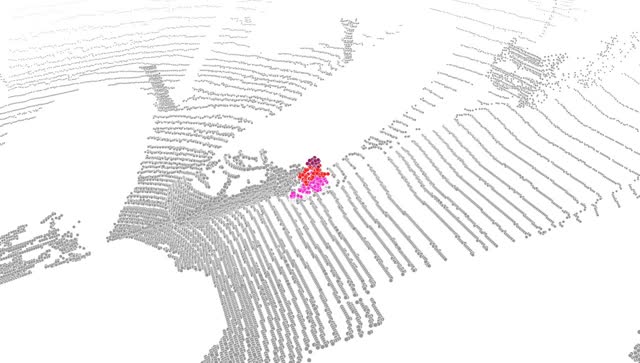}
        \end{overpic}
        \begin{overpic}[width=0.19\textwidth]{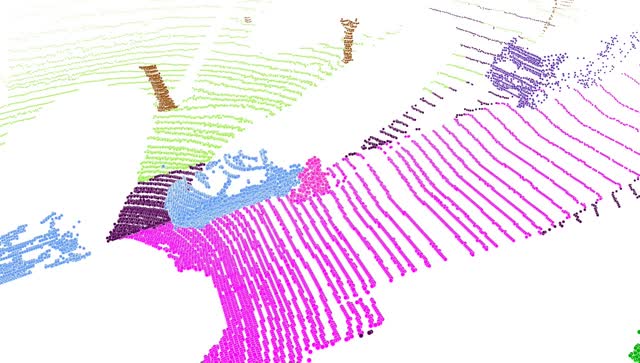}
        \end{overpic}
        \\
        \multicolumn{5}{c}{
        \begin{overpic}[width=0.99\textwidth]{images/qualitative/kitti/legend_kitti.pdf}
        \end{overpic}}
    \end{tabular}
    \caption{Qualitative comparison on SemanticKITTI from KITTI-$4^3$. EUMS$^\dag$~\cite{zhao2022novel} outputs are completely or partially wrong for the novel classes. \ourmethod improves the performance by providing correct and more homogeneous predictions.}
    \label{fig:supp_qualitative_kitti3}
\end{figure*}

Figs.~\ref{fig:supp_qualitative_kitti0}-\ref{fig:supp_qualitative_kitti3}, show qualitative results for SemanticKITTI from the split KITTI-$5^0$ (Fig.~\ref{fig:supp_qualitative_kitti0}), KITTI-$5^1$ (Fig.~\ref{fig:supp_qualitative_kitti1}), KITTI-$5^2$ (Fig.~\ref{fig:supp_qualitative_kitti2}) and KITTI-$4^3$ (Fig.~\ref{fig:supp_qualitative_kitti3}). We add ground-truth labels as the supervised reference.

\end{document}